\RequirePackage{fix-cm}
\documentclass[smallextended]{svjour3}       
\smartqed  
\usepackage{graphicx}
\usepackage{makeidx}  
\usepackage{amsmath} 
\usepackage{amssymb}
\usepackage{enumerate}
\usepackage{algorithmic}
\usepackage[toc,page]{appendix}
\usepackage[ruled,vlined]{algorithm2e}
\usepackage{graphicx} 
\usepackage{caption}
\usepackage{subcaption}
\usepackage{wrapfig}
\usepackage{sidecap}
\usepackage{framed}
\usepackage{enumitem}
\usepackage[dvipsnames]{xcolor}
\usepackage[colorlinks]{hyperref}
\usepackage{bm}
\usepackage{multirow}
\usepackage{ctable}
\usepackage{empheq}

\newenvironment{Proof}[1]{\par\noindent\textit{Proof:}\space#1}{\hfill $\blacksquare$}

\DeclareMathOperator*{\argmin}{arg\,min}
\DeclareMathOperator*{\argmax}{arg\,max}
\def\bbbr{{\rm I\!R}}
\def\bbbh{{\rm I\!H}}

\def\R{\mathrm{R}}

\def\N{\mathbb{N}}

\def\I{\mathbb{I}}
\def\P{\mathrm{P}}

\begin{document}

\title{An Online Prediction Algorithm for Reinforcement Learning with Linear Function Approximation using Cross Entropy Method
}

\titlerunning{A Cross Entropy based Prediction Method for RL}        

\author{Ajin George Joseph         \and
        Shalabh Bhatnagar 
}


\institute{Ajin George Joseph \at
	      Department of Computer Science \& Automation\\
              Indian Institute of Science, Bangalore \\
              Tel.: +91-80-22932368\\
              Fax:  +91-80-23602911\\
              \email{ajin@iisc.ac.in}           
           \and
           Shalabh  Bhatnagar \at
	      Department of Computer Science \& Automation and \\
	      the Robert Bosch Centre for Cyber Physical Systems\\
              Indian  Institute of Science, Bangalore
}

\date{Received: date / Accepted: date}

\maketitle

\begin{abstract}
	In this paper, we provide two new stable online algorithms for the problem of prediction in reinforcement learning, \emph{i.e.}, estimating the value function of a model-free Markov reward process  using the linear function approximation architecture and with memory and computation costs scaling quadratically in the size of the feature set. The algorithms employ the multi-timescale stochastic approximation variant of the very popular cross entropy (CE) optimization method which is a model based search method to find the global optimum of a real-valued function. A proof of convergence of the algorithms using the ODE method is provided. We supplement our theoretical results  with experimental comparisons. The algorithms achieve good performance fairly consistently on many RL benchmark problems with regards to computational efficiency, accuracy and stability.
\end{abstract}
\section{Introduction}
In this paper, we follow the reinforcement learning (RL) framework as described in \cite{sutton1998introduction,white1993survey,bertsekas1995dynamic}. The basic structure in this setting is the discrete time Markov decision process (MDP) which is a 4-tuple ($\mathbb{S}$, $\mathbb{A}$, $\mathrm{R}$, $\mathrm{P}$), where $\mathbb{S}$ denotes the set of \textit{states} and $\mathbb{A}$ is the set of \textit{actions}. We assume that the state and action spaces are finite. The function $\mathrm{R}: \mathbb{S} \times \mathbb{A} \times \mathbb{S} \rightarrow \bbbr$ is called the \textit{reward function}, where $\mathrm{R}(s, a, s^{\prime})$ represents the reward obtained in state $s$ after taking action $a$ and transitioning to $s^{\prime}$. Without loss of generality, we assume that the reward function is bounded, \emph{i.e.}, $|\mathrm{R}(\cdot,\cdot,\cdot)| \leq \mathrm{R}_{\mathrm{max}} < \infty$. Also, $\mathrm{P}:\mathbb{S} \times \mathbb{A} \times \mathbb{S} \rightarrow [0,1]$ is the \textit{transition probability kernel}, where $\mathrm{P}(s, a, s^{\prime}) = \mathbb{P}(s^{\prime} | s, a)$ is the probability of next state being $s^{\prime}$ conditioned on the fact that the current state is $s$ and the action taken is $a$. 
A \textit{stationary policy} $\pi:\mathbb{S} \rightarrow \mathbb{A}$ is a function from states to actions, where $\pi(s)$ is the action taken whenever the system is in state $s$ (independent of time)\footnote{The policy can also be stochastic in order to incorporate  exploration. In that case, for a given $s \in \mathbb{S}$, $\pi(\cdot\vert s)$ is a probability distribution over the action space $\mathbb{A}$.}. A given policy $\pi$ along with the transition kernel $\mathrm{P}$ determine the state dynamics of the system. For a given policy $\pi$, the system behaves as a Markov chain with transition matrix $\mathrm{P}^{\pi}(s,s^{\prime})$ = $\mathrm{P}(s,\pi(s),s^{\prime})$.

For a given policy $\pi$, the system evolves at each discrete time step and this process can be captured as a coupled sequence of transitions and rewards $\{\mathbf{s}_{0}, \mathbf{r}_{0}, \mathbf{s}_{1}, \mathbf{r}_{1}, \mathbf{s}_{2}, \mathbf{r}_{2}, \dots \}$, where $\mathbf{s}_{t}$ is the random variable which represents the state at time $t$, $\mathbf{s}_{t+1}$ is the transitioned state from $\mathbf{s}_{t}$ and $\mathbf{r}_{t} = \mathrm{R}(\mathbf{s}_t, \pi(\mathbf{s}_t), \mathbf{s}_{t+1})$ is the reward associated with the transition. In this paper, we are concerned with the problem of \textit{prediction}, {\em i.e.}, estimating the expected long run $\gamma$-discounted cost $V^{\pi} \in \bbbr^{\mathbb{S}}$ (also referred to as the \textit{value function}) corresponding to the given policy $\pi$. Here, given $s \in \mathbb{S}$, we let 
{\setlength{\abovedisplayskip}{7pt}\setlength{\belowdisplayskip}{6pt}\begin{flalign}
V^{\pi}(s) \triangleq \mathbb{E}\left[\sum_{t=0}^{\infty}\gamma^{t}\mathbf{r}_{t} \big \vert \mathbf{s}_{0} = s\right],
\end{flalign}}
where $\gamma \in [0,1)$ is a constant called  the \textit{discount factor} and $\mathbb{E}[\cdot]$ is the expectation over sample trajectories of states obtained in turn from  $\mathrm{P}^{\pi}$ when starting from the initial state $s$. $V^{\pi}$ satisfies the well known \textit{Bellman equation} \cite{bertsekas1995dynamic} under policy $\pi$, given by
{\setlength{\abovedisplayskip}{7pt}\setlength{\belowdisplayskip}{6pt}\begin{flalign}
V^{\pi} = \mathrm{R}^{\pi} + \gamma\mathrm{P}^{\pi}V^{\pi} \triangleq T^{\pi}V^{\pi},
\end{flalign}}
where $\mathrm{R}^\pi \triangleq (\mathrm{R}^\pi(s),s\in\mathbb{S})^\top$ with $\mathrm{R}^{\pi}(s)$ = $\mathbb{E}\left[\mathbf{r}_{t} \vert \mathbf{s}_t = s\right]$, $V^\pi \triangleq (V^\pi(s),s\in\mathbb{S})^\top$ and $T^\pi V^\pi \triangleq ((T^\pi V^\pi)(s),$
$ s \in \mathbb{S})^\top$, respectively. Here $T^\pi$ is called the \textit{Bellman operator}.\\

\noindent
\textbf{Prediction problem}\footnote{The prediction problem is related to policy evaluation except that the latter procedure evaluates the value of a policy given complete model information. We are however in an online setting where model information is completely unknown, however, a realization of the model dynamics in the form of a sample trajectory as described above is made available in an incremental fashion. The goal then is to predict at each time instant the value of each state in $\mathbb{S}$ (both observed and unobserved) under this constraint using the sample trajectory revealed till that instant.}\cite{sutton1988learning,bhatnagar2009convergent,sutton2009fast}:  In this paper, we follow  a generalized RL framework, where we assume that the model, \emph{i.e.}, $\mathrm{P}$ and $\mathrm{R}$ are inaccessible; only a sample trajectory $\{(\mathbf{s}_t, \mathbf{r}_{t}, \mathbf{s}^{\prime}_{t})\}_{t=0}^{\infty}$ is available where at each instant $t$, the state $\mathbf{s}_t$ of the triplet $(\mathbf{s}_t, \mathbf{r}_{t}, \mathbf{s}^{\prime}_{t})$ is sampled using an arbitrary distribution $\nu$ over $\mathbb{S}$ called the \textit{sampling distribution}, while the next state $\mathbf{s}^{\prime}_{t}$ is drawn using $\mathrm{P}^{\pi}(\mathbf{s}_t, \cdot)$ following the underlying Markov dynamics and $\mathbf{r}_{t}$ is the immediate reward for the transition, \emph{i.e.}, $\mathbf{r}_{t} = \mathrm{R}(\mathbf{s}_{t}, \pi(\mathbf{s}_{t}), \mathbf{s}^{\prime}_{t})$. We assume that $\nu(s) > 0$, $\forall s \in \mathbb{S}$. The goal of the prediction problem is to estimate the value function $V^{\pi}$ from the given sample trajectory. 

\begin{remark}\label{rem:rem1}
	\textit{The framework that we consider in this paper is a generalized setting, commonly referred to as the off-policy setting \footnote{One may find the term off-policy to be a misnomer in this context. Usually on-policy refers to RL settings where the underlying Markovian system is assumed ergodic and the sample trajectory provided follows the dynamics of the system. Hence, off-policy can be interpreted as a contra-positive statement of this definition of on-policy and in that sense, our setting is indeed off-policy. See \cite{sutton2009convergent} }. In the literature, one often finds the on-policy setting where the underlying Markovian system induced by the evaluation policy is assumed ergodic, \emph{i.e.}, aperiodic and irreducible, which directly implies the existence of a unique steady state distribution (stationary distribution). In such cases, the sample trajectory is presumed to be a continuous roll-out of a particular instantiation of the underlying transition dynamics in the form of $\{\mathbf{s}_0, \mathbf{r}_1, \mathbf{s}_1, \mathbf{r}_2, \dots \}$, where $\mathbf{s}_0$ is chosen arbitrarily. Since the system is ergodic, the distribution of the states in the sample trajectory will eventually follow the steady state distribution. Hence, the on-policy setting becomes a special case of the off-policy setting, where the sampling distribution is nothing but the stationary distribution and $\mathbf{s}_{t+1} = \mathbf{s}^{\prime}_{t}$, $\forall t \in \mathbb{N}$.}
\end{remark}

Unfortunately, the number of states $\vert \mathbb{S} \vert$ may be large in many practical applications \cite{kaelbling1996reinforcement,doya2000reinforcement}, for example, elevator dispatching \cite{crites1996improving}, robotics \cite{kober2013reinforcement} and board games such as Backgammon ($10^{20}$ states \cite{tesauro1995td}) and computer Go ($10^{170}$ states \cite{silver2007reinforcement}). The impending combinatorial blow-ups exemplify the underlying problem with the value function estimation, commonly referred to as the \textit{curse of dimensionality}. In this case, the value function is unrealizable due to both storage and computational limitations. Apparently one has to resort to  approximate solution methods where we sacrifice precision for computational tractability. A common approach in this context is the function approximation method \cite{sutton1998introduction}, where we approximate the value function of unobserved states using the knowledge of the observed states and their transitions.

In the \textit{linear function approximation technique}, a linear architecture consisting of a set of $k$ feature vectors ($\vert \mathbb{S} \vert$-dimensional) $\{\phi_{i} \in \bbbr^{\vert \mathbb{S} \vert}, 1 \leq i \leq k\}$,  where $1 \leq k \ll \vert \mathbb{S} \vert$, is chosen \emph{a priori}.  For a state $s \in \mathbb{S}$, we define
\begin{equation}\label{eqn:phieq}
\phi(s) \triangleq \begin{bmatrix}
         \phi_{1}(s) \\
          \phi_{2}(s) \\
          \vdots \\
          \phi_{k}(s) \\          
\end{bmatrix}_{k \times 1}, \hspace{2mm}      
\Phi \triangleq  \begin{bmatrix}
\phi(s_{1})^{\top} \\
\phi(s_{2})^{\top} \\
\vdots \\
\phi(s_{\vert \mathbb{S} \vert})^{\top}
\end{bmatrix}_{\vert \mathbb{S} \vert \times k},
\end{equation}
where the vector $\phi(s)$ is called the \textit{feature vector} corresponding to the state $s \in \mathbb{S}$, while the matrix $\Phi$ is called the \textit{feature matrix}.

Primarily, the task in linear function approximation is to find a weight vector $z \in \bbbr^{k}$ such that the predicted value function $\Phi z \approx V^{\pi}$. Given $\Phi$, the best approximation of $V^{\pi}$ is its projection on to the  closed subspace $\bbbh^{\Phi} = \{\Phi z |  z \in \bbbr^{k}\}$ (column space of $\Phi$) with respect to some norm on $\bbbr^{\vert \mathbb{S} \vert}$. Typically, one uses the weighted semi-norm $\Vert\cdot\Vert_{\nu}$ on $\bbbr^{\vert \mathbb{S} \vert}$, where $\nu(\cdot)$ is the sample probability distribution with which the states $\mathbf{s}_{t}$ occur in the sample trajectory. It is assumed that $\nu(s) > 0, \forall s \in \mathbb{S}$. The semi-norm  $\Vert\cdot\Vert_{\nu}$ on $\bbbr^{\vert \mathbb{S} \vert}$ is defined as $\Vert V \Vert_{\nu}^{2} = \sum_{s \in \mathbb{S}}V(s)^{2}\nu(s)$. The associated linear projection operator $\Pi^{\nu}$ is defined as $\Pi^{\nu} V^{\pi} = \argmin_{h \in \bbbh^{\Phi}} \Vert V^{\pi} - h \Vert_{\nu}^{2}$. It is not hard to derive the following closed form expression for $\Pi^{\nu}$.
{\setlength{\abovedisplayskip}{7pt}\setlength{\belowdisplayskip}{6pt}\begin{flalign}\label{eq:norm}
\Pi^{\nu} = \Phi(\Phi^{\top}D^{\nu}\Phi)^{-1}\Phi^{\top}D^{\nu},
\end{flalign}}
where $D^{\nu}$ is the diagonal matrix with $D^{\nu}_{ii} = \nu(s_i), i=1,\dots,\vert \mathbb{S} \vert$. On a technical note, observe that the projection is obtained by minimizing the squared $\nu$-weighted distance from the true value function $V^{\pi}$ and this distance is referred to as the \textit{mean squared error (MSE)}, \emph{i.e.},
{\setlength{\abovedisplayskip}{7pt}\setlength{\belowdisplayskip}{6pt}\begin{flalign}\label{eq:msedef}
\textrm{MSE}(z) \triangleq \Vert V^{\pi} - \Phi z \Vert^{2}_{\nu}, \hspace*{4mm} z \in \bbbr^{k}.
\end{flalign}}

However, it is hard to evaluate or even estimate $\Pi^{\nu}$ since it requires complete knowledge of the sampling distribution $\nu$ and also requires $\vert \mathbb{S} \vert$ amount of memory for storing $D^{\nu}$. Therefore, one has to resort to additional approximation techniques to estimate the projection $\Pi^{\nu} V^{\pi}$ which is indeed the prime objective of this paper.\\

\noindent
\textbf{Goal of this paper: }To find a vector $z^{*} \in \bbbr^{k}$ such that $\Phi z^{*} \approx \Pi^{\nu}V^{\pi}$ without knowing $\Pi^{\nu}$ or trying to estimate the same.\\

A caveat is in order. It is important to note that the efficacy of the learning method depends on the choice of the feature set $\{\phi_{i}\}$ \cite{lagoudakis2003least}. One can either utilize prior knowledge about the system to develop hard-coded features or employ off-the-shelf basis functions\footnote{Some of the commonly used basis functions are radial basis functions (RBF), polynomials, Fourier basis functions \cite{konidaris2011value} and cerebellar model articulation controller (CMAC) \cite{eldracher1994function}, to name a few.} from the literature. In this paper, we assume that a carefully chosen set of features is available \emph{a priori}.

\section{Related Work}
The existing algorithms can be broadly classified as
\begin{enumerate}
\item
	Linear methods which include temporal difference method (TD($\lambda$), $\lambda \in [0,1]$ \cite{sutton1988learning,tsitsiklis1997analysis}), gradient temporal difference methods (GTD \cite{sutton2009convergent}, GTD2 \cite{sutton2009fast}, TDC \cite{sutton2009fast}) and residual gradient (RG) schemes\cite{baird1995residual},  whose computational complexities are linear in $k$ and hence are good for large values of $k$ and 
\item
Second order methods which include least squares temporal difference (LSTD) \cite{bradtke1996linear,boyan2002technical} and least squares policy evaluation (LSPE) \cite{nedic2003least} whose computational complexities are quadratic in $k$ and are useful for moderate values of $k$. Second order methods, albeit computationally expensive, are seen to be more data efficient than others except in the case when trajectories are very small \cite{dann2014policy}.
\end{enumerate}

In cases where the Markov chain is ergodic (\emph{i.e.} irreducible and aperiodic) and the sampling distribution $\nu$ is the stationary distribution of the Markov chain, then with $\Phi$ being a full column rank matrix, the convergence of TD($\lambda$) is guaranteed \cite{tsitsiklis1997analysis}. But in cases where the sampling distribution $\nu$ is not the stationary distribution of the Markov chain or the projected subspace is a non-linear manifold, then TD($\lambda$) can diverge \cite{tsitsiklis1997analysis,baird1995residual}. However, both LSTD and LSPE algorithms are stable \cite{schoknecht2002optimality} and are also seen to be independent of the sampling distribution $\nu$. However, there do not exist any extensions of LSTD and LSPE to the non-linear function approximation.

Van Roy and Tsitsiklis \cite{tsitsiklis1997analysis} gave a different characterization for the stable limit point of TD($0$) as the fixed point of the \textit{projected Bellman operator} $\Pi^{\nu}T^{\pi}$,
\begin{equation}
\Phi z = \Pi^{\nu}T^{\pi} \Phi z,
\end{equation}
where $\nu$ is the stationary distribution of the underlying ergodic chain.

This characterization yields a new error function, the \textit{mean squared projected Bellman error (MSPBE)} which is defined as follows:
\begin{equation}
\textrm{MSPBE}(z) \triangleq \Vert \Phi z - \Pi^{\nu}T^{\pi} \Phi z \Vert_{\nu}^{2}\hspace*{1mm}, \hspace*{4mm} z \in \bbbr^{k}.
 \end{equation}
The LSTD algorithm \cite{bradtke1996linear,boyan2002technical} is a fitted value function method (least squares approach) obtained by directly solving MSPBE over the sample trajectory using sample averaging of the individual transitions. However, the LSPE method \cite{nedic2003least} solves MSPBE indirectly using a double minimization procedure where the primary minimizer finds the projection of the Bellman operator value using the least squares approach with the proximal Bellman operator value being obtained from the secondary gradient based minimizer. In \cite{sutton2009fast}, MSPBE is exquisitely manoeuvred to derive multiple stable $\Theta(k)$ algorithms like TDC and GTD2. A non-linear function approximation version of the GTD2 algorithm is also available \cite{bhatnagar2009convergent}. The method is shown to be stable and the convergence to the sub-optimal solutions is also guaranteed  under reasonably realistic assumptions \cite{bhatnagar2009convergent}. The sub-optimality of the solutions is expected as GTD2 is a gradient-based method and the convexity of the objective function does not always hold in non-linear function approximation settings.

Another pertinent error function is the \textit{mean squared Bellman residue ($\mathrm{MSBR}$)} which is defined as follows:
\begin{equation}\label{eq:msbrdef}
\mathrm{MSBR}(z) \triangleq \mathbb{E}\left[(\mathbb{E}\left[\delta_t(z)| \mathbf{s}_t\right])^{2}\right], z \in \bbbr^{k},
\end{equation}
where $\delta_{t}(z) \triangleq \mathbf{r}_{t} + \gamma z^{\top}\phi(\mathbf{s}^{\prime}_{t}) - \phi(\mathbf{s}_{t})$ is the temporal difference error under function approximation when $z$ is the associated approximation parameter. Note that \textrm{MSBR} is a measure of how closely the prediction vector represents the solution to the Bellman equation. 

\textit{Residual gradient (RG)} algorithm \cite{baird1995residual}  minimizes the error function MSBR directly using stochastic gradient search. Indeed, RG solves $\nabla_{z}\textrm{MSBR} = 0$ $\Rightarrow \mathbb{E}\Big[\mathbb{E}\left[\delta_t(z)| \mathbf{s}_t\right]\Big]\mathbb{E}\Big[\mathbb{E}\left[(\gamma\phi(\mathbf{s}_{t})-\phi(\mathbf{s}_t))| \mathbf{s}_t\right]\Big] = 0$. The above expression is a product of two expectations conditioned on the current state $\mathbf{s}_{t}$. Hence it requires two independent samples $\mathbf{s}^{\prime}_{t}$ and $\mathbf{s}^{\prime\prime}_{t}$ of the next state when in the current state $\mathbf{s}_{t}$. This is generally referred to as \textit{double sampling}. Even though the RG algorithm guarantees convergence, due to large variance, the convergence rate is small \cite{schoknecht2003td}. 

Eligibility traces \cite{sutton1988learning}  are a mechanism to accelerate learning by blending temporal difference methods with Monte Carlo simulation (averaging the values) and weighted using a geometric distribution with parameter $\lambda \in [0,1]$. 
Eligibility traces can be integrated into most of these algorithms\footnote{The algorithms with eligibility traces are named with $(\lambda)$ appended, for example TD$(\lambda)$, LSTD$(\lambda)$ \emph{etc}.}. In this paper, we do not consider the treatment of eligibility traces.

Table \ref{tab:compalgtable} provides a list of important TD based algorithms along with the associated error objectives. The algorithm complexities and other characteristics are also shown in the table.
\begin{table}
\begin{center}
\renewcommand{\arraystretch}{1.1}
\setlength\tabcolsep{1.1pt}
\begin{tabular}{llllll}
\specialrule{.1em}{.02em}{.02em} 
\hline\noalign{\smallskip}
   $\vert$ Algorithm \hspace*{1cm}$\vert$ & Complexity \hspace*{5mm}$\vert$ & Error \hspace{10mm}$\vert$ &Elig. Trace $\vert$ & Stability $\vert$ & NLFA$^{\ddagger}$ $\vert$\\
\noalign{\smallskip}
\hline
\specialrule{.1em}{.02em}{.02em} 
\noalign{\smallskip}
 	\hspace*{2mm}LSTD\hspace*{1cm} & $\Theta(k^{3})$ \hspace*{5mm}& MSPBE \hspace*{10mm}& $\surd$ & $\surd$ & $\times$ \\ 
	\hspace*{2mm}TD \hspace*{1cm} &$\Theta(k)$ \hspace*{5mm}& MSPBE \hspace*{10mm}& $\surd$ & $\times$ &$\surd$\\ 
    \hspace*{2mm}LSPE \hspace*{1cm}& $\Theta(k^{3})$\hspace*{5mm} & MSPBE \hspace*{10mm}& $\surd$ & $\surd$ & $\times$\\
	\hspace*{2mm}GTD \hspace*{1cm}& $\Theta(k)$\hspace*{5mm} & MSPBE \hspace*{10mm}& - & $\surd$ & $\times$\\
	\hspace*{2mm}GTD2 \hspace*{1cm}& $\Theta(k)$ \hspace*{5mm}& MSPBE \hspace*{10mm}& $\surd$ & $\surd$ & $\surd$\\
	\hspace*{2mm}TDC \hspace*{1cm}& $\Theta(k)$ \hspace*{5mm}& MSPBE \hspace*{10mm}& $\surd$ & $\surd$ & $\times$\\
    \hspace*{2mm}RG \hspace*{1cm}& $\Theta(k)$ \hspace*{5mm}& MSBR \hspace*{10mm}& $\surd$ & $\surd$ & $\times$\\	  
\hline
\specialrule{.1em}{.01em}{.01em} 
\end{tabular}
\end{center}
\caption{Comparison of the state-of-the-art function approximation RL algorithms.$^{\ddagger}$ NLFA: Non-linear function approximation}\label{tab:compalgtable}
\end{table}

Put succinctly, when linear function approximation is applied in an RL setting, the main task can be cast as an optimization problem whose objective function is one of the aforementioned error functions. Typically, almost all the state-of-the-art algorithms employ gradient search technique to solve the minimization problem. In this paper, we apply a gradient-free technique called the \textit{cross entropy (CE) method} instead to find the minimum. By `\textit{gradient-free}', we mean the algorithm does not incorporate information on the gradient of the objective function, rather it uses the function values themselves. The cross entropy method as such lies within the general class of \textit{model based search methods} \cite{zlochin2004model}. Other methods in this class are \textit{model reference adaptive search (MRAS)} \cite{hu2007model}, \textit{gradient-based adaptive stochastic search for simulation optimization (GASSO)} \cite{zhou2014simulation}, \textit{ant colony optimization (ACO)} \cite{dorigo1997ant} and \textit{estimation of distribution algorithms (EDAs)} \cite{muhlenbein1996recombination}. Model based search methods have been applied to the control problem\footnote{The problem here is to find the optimal basis of the MDP.} in \cite{hu2008model,mannor2003cross,busoniu2009policy} and in basis adaptation\footnote{The basis adaptation problem is to find the best parameters of the basis functions for a given policy.} \cite{menache2005basis}, but this is the first time such a procedure has been applied to the prediction problem. However, due to the naive batch based approach of the original CE method, it cannot be directly applied to the online RL setting. In this paper, therefore, we propose two incremental, adaptive, online algorithms which solve MSBR and MSPBE respectively by employing a stochastic approximation version of the cross entropy method proposed in \cite{genstochce2016,ajincedet,predictsce2018}.
\subsubsection{Our Contributions}
The \textit{cross entropy (CE) method} \cite{rubinstein2013cross,de2005tutorial} is a model based search algorithm to find the global maximum of a given real valued objective function. In this paper, we propose for the first time, an adaptation of this method to the problem of parameter tuning in order to find the best estimates of the value function $V^{\pi}$ for a given policy $\pi$ under the linear function approximation architecture. We propose two prediction algorithms using the multi-timescale stochastic approximation framework \cite{robbins1951stochastic,borkar1997stochastic,kushner2012stochastic} which minimize MSPBE and MSBR respectively. The algorithms possess the following attractive features:
\begin{enumerate}
\item
A remodelling of the famous CE method to a model-free MDP framework using the stochastic approximation framework. 
\item
Stable with minimal restrictions on both the structural properties of the underlying Markov chain and on the sample trajectory.
\item
Minimal restriction on the feature set.
\item
Computational complexity is quadratic in the number of features (this is a significant improvement compared to the cubic complexity of the least squares algorithms).
\item
Competitive with least squares and other state-of-the-art algorithms in terms of accuracy.
\item
Algorithms are incremental update, adaptive, streamlined and online.
\item
Algorithms provide guaranteed convergence to the global minimum of MSPBE (or MSBR).
\item
Relative ease in extending the algorithms to  non-linear function approximation settings.
\end{enumerate} 
A noteworthy observation is that under linear architecture, both MSPBE and MSBR are strongly convex functions \cite{dann2014policy} and hence their local and global minima overlap. Hence, the fact that CE method finds the global minima as opposed to local minima, unlike gradient search, does not provide any tangible advantage in terms of the quality of the solution. Nonetheless, in the case of non-linear function approximators, the convexity property does not hold in general and so there may exist multiple local minima in the objective and the gradient search schemes would get stuck in local optima unlike CE based search. We have not explored analytically the non-linear case in this paper. Notwithstanding, we have applied our algorithm to the non-linear MDP setting defined in section $X$ of \cite{tsitsiklis1997analysis} and the results obtained are quite impressive. The MDP setting in \cite{tsitsiklis1997analysis} is a classic example where TD($0$) is shown to diverge and GTD2 is shown to produce sub-optimal solutions. This demonstrates the robustness of our algorithm which is quite appealing, considering the fact that the state-of-the-art RL algorithms are specifically designed to perform in a linear environment and extending them to domains beyond the realm of linearity is quite tedious and often mere impossible. In view of all these alluring features, our approach can be viewed as a significant first step towards efficiently using model based search for policy evaluation in a generalized RL environment.

\section{Summary of Notation}\label{sec:summary}
We use $\mathbf{X}$ for random variable and $x$ for deterministic variable. Let $\mathbb{I}_{k \times k}$ and $0_{k \times k}$ be the identity matrix and the zero matrix with dimensions $k \times k$ respectively. For set $A$,  $I_{A}$ represents the indicator function of $A$, \emph{i.e.}, $I_{A}(x) = 1$ if $x \in A$ and $0$ otherwise. Let $f_{\theta}:\bbbr^{n} \rightarrow \bbbr_{+}$ denote the \textit{probability density function} (PDF) over $\bbbr^{n}$ parametrized by $\theta$. Let $\mathbb{E}_{\theta}[\cdot]$ and $P_{\theta}$ denote the \textit{expectation} and the induced \textit{probability measure} \emph{w.r.t.} $f_{\theta}$. For $\rho \in (0,1)$ and $\mathcal{H}:\bbbr^{n} \rightarrow \bbbr$, let $\gamma_{\rho}(\mathcal{H}, \theta)$ denote the $(1-\rho)$-quantile of $\mathcal{H}(\mathbf{X})$ \emph{w.r.t.} $f_{\theta}$, \emph{i.e.},
\begin{equation}\label{eq:quantdef}
	\gamma_{\rho}(\mathcal{H}, \theta) \triangleq \sup\{\ell \in \bbbr \hspace*{1mm} \vert \hspace*{1mm} P_{\theta}(\mathcal{H}(\mathbf{X}) \geq \ell) \geq \rho \}.
\end{equation}
Let $int(A)$ be the \textit{interior} of set $A$. Let $\mathcal{N}_{n}(m, V)$ represent the $n$-variate Gaussian distribution with mean vector $m \in \bbbr^{n}$ and covariance matrix $V \in \bbbr^{n \times n}$. A function $L:\bbbr^{n} \rightarrow \bbbr$ is \textit{Lipschitz continuous}, if $\exists K \geq 0$ \emph{s.t.} $\vert L(x) - L(y) \vert \leq K\Vert x - y \Vert$, $\forall x, y \in \bbbr^{n}$, where $\Vert \cdot \Vert$ is some norm defined on $\bbbr^{n}$.

\section{Background: The CE Method}
To better understand our algorithm, we explicate the original CE method first.
\subsection{Objective of CE}
The \textit{cross entropy (CE) method} \cite{rubinstein2013cross,hu2009performance,de2005tutorial} solves problems of the following form:
\[\textrm{ Find } \hspace*{4mm} x^{*} \in \argmax_{x \in \mathcal{X} \subseteq \bbbr^{m}} \mathcal{H}(x),\]
where $\mathcal{H}(\cdot)$ is a multi-modal real-valued function and $\mathcal{X}$ is called the \textit{solution space}.

The goal of the CE method is to find an optimal ``\textit{model}" or probability distribution over the solution space $\mathcal{X}$ which concentrates on the global maxima of $\mathcal{H}(\cdot)$. The CE method adopts an iterative procedure where at each iteration $t$, a search is conducted on a space of parametrized probability distributions $\{f_{\theta} \vert \theta \in \Theta\}$ over $\mathcal{X}$, where $\Theta$ (subset of the multi-dimensional Euclidean space) is the parameter space, to find a distribution parameter $\theta_t$ which reduces the \emph{Kullback-Leibler (KL)} divergence (also called the cross entropy distance) \cite{kullback1959statistics}  from the optimal model. The most commonly used class here is the \emph{natural exponential family of distributions (NEF)}.\\\\
\textbf{Natural exponential family of distributions} \cite{morris1982natural}:  These are denoted as $\mathcal{C} \triangleq$ $\{f_{\theta}(x) = h(x)e^{\theta^{\top}\Gamma(x)-K(\theta)} \mid \theta\in \Theta \subseteq \bbbr^d\}, \textrm{ where }$ $h:\bbbr^{m} \longrightarrow \bbbr$, $\Gamma:\bbbr^{m} \longrightarrow \bbbr^{d}$ and  $K:\bbbr^{d} \longrightarrow \bbbr$. By rearranging the parameters, we can show that the Gaussian distribution with mean vector $\mu$ and the covariance matrix $\Sigma$ belongs to $\mathcal{C}$. In this case,
\begin{equation} \label{eq:gdist}
f_{\theta}(x) = \frac{1}{\sqrt{(2\pi)^{m}|\Sigma|}}\exp{\{-\frac{1}{2}(x-\mu)^{\top}\Sigma^{-1}(x-\mu)\}},
\end{equation}
and one may let
${\displaystyle h(x) = \frac{1}{\sqrt{(2\pi)^{m}}}}$, $\Gamma(x) = (x, xx^{\top})^{\top}$ and  ${\displaystyle \theta = (\Sigma^{-1} \mu,\hspace*{1mm}-\frac{1}{2}\Sigma^{-1})^{\top}}$.\vspace*{1mm}\\
\begin{itemize}
\item[$\circledast$]\textbf{\textit{Assumption (A1)}:} The parameter space $\Theta$ is compact.
\end{itemize}
\subsection{CE Method (Ideal Version)}
The CE method aims to find a sequence of model parameters ${\{\theta_t\}}_{t \in \mathbb{N}}$, where $\theta_t \in \Theta$ and an increasing sequence of thresholds ${\{\gamma_{t}\}}_{t \in \mathbb{N}}$ where $\gamma_t \in \bbbr$, with the property that the event $\{\mathcal{H}(\mathbf{X}) \geq \gamma_{t}\}$ is a very high probability event with respect to the probability measure induced by the model parameter $\theta_{t}$. By assigning greater weight to higher values of $\mathcal{H}$ at each iteration, the expected behaviour of the probability distribution sequence should improve. The most common choice for $\gamma_{t+1}$ is $\gamma_{\rho}({\mathcal{H}}, \theta_t)$, the $(1-\rho)$-quantile of $\mathcal{H}(\mathbf{X})$ \emph{w.r.t.} the probability density function $f_{\theta_{t}}$,  where $\rho \in (0,1)$ is set \emph{a priori} for the algorithm. We take Gaussian distribution as the preferred choice for $f_{\theta}$ in this paper. In this case, the model parameter is $\theta = (\mu, \Sigma)^{\top}$ where $\mu \in \bbbr^{m}$ is the mean vector and $\Sigma \in \bbbr^{m \times m}$ is the covariance matrix.

The CE algorithm is an iterative procedure which starts with an initial value $\theta_0 = (\mu_{0}, \Sigma_0)^{\top}$ of the mean vector and the covariance matrix tuple and at each iteration $t$, a new parameter $\theta_{t+1} = (\mu_{t+1}, \Sigma_{t+1})^{\top}$ is derived from the previous value $\theta_t$ as follows (from Section 4 of \cite{hu2007model}):
\begin{equation}\label{eq:opt1}
	\theta_{t+1} = \argmax_{\theta \in \Theta}\mathbb{E}_{\theta_{t}}\left[S(\mathcal{H}(\mathbf{X}))I_{\{\mathcal{H}(\mathbf{X}) \geq \gamma_{t+1}\}}\log{f_\theta(\mathbf{X})}\right],
\end{equation}
where $S:\bbbr \rightarrow \bbbr_{+}$  is a  positive and strictly monotonically increasing function.\hspace{2mm}\\
If the gradient \emph{w.r.t.} $\theta$ of the objective function in Eq. (\ref{eq:opt1}) is equated to $0$, considering Gaussian PDF for $f_\theta$ (\emph{i.e.}, using the expression provided in Eq. (\ref{eq:gdist}) for $f_{\theta}$)and $\gamma_{t+1} = \gamma_{\rho}(\mathcal{H}, \theta_t)$ , we obtain the following:
\begin{align} 
&\mu_{t+1} = \frac{\mathbb{E}_{\theta_{t}}\left[\mathbf{g_{1}}\left(\mathcal{H}(\mathbf{X}), \mathbf{X}, \gamma_{\rho}(\mathcal{H}, \theta_t)\right)\right]}{\mathbb{E}_{\theta_{t}}\left[\mathbf{g_{0}}\left(\mathcal{H}(\mathbf{X}), \gamma_{\rho}(\mathcal{H}, \theta_t)\right)\right]} \triangleq \Upsilon_{1}(\mathcal{H}, \theta_{t}),\hspace{0mm} \label{eq:sigmaideal1}\\\nonumber\\
&\Sigma_{t+1} = \frac{\mathbb{E}_{\theta_{t}}\left[\mathbf{g_{2}}\left(\mathcal{H}(\mathbf{X}), \mathbf{X}, \gamma_{\rho}(\mathcal{H}, \theta_t), \Upsilon_{1}(\mathcal{H}, \theta_t)\right)\right]}{\mathbb{E}_{\theta_{t}}\left[\mathbf{g_{0}}\left(\mathcal{H}(\mathbf{X}), \gamma_{\rho}(\mathcal{H}, \theta_t)\right)\right]} \triangleq \Upsilon_{2}(\mathcal{H}, \theta_{t}).\hspace{0mm} \label{eq:sigmaideal2}
\end{align}
where
\begin{align}
&\hspace*{0mm}\mathbf{g_{0}}\bm{(}\mathcal{H}(x), \gamma\bm{)} \triangleq S(\mathcal{H}(x))I_{\bm{\{}\mathcal{H}(x) \geq \gamma\bm{\}}}, \hspace*{0mm}\\
&\mathbf{g_{1}}\bm{(}\mathcal{H}(x), x, \gamma\bm{)} \triangleq S(\mathcal{H}(x))I_{\bm{\{}\mathcal{H}(x) \geq \gamma\bm{\}}}x, \hspace*{0mm}\\
&\hspace*{0mm}\mathbf{g_{2}}\bm{(}\mathcal{H}(x), x, \gamma, \mu\bm{)} \triangleq S(\mathcal{H}(x))I_{\bm{\{}\mathcal{H}(x) \geq \gamma\bm{\}}}(x-\mu)(x-\mu)^{\top}.
\end{align}

\begin{remark}\textit{The function $S(\cdot)$ in Eq. (\ref{eq:opt1}) is positive and strictly monotonically increasing and is used to account for the cases when the objective function $\mathcal{H}(x)$ takes negative values for some $x$. Note that in the expression of $\mu_{t+1}$ in Eq. (\ref{eq:sigmaideal1}), $x$ is being weighted with $S(\mathcal{H}(x))$ in the region $\{x \vert \mathcal{H}(x) \geq \gamma_{t+1}\}$. Since the function $S$ is positive and strictly monotonically increasing, the region where $\mathcal{H}(x)$ is higher (hence $S(\mathcal{H}(x))$ is also higher) is given more weight and hence $\mu_{t+1}$ concentrates in the region where $\mathcal{H}(x)$ takes higher values. In case where $\mathcal{H}(\cdot)$ is positive, we can choose $S(x) = x$. However, in general scenarios, where $\mathcal{H}(\cdot)$ takes positive and negative values, the identity function is not an appropriate choice since the effect of the positive weights is reduced by the negative ones. In such cases, we take $S(x) = exp(rx), r \in \bbbr_{+}$.}
\end{remark}
Thus the ideal CE algorithm can be expressed using the following recursion:
\begin{flalign}\label{eqn:cerecn}
\theta_{t+1} = (\Upsilon_1(\mathcal{H}, \theta_t), \Upsilon_2(\mathcal{H}, \theta_t))^{\top}.
\end{flalign}
An illustration demonstrating the evolution of the model parameters of the CE method with Gaussian distribution during the optimization of a multi-modal objective function is provided in Fig. \ref{fig:evtrack} of Appendix.
\section{Comparison of the Objectives: MSPBE and MSBR}
\begin{figure}[!h]
	\centering
	{\includegraphics[height=45mm, width=85mm]{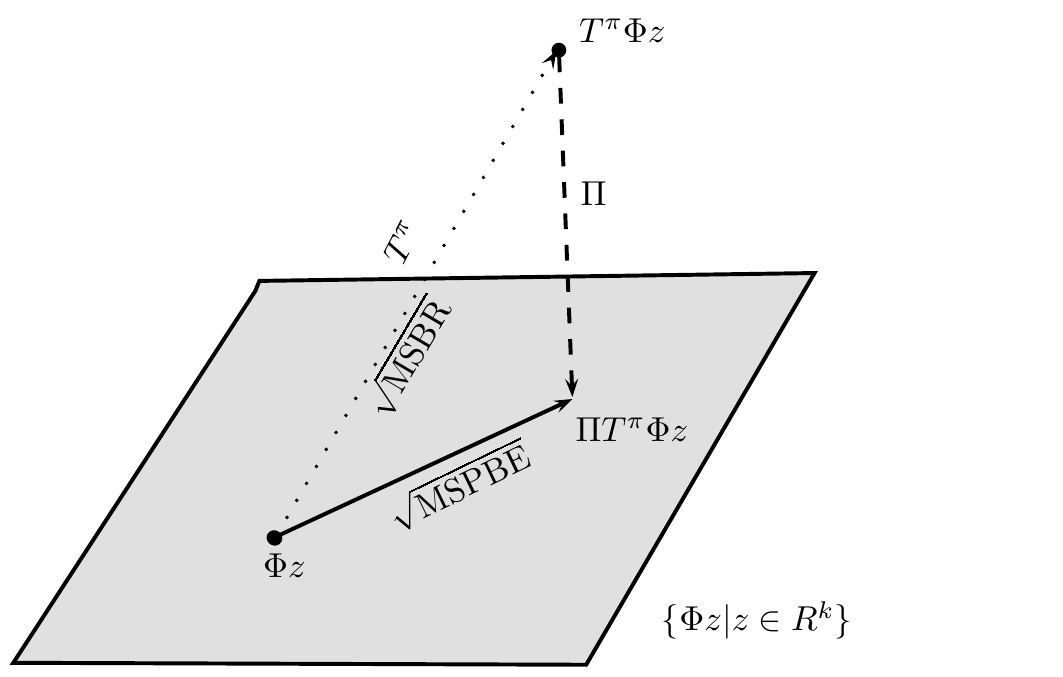}}
	\caption{Diagram depicting the relationship between the error functions \textrm{MSPBE} and \textrm{MSBR}.}\label{fig:errfnrel}
\end{figure}

This question is critical since  most reinforcement learning algorithms can be characterized via some optimization problem which minimizes either MSBR or MSPBE. A comprehensive comparison of the two error functions is available in the literature \cite{schoknecht2003td,schoknecht2002optimality,scherrer2010should}. A direct relationship between MSBR and MSPBE  can be easily established as follows:
\begin{equation}\label{eq:msbrmspberel}
\mathrm{MSBR}(z) = \mathrm{MSPBE}(z) + \Vert T^{\pi}\Phi z - \Pi^{\nu} T^{\pi}\Phi z \Vert^{2}, \hspace*{2mm} z \in \bbbr^{k}.
\end{equation}
This follows directly from Babylonian-Pythagorean  theorem and the fact that $\left(T^{\pi}\Phi z - \Pi^{\nu} T^{\pi}\Phi z\right)$ $\bot$ $\left(\Pi^{\nu} T^{\pi}\Phi z - \Phi z\right)$, $\forall z \in \bbbr^{k}$.
A vivid depiction of this relationship is shown in Fig. \ref{fig:errfnrel}.\\

If  the columns of the feature matrix $\Phi$ are linearly independent, then both the error functions MSBR and MSPBE are strongly convex\cite{dann2014policy}. However, the respective minima of MSBR and MSPBE are related depending on whether the feature set is perfect or not. A feature set is \emph{perfect} if $V^{\pi} \in \{\Phi z \vert z \in \bbbr^{k}\}$. In the perfect case, $\exists z_{0} \in \bbbr^{k}$ \emph{s.t.} $\Phi z_{0} = V^{\pi}$ and hence MSBR($z_{0}$) = 0. Since MSBR$(z) \geq 0$, $\forall z \in \bbbr^{k}$, we have $z_0 = \argmin_{z}{\textrm{MSBR}(z)}$. Now from (\ref{eq:msbrmspberel}), we get $\textrm{MSPBE}(z_0) = 0$ and $z_0 = \argmin_{z}{\textrm{MSPBE}(z)}$ (again since MSPBE$(z) \geq 0$, $\forall z \in \bbbr^{k}$). Hence in the perfect feature set scenario, the respective minima of MSBR and MSPBE coincide. However, in the imperfect case, they might differ since MSPBE($z$) $\neq$ MSBR($z$) for some $z \in \mathcal{Z}$ (follows from Eq. (\ref{eq:msbrmspberel})).\\

In \cite{scherrer2010should,williams1993tight}, a  relationship between MSBR and MSE is provided as shown in (\ref{eq:msemsbr}). Recall that MSE is the error which defines the projection operator $\Pi^{\nu}$ in the linear function approximation setting. It is found that, for a given $\nu$ with $\nu(s) > 0, \forall s \in \mathbb{S}$, 
\begin{equation}\label{eq:msemsbr}
\sqrt{\mathrm{MSE}(z)} \leq \frac{\sqrt{C(\nu)}}{1-\gamma}\sqrt{\mathrm{MSBR}(z)},
\end{equation}
where $C(\nu) = \max_{s,s^{\prime}}{\frac{\mathrm{P}^{\pi}(s, s^{\prime})}{\nu(s)}}$. This bound (albeit loose) ensures that the minimization of MSBR is indeed stable and the solution so obtained cannot be too far from the projection $\Pi^{\nu}V^{\pi}$. A noticeable drawback with MSBR is the statistical overhead brought about by the double sampling required for its estimation. To elaborate this, recall that MSBR($z$) = $\mathbb{E}\Big[\mathbb{E}\left[\delta_t(z)| \mathbf{s}_t\right]\mathbb{E}\left[\delta_t(z)| \mathbf{s}_t\right]\Big]$ (from Eq. (\ref{eq:msbrdef})). In the above expression of MSBR, we have a product of two conditional expectations conditioned on the current state $\mathbf{s}_{t}$. This implies that to estimate MSBR, one requires two independent samples of the next state, given the current state $\mathbf{s}_{t}$. Another drawback which was observed in the literature is the large variance incurred while estimating MSBR \cite{dann2014policy,scherrer2010should}, which inversely affects the rate of convergence of the optimization procedure. Also, in settings where only a finite length sample trajectory is available, the larger stochastic noise associated with the MSBR estimation will produce inferior quality solutions. 
MSPBE is attractive in the sense that double sampling is not required and there is sufficient empirical evidence  \cite{dann2014policy} to believe that the minimum of MSPBE often has low MSE. The absence of double sampling is quite appealing, since for large complex MDPs obtaining sample trajectories is itself tedious, let alone double samples. Also, MSPBE when integrated with control algorithms is also shown to produce better quality policies\cite{lagoudakis2003least}. Another less significant advantage is the fact that MSPBE($z$) $\leq$ MSBR($z$), $\forall z$ (follows from Eq. (\ref{eq:msbrmspberel})). This implies that the optimization algorithm can work with smaller objective function values compared to MSBR.

\noindent
Now, we explore here both the error functions analytically:
\subsection{MSPBE}
In \cite{sutton2009fast}, a compact expression for MSPBE is provided as follows: 
\begin{equation}
\mathrm{MSPBE}(z) = (\Phi^{\top} D^{\nu} (T^{\pi} V_{z}-V_{z}))^{\top}(\Phi^{\top}D^{\nu}\Phi)^{-1}
(\Phi^{\top}D^{\nu}(T^{\pi} V_{z}-V_{z})),
\end{equation}
where  $V_{z} = \Phi z$, while $\Phi$ and $D^{\nu}$ are defined in Eqs. (\ref{eqn:phieq}) and (\ref{eq:norm}) respectively. 
Now the expression $\Phi^{\top}D^{\nu} (T^{\pi}V_{z}-V_{z})$ is further rewritten as 
\begin{multline}\label{eqn:comp1}
\Phi^{\top}D^{\nu}(T^{\pi}V_{z}-V_{z}) = \mathbb{E}\left[\mathbb{E}\left[\phi_t(\mathbf{r}_{t}+
\gamma z^{\top}\phi^{\prime}_{t} - z^{\top}\phi_{t})| \mathbf{s}_t\right]\right]
\\ = \mathbb{E}\left[\mathbb{E}\left[\phi_t \mathbf{r}_{t} \vert \mathbf{s}_t\right]\right] +   \mathbb{E}\left[\mathbb{E}\left[\phi_t(\gamma\phi^{\prime}_{t} -\phi_t)^{\top} \vert \mathbf{s}_t\right]\right]z, \hspace*{3mm}\\ \mathrm{where} \hspace{1mm} \phi_t \triangleq \phi(\mathbf{s}_t) \textrm{ and } \phi^{\prime}_t \triangleq \phi(\mathbf{s}^{\prime}_t).
\hspace*{5mm}
\end{multline}
\begin{equation}\label{eqn:comp2}
\hspace*{0mm}\textrm{Also,} \hspace*{1cm} \Phi^{\top}D^{\nu}\Phi = \mathbb{E}\left[\phi_t \phi_{t}^{\top}\right].\hspace*{40mm}
\end{equation}
Putting all together we get,
\begin{flalign*}
\mathrm{MSPBE}(z) &= \left(\mathbb{E}\left[\mathbb{E}\left[\phi_t \mathbf{r}_{t} \vert \mathbf{s}_t\right]\right] +   \mathbb{E}\left[\mathbb{E}\left[\phi_t(\gamma\phi^{\prime}_{t} -\phi_t)^{\top} \vert \mathbf{s}_t\right]\right]z\right)^{\top}(\mathbb{E}\left[\phi_t \phi_{t}^{\top}\right])^{-1} \\ &\hspace*{30mm}\left(\mathbb{E}\left[\mathbb{E}\left[\phi_t \mathbf{r}_{t} \vert \mathbf{s}_t\right]\right] +   \mathbb{E}\left[\mathbb{E}\left[\phi_t(\gamma\phi^{\prime}_{t} -\phi_t)^{\top} \vert \mathbf{s}_t\right]\right]z\right)
\end{flalign*}
\begin{equation}\label{eq:stdetobj}
\hspace*{-20mm}= \left(\omega^{(0)}_{*} + \omega^{(1)}_{*} z\right)^{\top} \omega^{(2)}_{*} \left(\omega^{(0)}_{*} + \omega^{(1)}_{*} z\right),
\end{equation}
where $\omega^{(0)}_{*} \triangleq \mathbb{E}\left[\mathbb{E}\left[\phi_t \mathbf{r}_{t} \vert \mathbf{s}_t\right]\right]$, $\omega^{(1)}_{*} \triangleq  \mathbb{E}\left[\mathbb{E}\left[\phi_t(\gamma\phi^{\prime}_{t} -\phi_t)^{\top} \vert \mathbf{s}_t\right]\right]$ and \\ $\omega^{(2)}_{*} \triangleq (\mathbb{E}\left[\phi_t \phi_{t}^{\top}\right])^{-1}$.\\

This is a quadratic function in $z$. Note that in the above expression, the parameter vector $z$ and the stochastic component involving $\mathbb{E}[\cdot]$ are decoupled. Hence the stochastic component can be estimated or tracked independent of the parameter vector $z$. 

\subsection{MSBR}
We execute a similar decoupling procedure to the MSBR function. Indeed, from Eq. (\ref{eq:msbrdef}), we have
\begin{flalign*}
	&\mathrm{MSBR}(z) = \mathbb{E}\Big[\mathbb{E}^{2}[\delta_{t}(z) \vert \mathbf{s}_{t}]\Big] = \mathbb{E}\Big[\mathbb{E}^{2}[\mathbf{r}_{t}+\gamma z^{\top}\phi^{\prime}_{t} - z^{\top}\phi_{t}\big| \mathbf{s}_t]\Big]
	\\ &= \mathbb{E}\Big[\mathbb{E}^{2}[\mathbf{r}_{t} + \gamma z^{\top}\phi^{\prime}_{t} \big\vert \mathbf{s}_{t}]\Big]-2\mathbb{E}\Big[\mathbb{E}[\mathbf{r}_{t}+\gamma z^{\top}\phi^{\prime}_{t}\big\vert \mathbf{s}_{t}]\mathbb{E}[z^{\top}\phi_{t}\big| \mathbf{s}_t]\Big] + z^{\top}\mathbb{E}[\phi_t\phi^{\top}_t]z
	\\&=\mathbb{E}\Big[\mathbb{E}^{2}\left[\mathbf{r}_{t}\big\vert\mathbf{s}_{t}\right]\Big] + \gamma^{2}z^{\top}\mathbb{E}\Big[\mathbb{E}[\phi^{\prime}_{t}\big\vert \mathbf{s}_{t}]\mathbb{E}\left[\phi^{\prime}_{t}\big\vert \mathbf{s}_{t}\right]^{\top}\Big]z + 2z^{\top}\mathbb{E}\Big[\mathbb{E}[\mathbf{r}_{t} \vert \mathbf{s}_t]\mathbb{E}[\phi^{\prime}_{t}\big| \mathbf{s}_t]\Big] - \\  &\hspace*{20mm} 2z^{\top}\mathbb{E}\Big[\mathbb{E}\left[\mathbf{r}_{t} \vert \mathbf{s}_t\right]\phi_{t}\Big] - 2\gamma z^{\top}\mathbb{E}\Big[\mathbb{E}[\phi^{\prime}_{t}\big\vert \mathbf{s}_{t}]\phi_{t}^{\top}\Big]z + z^{\top}\mathbb{E}[\phi_t\phi^{\top}_t]z
	\\&=\mathbb{E}\Big[\mathbb{E}^{2}\left[\mathbf{r}_{t}\big\vert\mathbf{s}_{t}\right]\Big] + \gamma^{2}z^{\top}\mathbb{E}\Big[\mathbb{E}[\phi^{\prime}_{t}\big\vert \mathbf{s}_{t}]\mathbb{E}\left[\phi^{\prime}_{t}\big\vert \mathbf{s}_{t}\right]^{\top}\Big]z + \\ &\hspace*{20mm}2z^{\top}\mathbb{E}\Big[\mathbb{E}[\mathbf{r}_{t} \vert \mathbf{s}_t]\big(\mathbb{E}[\phi^{\prime}_{t}\big| \mathbf{s}_t]-\phi_{t}\big)\Big] + z^{\top}\mathbb{E}\Big[\big(\phi_t - 2\gamma \mathbb{E}[\phi^{\prime}_{t}\big\vert \mathbf{s}_{t}]\big)\phi_{t}^{\top}\Big]z.
\end{flalign*}
Therefore,
\begin{flalign}\label{eqn:msbrclform}
	\mathrm{MSBR}(z) &= \upsilon^{(0)}_{*} + z^{\top}\upsilon^{(1)}_{*}z + 2z^{\top}\upsilon^{(2)}_{*} + z^{\top}\upsilon^{(3)}_{*}z,\nonumber\\
	&= \upsilon^{(0)}_{*} +  z^{\top}(\upsilon^{(1)}_{*}+\upsilon^{(3)}_{*})z + 2z^{\top}\upsilon^{(2)}_{*},
\end{flalign}
where $\upsilon^{(0)}_{*} \triangleq \mathbb{E}\Big[\mathbb{E}^{2}\left[\mathbf{r}_{t}\big\vert\mathbf{s}_{t}\right]\Big]$, $\upsilon^{(1)}_{*} \triangleq  \gamma^{2}\mathbb{E}\Big[\mathbb{E}[\phi^{\prime}_{t}\big\vert \mathbf{s}_{t}]\mathbb{E}\left[\phi^{\prime}_{t}\big\vert \mathbf{s}_{t}\right]^{\top}\Big]$,\\ $\upsilon^{(2)}_{*} \triangleq \mathbb{E}\Big[\mathbb{E}[\mathbf{r}_{t} \vert \mathbf{s}_t]\big(\mathbb{E}[\phi^{\prime}_{t}\big| \mathbf{s}_t]-\phi_{t}\big)\Big]$ and $\upsilon^{(3)}_{*} \triangleq \mathbb{E}\Big[\big(\phi_t - 2\gamma \mathbb{E}[\phi^{\prime}_{t}\big\vert \mathbf{s}_{t}]\big)\phi_{t}^{\top}\Big]$.

\section{Proposed Algorithms}\label{sec:anatomy}
\textit{ We propose a generalized algorithm to approximate the value function $V^{\pi}$ (for a given policy $\pi$) with linear  function approximation by minimizing either MSPBE or MSBR, where the optimization is performed using a multi-timescale stochastic approximation variant of the CE algorithm}. Since the CE method is a maximization algorithm, the objective function in the optimization problem here is the negative of MSPBE and MSBR. To state it more formally: In this paper, we solve the following two optimization problems:\\
\begin{enumerate}
\item
{\setlength{\abovedisplayskip}{-12pt}\begin{flalign}\label{eqn:mspbeobj}
z_{p}^{*} = \argmin_{z \in \mathcal{Z} \subset \bbbr^{k}} \mathrm{MSPBE}(z) = \argmax_{z \in \mathcal{Z} \subset \bbbr^{k}}  \mathcal{J}_{p}(z),\\
\hspace{20mm} \mathrm{ where } \hspace{2mm} \mathcal{J}_{p} = -\mathrm{MSPBE}. \nonumber
\end{flalign}}
\item
{\setlength{\abovedisplayskip}{-10pt}\begin{flalign}\label{eqn:msbrobj}
z_{b}^{*} = \argmin_{z \in \mathcal{Z} \subset \bbbr^{k}} \mathrm{MSBR}(z) = \argmax_{z \in \mathcal{Z} \subset \bbbr^{k}}  \mathcal{J}_{b}(z),\\
\hspace{20mm} \mathrm{ where } \hspace{2mm} \mathcal{J}_{b} = -\mathrm{MSBR}. \nonumber
\end{flalign}}
\end{enumerate}

Here $\mathcal{Z}$ is the solution space, \emph{i.e.}, the space of parameter values of the function approximator. We also define $\mathcal{J}_{p}^{*} \triangleq \mathcal{J}_{p}(z_{p}^{*})$ and $\mathcal{J}_{b}^{*} \triangleq \mathcal{J}_{b}(z_{b}^{*})$.\\

\begin{itemize}
	\item[$\circledast$]\textbf{\textit{Assumption (A2)}:} The solution space $\mathcal{Z}$ is compact, \emph{i.e.}, it is closed and bounded.
\end{itemize}
\noindent
A few annotations about the algorithms are in order:\vspace*{4mm}\\
\textbf{1. Tracking the objective functions $\mathcal{J}_{p}, \mathcal{J}_{b}$: } Recall that the goal of the paper is to develop an online and incremental prediction algorithm. This implies that the algorithm has to estimate the value function by recalibrating the prediction vector incrementally  as new transitions of the sample trajectory are revealed. Note that the sample trajectory is simply a roll-out of an arbitrary realization of the underlying Markovian dynamics in the form of state transitions and their associated rewards and we assume that the sample trajectory satisfies the following assumption:
\begin{itemize}
\item[$\circledast$] \textit{\textbf{Assumption (A3):}}
	A sample trajectory $\{(\mathbf{s}_{t}, \mathbf{r}_{t}, \mathbf{s}^{\prime}_{t})\}_{t=0}^{\infty}$ is given, where  $\mathbf{s}_{t} \sim \nu(\cdot)$, $\mathbf{s}^{\prime}_{t} \sim \mathrm{P}^{\pi}(\mathbf{s}_{t}, \cdot)$ and $\mathbf{r}_{t} = \mathrm{R}(\mathbf{s}_{t}, \pi(\mathbf{s}_{t}), \mathbf{s}^{\prime}_{t})$. Let $\nu(s) > 0 $, $\forall s \in \mathbb{S}$. Also, let $\phi_t, \phi^{\prime}_{t}$, and $\mathbf{r}_{t}$ have uniformly bounded second moments. And the matrix $\mathbb{E}\left[\phi_{t}\phi_{t}^{\top}\right]$ is non-singular. 
\end{itemize}
In (A3), the uniform boundedness of the second  moments of $\phi_t, \phi^{\prime}_{t}$, and $\mathbf{r}_{t}$ directly follows in the case of finite state MDPs. However, the non-singularity requirement of  the matrix $\mathbb{E}\left[\phi_{t}\phi_{t}^{\top}\right]$ is strict and one can ensure this condition by appropriately choosing the feature set\footnote{A sufficient condition is the columns of the feature matrix $\Phi$ are linearly independent.}.\vspace*{0mm}\\

Now recall that in the analytic closed-form expression (Eq. (\ref{eq:stdetobj})) of the objective function $\mathcal{J}_{p}(\cdot)$, we have isolated the stochastic and the deterministic parts. The stochastic part can be identified by the tuple  $\omega_{*} \triangleq (\omega^{(0)}_{*}, \omega^{(1)}_{*}, \omega^{(2)}_{*})^{\top}$. So if we can find ways to track $\omega_{*}$, then it implies that we can track the objective function $\mathcal{J}_{p}(\cdot)$. This is the line of thought we follow here. In our algorithm, we track $\omega_{*}$ by maintaining a time indexed random vector $\omega_{t} \triangleq (\omega^{(0)}_{t}, \omega^{(1)}_{t}, \omega^{(2)}_{t})^{\top}$, where $\omega^{(0)}_{t} \in \bbbr^{k}$, $\omega^{(1)}_{t} \in \bbbr^{k \times k}$ and $\omega^{(2)}_{t} \in \bbbr^{k \times k}$. Here $\omega^{(i)}_{t}$ independently tracks $\omega^{(i)}_{*}$, $0 \leq i \leq 2$. We show here that $\lim_{t \rightarrow \infty}\omega^{(i)}_{t} = \omega^{(i)}_{*}$, with probability one, $0 \leq i \leq 2$. Now the stochastic recursion to track $\omega_{*}$ is given by
\begin{equation}
\omega_{t+1} = \omega_{t} + \alpha_{t+1}\Delta \omega_{t+1}.
\end{equation}
The increment term $\Delta \omega_{t+1} \triangleq (\Delta\omega^{(0)}_{t+1}, \Delta\omega^{(1)}_{t+1}, \Delta\omega^{(2)}_{t+1})^{\top}$ used for this recursion is defined as follows:
\begin{equation}\label{eq:omginc}
\hspace*{20mm}\left.
\begin{aligned}
\bigtriangleup{\omega}^{(0)}_{t+1}\hspace*{2mm} \triangleq& \hspace*{2mm}\mathbf{r}_{t}\phi_{t}-{\omega}^{(0)}_{t},\\
\bigtriangleup{\omega}^{(1)}_{t+1}\hspace*{2mm} \triangleq& \hspace*{2mm}\phi_{t}(\gamma\phi^{\prime}_{t}-\phi_{t})^{\top}-{\omega}^{(1)}_{t}, \\
\bigtriangleup{\omega}^{(2)}_{t+1}\hspace*{2mm} \triangleq& \hspace*{2mm}\mathbb{I}_{k \times k} -  \phi_{t} \phi_{t}^{\top}{\omega}^{(2)}_{t},
\end{aligned}
\hspace*{30mm}\right\}
\end{equation}
where $\phi_t \triangleq \phi(\mathbf{s}_{t})$ and $\phi^{\prime}_{t} \triangleq \phi(\mathbf{s}^{\prime}_{t})$.\\\\
Now we define the estimate of $\mathcal{J}_{p}(\cdot)$ at time $t$ as follows:\\
For a given $z \in \mathcal{Z}$,  
\begin{equation}
\bar{\mathcal{J}}_{p}(\omega_{t}, z) \triangleq -\left({\omega}^{(0)}_{t} + {\omega}^{(1)}_{t}z\right)^{\top}{\omega}^{(2)}_{t}\left({\omega}^{(0)}_{t} + {\omega}^{(1)}_{t}z\right).
\end{equation}
Superficially, it is similar to the expression of $\mathcal{J}_{p}$ in Eq. (\ref{eq:stdetobj}) except for $\omega_{t}$ replacing $\omega_{*}$. Since $\omega_{t}$ tracks $\omega_{*}$, it is easy to verify that $\bar{\mathcal{J}}_{p}(\omega_{t}, z)$ indeed tracks $\mathcal{J}_{p}(z)$ for a given $z \in \mathcal{Z}$.\\

Similarly, in the case of MSBR, we require the following double sampling assumption on the sample trajectory:\vspace*{0mm}\\
\noindent
\begin{itemize}
	\item[$\circledast$] \textbf{\textit{Assumption (A3)$^{\prime}$}: }A sample trajectory $\{(\mathbf{s}_{t}, \mathbf{r}_{t}, \mathbf{r}^{\prime}_{t}, \mathbf{s}^{\prime}_{t}, \mathbf{s}^{\prime\prime}_{t})\}_{t=0}^{\infty}$ is provided, where $\mathbf{s}_{t} \sim \nu(\cdot)$, $\mathbf{s}^{\prime}_{t} \sim \mathrm{P}^{\pi}(\mathbf{s}_{t}, \cdot)$, $\mathbf{s}^{\prime\prime}_{t} \sim \mathrm{P}^{\pi}(\mathbf{s}_{t}, \cdot)$ with $\mathbf{s}^{\prime}_{t}$ and $\mathbf{s}^{\prime\prime}_{t}$ sampled independently. Also, $\mathbf{r}_{t} = \mathrm{R}(\mathbf{s}_{t}, \pi(\mathbf{s}_{t}), \mathbf{s}^{\prime}_{t})$ and $\mathbf{r}^{\prime}_{t} = \mathrm{R}(\mathbf{s}_{t}, \pi(\mathbf{s}_{t}), \mathbf{s}^{\prime\prime}_{t})$. Let $\nu(s)> 0 $, $\forall s \in \mathbb{S}$. Further, let $\phi_t, \phi^{\prime}_{t}, \phi^{\prime\prime}_{t}, \mathbf{r}_{t}$, and $\mathbf{r}^{\prime}_{t}$ have uniformly bounded second moments (where $\phi_t \triangleq \phi(\mathbf{s}_{t}), \phi^{\prime}_{t} \triangleq \phi(\mathbf{s}^{\prime}_{t}), \phi^{\prime\prime}_{t} \triangleq \phi(\mathbf{s}^{\prime\prime}_{t})$).
\end{itemize}

Assumption (A3)$^{\prime}$ does not contain any non-singularity condition. However, it demands the availability of two independent transitions $(\mathbf{s}^{\prime}_{t}, \mathbf{r}_{t}) $ and $(\mathbf{s}^{\prime\prime}_{t}, \mathbf{r}^{\prime}_{t})$ given the current state $\mathbf{s}_{t}$. This requirement is referred to as the \textit{double sampling}.\vspace*{1mm}\\
We maintain the time indexed random vector $\upsilon_{t} \triangleq (\upsilon^{(0)}_{t}, \upsilon^{(1)}_{t}, \upsilon^{(2)}_{t}, \upsilon^{(3)}_{t})^{\top}$, where $\upsilon^{(0)}_{t} \in \bbbr$, $\upsilon^{(1)}_{t} \in \bbbr^{k \times k}$, $\upsilon^{(2)}_{t} \in \bbbr^{k \times 1}$ and $\upsilon^{(3)}_{t} \in \bbbr^{k \times k}$. 
Now the stochastic recursion to track $\upsilon_{*}$ is given by
\begin{equation}\label{eqn:alupsupd}
\upsilon_{t+1} = \upsilon_{t} + \alpha_{t+1}\Delta \upsilon_{t+1}.
\end{equation}
The increment term $\Delta \upsilon_{t+1} \triangleq (\upsilon^{(0)}_{t+1}, \upsilon^{(1)}_{t+1}, \upsilon^{(2)}_{t+1}, \upsilon^{(3)}_{t+1})^{\top}$ used in the above recursion is defined as follows:
\begin{equation}\label{eq:upsiloninc}
\left.
\begin{aligned}
\bigtriangleup{\upsilon}^{(0)}_{t+1}\hspace*{2mm} \triangleq& \hspace*{2mm}\mathbf{r}_{t}\mathbf{r}^{\prime}_{t} - {\upsilon}^{(0)}_{t},\\
\bigtriangleup{\upsilon}^{(1)}_{t+1}\hspace*{2mm} \triangleq& \hspace*{2mm}\gamma^{2}\phi^{\prime}_{t}\phi^{\prime\prime\top}_{t} - {\upsilon}^{(1)}_{t}, \\
\bigtriangleup{\upsilon}^{(2)}_{t+1}\hspace*{2mm} \triangleq& \hspace*{2mm}\mathbf{r}_{t}\big(\phi^{\prime}_{t}-\phi_{t}\big) - \upsilon^{(2)}_{t},\\
\bigtriangleup{\upsilon}^{(3)}_{t+1}\hspace*{2mm} \triangleq& \hspace*{2mm}\big(\phi_t - 2\gamma\phi^{\prime}_{t}\big)\phi_{t}^{\top} - \upsilon^{(3)}_{t}.
\end{aligned}
\hspace*{20mm}\right\}
\end{equation}
We also define the estimate of $\mathcal{J}_{b}(\cdot)$ at time $t$ as follows:\vspace*{2mm}\\
For a given $z \in \mathcal{Z}$,  
\begin{equation}\label{eq:msbrobj-est}
\bar{\mathcal{J}}_{b}(\upsilon_{t}, z) \triangleq -\left(\upsilon^{(0)}_{t} + z^{\top}(\upsilon^{(1)}_{t}+\upsilon^{(3)}_{t})z + 2z^{\top}\upsilon^{(2)}_{t}\right).
\end{equation}
\noindent
\textbf{2. Tracking the ideal CE method: }The ideal CE method defined in Eq. (\ref{eqn:cerecn}) is computationally intractable due to the inherent hardness involved in computing the quantities $\mathbb{E}_{\theta_t}[\cdot]$ and $\gamma_{\rho}(\cdot, \cdot)$ efficiently (hence the tag name ``ideal''). There are multiple ways one can track the ideal CE method. In this paper, we consider the efficient tracking of the ideal CE method using the stochastic approximation (SA) framework proposed in \cite{genstochce2016,ajincedet,predictsce2018}. The stochastic approximation approach is efficient both computationally and storage wise when compared to the rest of the state-of-the-art CE tracking methods.The SA variant is also shown to exhibit global optimum convergence,
\emph{i.e.}, the model sequence $\{\theta_t\}_{t \in \mathbb{N}}$ converges to the degenerate distribution concentrated on any of the global optima of the
objective function. The SA version of the CE method consists of three stochastic recursions which are defined as follows:
\begin{flalign}
\bullet \textbf{ Tracking }&\textbf{$\gamma_{\rho}(\mathcal{J}_{p}, \theta)$:}\hspace*{5mm} \gamma_{t+1} = \gamma_{t} - \beta_{t+1} \Delta \gamma_{t+1}(\mathbf{Z}_{t+1}), \hspace*{3mm}\textrm{ where }\mathbf{Z}_{t+1} \sim f_{\theta}\nonumber \\
&\textrm{ and } \Delta \gamma_{t+1}(x) \triangleq -(1-\rho)\I_{\{\bar{\mathcal{J}}_{p}(\omega_{t}, x) \geq \gamma_t\}}+\rho\I_{\{\bar{\mathcal{J}}_{p}(\omega_{t}, x) \leq \gamma_t\}}.
\end{flalign}
\begin{flalign}
\bullet &\textbf{ Tracking }\textbf{$\Upsilon_{1}(\mathcal{J}_{p}, \theta)$:}\hspace*{5mm}\xi^{(0)}_{t+1} = \xi^{(0)}_{t}+\beta_{t+1}\Delta \xi^{(0)}_{t+1}(\mathbf{Z}_{t+1}), \hspace*{3mm}\textrm{ where }\mathbf{Z}_{t+1} \sim f_{\theta}\nonumber\\
&\hspace*{5mm}\textrm{ and }\Delta \xi^{(0)}_{t+1}(x) \triangleq \mathbf{g_{1}}(\bar{\mathcal{J}}_{p}(\omega_{t}, x), x, \gamma_t) - \xi^{(0)}_t \mathbf{g_{0}}(\bar{\mathcal{J}}_{p}(\omega_{t}, x), \gamma_t).
\end{flalign}
\begin{flalign}
\bullet &\textbf{ Tracking }\textbf{$\Upsilon_{2}(\mathcal{J}_{p}, \theta)$:}\hspace*{5mm}\xi^{(1)}_{t+1} = \xi^{(1)}_{t} + \beta_{t+1} \Delta \xi^{(1)}_{t}(\mathbf{Z}_{t+1}),\hspace*{3mm}\textrm{ where }\mathbf{Z}_{t+1} \sim f_{\theta}\nonumber\\
&\hspace*{5mm}\textrm{ and }\Delta \xi^{(1)}_{t+1}(x) \triangleq \mathbf{g_{2}}(\bar{\mathcal{J}}_{p}(\omega_{t}, x), x, \gamma_t, \xi^{(0)}_t) - \xi^{(1)}_t \mathbf{g_{0}}(\bar{\mathcal{J}}_{p}(\omega_{t}, x), \gamma_t).
\end{flalign}
Note that the above recursions are defined for the objective function $\mathcal{J}_{p}$. However, in the case of $\mathcal{J}_{b}$, the recursions are similar except for $\mathcal{J}_{b}$ replacing $\mathcal{J}_{p}$ and $\upsilon_t$ replacing $\omega_t$ wherever required.\\

\noindent 
\textbf{3. Learning rates and timescales: }Our algorithms use two learning rates $\{\alpha_{t}\}_{t \in \mathbb{N}}$ and $\{\beta_{t}\}_{t \in \mathbb{N}}$, which are deterministic, positive, non-increasing, predetermined (chosen a priori)  and satisfy the following conditions:
\begin{equation}\label{eqn:learnrt}
\sum_{t=1}^{\infty}\alpha_{t} = \sum_{t=1}^{\infty}\beta_{t} = \infty, \hspace{8mm} 
\sum_{t=1}^{\infty}\left(\alpha_{t}^{2}+\beta_{t}^{2}\right) < \infty, \hspace{8mm}
\lim_{t \rightarrow \infty}\frac{\alpha_t}{\beta_t} = 0.
\end{equation}
In a multi-timescale stochastic approximation setting \cite{borkar1997stochastic}, it is important to understand the difference between timescale and learning rate. The timescale of a stochastic recursion is defined by its learning rate (also referred to as step-size). Note that from the conditions imposed on the learning rates $\{\alpha_t\}_{t \in \mathbb{N}}$ and $\{\beta_t\}_{t \in \mathbb{N}}$ in Eq. (\ref{eqn:learnrt}), we have $\frac{\alpha_t}{\beta_t} \rightarrow 0$. So $\alpha_{t}$ decays to $0$ relatively faster than $\beta_{t}$. Hence the timescale obtained from $\{\beta_t\}_{t \in \mathbb{N}}$ is considered faster as compared to the other. So in a multi-timescale stochastic recursion scenario, the evolution of the recursion controlled by $\{\alpha_{t}\}$ (that converges relatively faster to $0$)  is slower compared to the recursions controlled by $\{\beta_{t}\}$. This is because the increments are weighted by their learning rates, \emph{i.e.}, the learning rates control the quantity of change that occurs to the variables when the update is executed. 
\begin{figure}[!h]
	\centering
	\vskip -1mm
	{\includegraphics[scale=0.32]{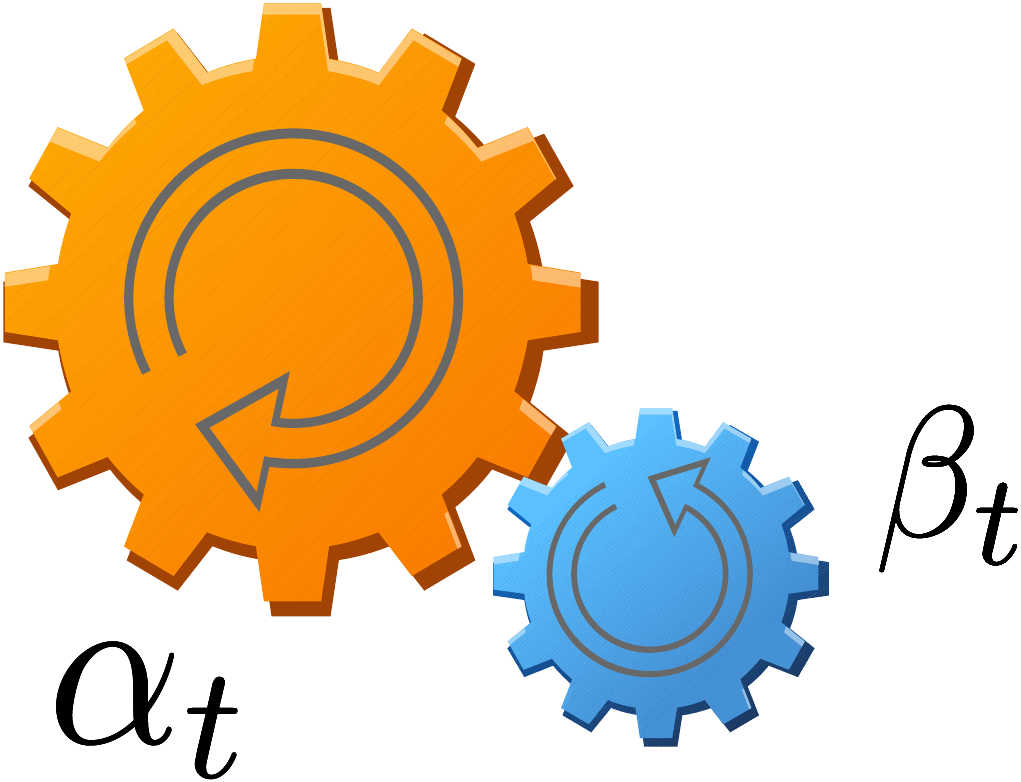}}
	\vskip -1mm
\end{figure}%
When observed from the faster timescale recursion, one can consider the slower timescale recursion to be almost stationary, while when viewed from the slower timescale, the faster timescale recursion appears to have converged. This attribute of the multi-timescale recursions are very important in the analysis of the algorithm. In the analysis, when studying the asymptotic behaviour of a particular stochastic recursion, we can consider the variables of other recursions which are on slower timescales to be constant. In our algorithm, the recursion of $\omega_t$ and $\theta_t$  proceed along the slowest timescale and so updates of $\omega_{t}$ appear to be quasi-static when viewed from the timescale on which the recursions governed by $\beta_{t}$ proceed. The  recursions of $\gamma_t, \xi^{(0)}_t$ and $\xi^{(1)}_t $ proceed along the faster timescale and hence appear equilibrated when viewed from the slower recursion. The coherent behaviour exhibited by the algorithms is primarily attributed to the timescale differences obeyed by the various recursions.\vspace*{2mm}\\

The algorithm SCE-MSPBEM acronym for \textit{stochastic  cross entropy-mean squared projected Bellman error minimization} that minimizes the mean squared projected Bellman error (MSPBE) by incorporating a multi-timescale stochastic approximation variant of the cross entropy (CE) method is formally presented in Algorithm \ref{algo:sce-mspbem}.\vspace*{2mm}\\
\newpage
\noindent
\scalebox{0.95}{
\begin{minipage}{1.05\linewidth}
\begin{algorithm}[H]
	\KwData{ $\alpha_t, \beta_t, c_t \in (0,1)$, $c_t \rightarrow 0$, $\epsilon_1, \lambda, \rho \in (0, 1)$, \hspace*{1mm} $S(\cdot): \bbbr \rightarrow \bbbr_{+}$\\}
	\vspace*{1mm}
	\textbf{Initialization:} $\gamma_0 = 0$, $\gamma^{p}_0 = -\infty$, $\theta_0 = (\mu_0, \Sigma_0)^{\top}$, $T_0 = 0$, $\xi^{(0)}_{t} = 0_{k \times 1}$, $\xi^{(1)}_{t} = 0_{k \times k}$, $\omega^{(0)}_{0} = 0_{k \times 1}, \omega^{(1)}_{0} = 0_{k \times k}, \omega^{(2)}_{0} = 0_{k \times k}, \theta^{p}=NULL$	\vspace*{2mm}\\
	\ForEach{ $(\mathbf{s}_{t}, \mathbf{r}_{t}, \mathbf{s}^{\prime}_{t})$ of the sample trajectory }{
		\vspace*{-4mm}
	\hspace*{70mm}\textit{/*Traj follows (A3)*/}
	\vspace*{2mm}
	{\setlength{\abovedisplayskip}{-2pt}\begin{flalign}\label{eq:dtmix}
	\hspace*{-45mm}\mathbf{Z}_{t+1}   \sim \widehat{f}_{\theta_{t}},\textrm{ where } \widehat{f}_{\theta_{t}} = (1-\lambda)f_{\theta_{t}} + \lambda f_{\theta_0}
	\end{flalign}}
	{\setlength{\abovedisplayskip}{-6pt}\begin{flalign}
	&\hspace*{-2mm}\textbf{Estimate Objective Function $\mathcal{J}_{p}$: }\nonumber\\
	&\hspace*{-2mm}\bm{{\color{Red}\omega_{t+1} = \omega_{t} + \alpha_{t+1}\Delta \omega_{t+1}}}\label{eqn:alomgupd}\\
	&\hspace*{-2mm}\bm{{\color{Red}\bar{\mathcal{J}}_{p}(\omega_{t}, \mathbf{Z}_{t+1}) = -(\omega^{(0)}_{t} + \omega^{(1)}_{t}\mathbf{Z}_{t+1})^{\top}\omega^{(2)}_{t}(\omega^{(0)}_{t} + \omega^{(1)}_{t}\mathbf{Z}_{t+1})}}\label{eqn:jval}
	\end{flalign}}
	{\setlength{\abovedisplayskip}{-5pt}\begin{flalign}\label{eqn:algamma}
	\hspace*{-22mm}\textbf{Track $\gamma_{\rho}(\mathcal{J}_{p}, \widehat{\theta}_{t})$: }\hspace*{12mm}
	\gamma_{t+1} = \gamma_{t} -  \beta_{t+1} \Delta \gamma_{t+1}(\mathbf{Z}_{t+1})\hspace*{0mm}
	\end{flalign}}
{\setlength{\abovedisplayskip}{-8pt}\begin{flalign} \label{eqn:xi0}
	\hspace*{-21mm}\textbf{Track $\Upsilon_{1}(\mathcal{J}_{p}, \widehat{\theta}_t)$: } 
	\hspace*{12mm}\xi^{(0)}_{t+1} =& \xi^{(0)}_{t}+\beta_{t+1}\Delta \xi^{(0)}_{t+1}(\mathbf{Z}_{t+1})
	\end{flalign}}
{\setlength{\abovedisplayskip}{-8pt}\begin{flalign}\label{eqn:xi1}
	\hspace*{-21mm}\textbf{Track $\Upsilon_{2}(\mathcal{J}_{p}, \widehat{\theta}_t)$: } 				
	\hspace*{12mm}\xi^{(1)}_{t+1} =& \xi^{(1)}_{t}+\beta_{t+1}\Delta \xi^{(1)}_{t+1}(\mathbf{Z}_{t+1})
	\end{flalign}}
	\If{$\theta^{p}$ $\neq$ $NULL$}{
	\begin{equation}\label{eqn:algammaold}
	\left.
	\begin{aligned}
	\mathbf{Z}^{p}_{t+1}  &\sim \widehat{f}_{\theta^{p}} \triangleq \lambda f_{\theta_{0}} + (1-\lambda)f_{\theta^{p}}\\
	\gamma^{p}_{t+1} &= \gamma^{p}_{t} -  \beta_{t+1} \Delta \gamma^{p}_{t+1}(\mathbf{Z}^{p}_{t+1})
	\end{aligned}
	\hspace*{20mm}\right\}
	\end{equation}}
	\vspace*{2mm}
	{\setlength{\abovedisplayskip}{0pt}\begin{flalign}\label{eq:Tt}
	\hspace*{-2mm}\textbf{Compare Thresholds: }
	T_{t+1} = T_{t} +  c(\I_{\{\gamma_{t+1} > \gamma^{p}_{t+1}\}} - \I_{\{\gamma_{t+1} \leq \gamma^{p}_{t+1}\}} - T_{t})\hspace{0mm}
	\end{flalign}}
	\eIf{$T_{t+1} > \epsilon_1$}{
	{\setlength{\abovedisplayskip}{0pt}\setlength{\belowdisplayskip}{0pt}\begin{flalign*}
		\hspace*{-32mm}\textbf{Save Old Model: }\gamma^{p}_{t+1} = \gamma_{t}; \hspace*{4mm}\theta^{p} = \theta_{t}
	\end{flalign*}}
	{\setlength{\abovedisplayskip}{-3pt}\setlength{\belowdisplayskip}{0pt}\begin{flalign}\label{eq:thupd}
		\textbf{Update Model: }\theta_{t+1} = \theta_{t} + \alpha_{t+1}\left((\xi^{(0)}_{t}, \xi^{(1)}_{t})^{\top} - \theta_{t}\right) \hspace{2mm}
	\end{flalign}}
	{\setlength{\abovedisplayskip}{-3pt}\setlength{\belowdisplayskip}{0pt}\begin{flalign}\label{eq:gmstarupd}
			\hspace*{-33mm}\textbf{Reset Parameters: } T_{t} = 0; \hspace*{4mm} c = c_{t}
	\end{flalign}}
	}{
		\hspace*{5mm}$\gamma^{p}_{t+1} = \gamma^{p}_{t}$; \hspace{4mm} $\theta_{t+1} = \theta_{t}$\\
	}
	$t := t+1$;	
	}	
	\caption{SCE-MSPBEM\label{algo:sce-mspbem}}
\end{algorithm}
\end{minipage}}\vspace*{2mm}\\
The algorithm SCE-MSBRM acronym for \textit{stochastic  cross entropy-mean squared Bellman residue minimization} that minimizes the mean squared Bellman residue (MSBR) by incorporating a multi-timescale stochastic approximation variant of the cross entropy (CE) method is formally presented in Algorithm \ref{algo:sce-msbr}.\\
\noindent
\scalebox{0.95}{
	\begin{minipage}{1.04\linewidth}
		\begin{algorithm}[H]
			\KwData{ $\alpha_t, \beta_t, c_t \in (0,1)$, $c_t \rightarrow 0$, $\epsilon_1, \lambda, \rho \in (0, 1)$, \hspace*{1mm} $S(\cdot): \bbbr \rightarrow \bbbr_{+}$\\}
			\vspace*{1mm}
			\textbf{Initialization:} $\gamma_0 = 0$, $\gamma^{p}_0 = -\infty$, $\theta_0 = (\mu_0, \Sigma_0)^{\top}$, $T_0 = 0$, $\xi^{(0)}_{t} = 0_{k \times 1}$, $\xi^{(1)}_{t} = 0_{k \times k}$, $\upsilon^{(0)}_{0} = 0, \upsilon^{(1)}_{0} = 0_{k \times k}, \upsilon^{(2)}_{0} = 0_{k \times 1}, \upsilon^{(3)}_{0} = 0_{k \times k}, \theta^{p} = NULL$	\vspace*{2mm}\\
			\ForEach{ $(\mathbf{s}_{t}, \mathbf{r}_{t}, \mathbf{r}^{\prime}_{t}, \mathbf{s}^{\prime}_{t}, \mathbf{s}^{\prime\prime}_{t})$ of the sample trajectory }{
		\vspace*{-4mm}
			\hspace*{78mm}\textit{/*Trajectory follows}\\ \hspace*{83mm}$(A3)^{\prime}$*/
				\vspace*{2mm}
				{\setlength{\abovedisplayskip}{-2pt}\begin{flalign}\label{eq:brdtmix}
					\hspace*{-42mm}\mathbf{Z}_{t+1}   \sim \widehat{f}_{\theta_{t}}, \textrm{ where } \widehat{f}_{\theta_{t}} = (1-\lambda)f_{\theta_{t}} + \lambda f_{\theta_0}
					\end{flalign}}
				{\setlength{\abovedisplayskip}{-6pt}\begin{flalign}
					&\hspace*{-3mm}\textbf{Estimate Objective Function $\mathcal{J}_{b}$: }\nonumber\\
					&\hspace*{-2mm}\bm{{\color{Red}\upsilon_{t+1} = \upsilon_{t} + \alpha_{t+1}\Delta \upsilon_{t+1}}}\label{eqn:bralupsupd}\\
					&\hspace*{-2mm}\bm{{\color{Red}\bar{\mathcal{J}}_{b}(\upsilon_{t}, \mathbf{Z}_{t+1}) = -\left(\upsilon^{(0)}_{t} + \mathbf{Z}_{t+1}^{\top}(\upsilon^{(1)}_{t}+\upsilon^{(3)}_{t})\mathbf{Z}_{t+1} + 2\mathbf{Z}_{t+1}^{\top}\upsilon^{(2)}_{t}\right)}}\label{eqn:brjval}
					\end{flalign}}
				{\setlength{\abovedisplayskip}{-5pt}\begin{flalign}\label{eqn:bralgamma}
					\hspace*{-22mm}\textbf{Track $\gamma_{\rho}(\mathcal{J}_{b}, \widehat{\theta}_{t})$: }\hspace*{12mm}
					\gamma_{t+1} = \gamma_{t} -  \beta_{t+1} \Delta \gamma_{t+1}(\mathbf{Z}_{t+1})\hspace*{0mm}
					\end{flalign}}
				{\setlength{\abovedisplayskip}{-8pt}\begin{flalign} \label{eqn:brxi0}
					\hspace*{-21mm}\textbf{Track $\Upsilon_{1}(\mathcal{J}_{b}, \widehat{\theta}_t)$: } 
					\hspace*{12mm}\xi^{(0)}_{t+1} =& \xi^{(0)}_{t}+\beta_{t+1}\Delta \xi^{(0)}_{t+1}(\mathbf{Z}_{t+1})
					\end{flalign}}
				{\setlength{\abovedisplayskip}{-8pt}\begin{flalign}\label{eqn:brxi1}
					\hspace*{-21mm}\textbf{Track $\Upsilon_{2}(\mathcal{J}_{b}, \widehat{\theta}_t)$: } 				
					\hspace*{12mm}\xi^{(1)}_{t+1} =& \xi^{(1)}_{t}+\beta_{t+1}\Delta \xi^{(1)}_{t+1}(\mathbf{Z}_{t+1})
					\end{flalign}}
				\If{$\theta^{p}$ $\neq$ $NULL$}{
					\begin{equation}\label{eqn:bralgammaold}
					\left.
					\begin{aligned}
					\mathbf{Z}^{p}_{t+1}  &\sim \widehat{f}_{\theta^{p}} \triangleq \lambda f_{\theta_{0}} + (1-\lambda)f_{\theta^{p}}\\
					\gamma^{p}_{t+1} &= \gamma^{p}_{t} -  \beta_{t+1} \Delta \gamma_{t+1}(\mathbf{Z}^{p}_{t+1})
					\end{aligned}
					\hspace*{20mm}\right\}
					\end{equation}}
				\vspace*{2mm}
				{\setlength{\abovedisplayskip}{0pt}\begin{flalign*}
					\hspace*{-2mm}\textbf{Compare Thresholds: }
					T_{t+1} = T_{t} +  c(\I_{\{\gamma_{t+1} > \gamma^{p}_{t+1}\}} - \I_{\{\gamma_{t+1} \leq \gamma^{p}_{t+1}\}} - T_{t})\hspace{1mm}
					\end{flalign*}}
				\eIf{$T_{t+1} > \epsilon_1$}{
					{\setlength{\abovedisplayskip}{0pt}\setlength{\belowdisplayskip}{0pt}\begin{flalign*}
						\hspace*{-32mm}\textbf{Save Old Model: }\gamma^{p}_{t+1} = \gamma_{t}; \hspace*{4mm}\theta^{p} = \theta_{t}
						\end{flalign*}}
					{\setlength{\abovedisplayskip}{-3pt}\setlength{\belowdisplayskip}{0pt}\begin{flalign}\label{eq:brthupd}
						\textbf{Update Model: }\theta_{t+1} = \theta_{t} + \alpha_{t+1}\left((\xi^{(0)}_{t}, \xi^{(1)}_{t})^{\top} - \theta_{t}\right) \hspace{2mm}
						\end{flalign}}
					{\setlength{\abovedisplayskip}{-3pt}\setlength{\belowdisplayskip}{0pt}\begin{flalign}\label{eq:brgmstarupd}
						\hspace*{-33mm}\textbf{Reset Parameters: } T_{t} = 0; \hspace*{4mm} c = c_{t}
						\end{flalign}}
				}{
					\hspace*{5mm}$\gamma^{p}_{t+1} = \gamma^{p}_{t}$; \hspace{4mm} $\theta_{t+1} = \theta_{t}$\\
				}
				$t := t+1$;	
			}			\caption{SCE-MSBRM\label{algo:sce-msbr}}
		\end{algorithm}
\end{minipage}}

\section{Convergence Analysis}
Observe that the algorithms are multi-timescale stochastic approximation algorithms \cite{borkar1997stochastic} involving multiple stochastic recursions piggybacking each other. The primal recursions which typify the algorithms are the stochastic recursions which update the model parameters $\theta_t$ (Eq. (\ref{eq:thupd}) of Algorithm \ref{algo:sce-mspbem} and Eq. (\ref{eq:brthupd}) of Algorithm \ref{algo:sce-msbr}), where the model parameters $\theta_t$ are calibrated to ensure their evolution towards the degenerate distribution concentrated on the global optimum ($z^{*}_p$ for Algorithm \ref{algo:sce-mspbem} and $z_{b}^{*}$ for Algorithm \ref{algo:sce-msbr}). Nonetheless, not disregarding the relevance of the remaining recursions which are all too vital and should augment each other and the primal recursion in achieving the desideratum. Therefore to analyze the limiting behaviour of the algorithms, one has to study the asymptotic behaviour of the individual recursions, \emph{i.e.}, the effectiveness of the variables involved in tracking the true quantities. For analyzing the asymptotic behaviour of the algorithms, we apply the  ODE based analysis from \cite{ljung1977analysis,kushner2012stochastic,kubrusly1973stochastic,borkar2008stochastic,benveniste2012adaptive}. In this method of analysis, for each individual stochastic recursion, we identify an associated ODE whose asymptotic (limiting) behaviour is similar to that of the stochastic recursion. In other words, the stochastic recursion eventually tracks the associated ODE. Subsequently, a qualitative analysis of the associated ODE is performed to study its limiting behaviour and it is argued that the stochastic recursion asymptotically converges almost surely to the set of stable fixed points of the ODE (See Chapter 2 of \cite{borkar2008stochastic} or Chapter 5 of \cite{kushner2012stochastic} or Chapter 2 of \cite{benveniste2012adaptive}).

\subsection{Outline of the Proof}
The roadmap followed in the analysis of the algorithms is as follows:
\begin{enumerate}[itemsep=2mm,leftmargin=4mm]
\item
	First and foremost, in the case of Algorithm \ref{algo:sce-mspbem}, we study the asymptotic behaviour of the stochastic recursion (\ref{eqn:alomgupd}). We show in Lemma \ref{lemma:lm1} that the stochastic sequence $\{\omega_{t}\}$ indeed tracks the true quantity $\omega_{*}$ which defines the true objective function $\mathcal{J}_{p}$. Note that the recursion (\ref{eqn:alomgupd}) is independent of other recursions and hence can be analyzed independently. The composition of the analysis (proof of Lemma \ref{lemma:lm1}) apropos of the limiting behaviour of $\{\omega_t\}$ involves mutlitple steps such as analyzing the nature of growth of the stochastic sequence, identifying the character of the implicit noise extant in the stochastic recursion, exploring the existence of finite bounds of the noise sequence (we solicit probabilistic analysis \cite{borkar2012probability} to realize the above steps), ensuring with certainty the stability of the stochastic sequence (we appeal to Borkar-Meyn theorem \cite{borkar2008stochastic}) and finally the qualitative analysis of the limit points of the associated ODE of the stochastic recursion (we seek assistance from dynamical systems theory \cite{perko2013differential}). 
\item
Similarly, in the case of Algorithm \ref{algo:sce-msbr}, we study the asymptotic behaviour of the stochastic recursion (\ref{eqn:bralupsupd}). We show in Lemma \ref{lemma:lm2} that the stochastic sequence $\{\upsilon_{t}\}$ certainly tracks the true quantity $\upsilon_{*}$ which defines the true objective function $\mathcal{J}_{b}$. The composition of the proof of Lemma \ref{lemma:lm2} follows similar discourse as that of Lemma \ref{lemma:lm1}. 
\item
	Since the proposed algorithms are multi-timescale stochastic approximation algorithms, their asymptotic behaviour depends heavily on the timescale differences induced by the step-size schedules $\{\alpha_{t}\}_{t \in \N}$ and $\{\beta_{t}\}_{t \in \N}$. The timescale differences allow the different individual recursions in a multi-timescale setting to learn at different rates. Since $\frac{\alpha_t}{\beta_t} \rightarrow 0$, the step-size $\{\beta_t\}_{t \in \N}$ decays to $0$ at a relatively slower rate than $\{\alpha_t\}_{t \in \N}$ and therefore the increments in the recursions (\ref{eqn:algamma})-(\ref{eqn:xi1})  which are controlled by $\beta_t$ are relatively larger and hence appear to converge relatively faster than the recursions (\ref{eqn:alomgupd})-(\ref{eqn:jval}) and (\ref{eq:thupd}) which are controlled by $\alpha_t$ when viewed from the latter. So, considering a finite, yet sufficiently long time window, the relative evolution of  the variables from the slower timescale $\alpha_{t}$, \emph{i.e.}, $\omega_{t}$ and $\theta_t$ to their steady-state form is indeed slow and in fact can be considered quasi-stationary when viewed from the evolutionary path of the faster timescale $\beta_t$. See Chapter 6 of \cite{borkar2008stochastic} for a succinct description on multi-timescale stochastic approximation algorithms. Hence, when viewed from the timescale of the recursions (\ref{eqn:algamma})-(\ref{eqn:xi1}), one may consider $\omega_{t}$ and $\theta_{t}$ to be fixed. This is a standard technique used in analyzing multi-timescale stochastic approximation algorithms. Following this course of analysis, we obtain Lemma \ref{lmn:xiconv} which characterizes the asymptotic behaviour of the stochastic recursions (\ref{eqn:algamma})-(\ref{eqn:xi1}). The original paper \cite{predictsce2018} apropos of the stochastic approximation version of the CE method (proposed for a generalized optimization setting) establishes claims  synonymous to Lemma \ref{lmn:xiconv} and hence we skip the proof of the lemma, nonetheless, we provide references to the same.\\

The results in Lemma \ref{lmn:xiconv} attest to validate that under the quasi-stationary hypothesis of $\omega_t \equiv \omega$ and $\theta_t \equiv \theta$, the stochastic sequence $\{\gamma_t\}$ tracks the true quantile $\gamma_{\rho}(\bar{\mathcal{J}}_{p}(\omega, \cdot),\widehat{\theta})$ ((1) of Lemma \ref{lmn:xiconv}), while the stochastic sequences $\{\xi^{(0)}_{t}\}$ and $\{\xi^{(1)}_{t}\}$ track the ideal CE model parameters $\Upsilon_1(\bar{\mathcal{J}}_{p}(\omega, \cdot), \widehat{\theta})$ and $\Upsilon_2(\bar{\mathcal{J}}_{p}(\omega, \cdot), \widehat{\theta})$ respectively ((2-3) of Lemma \ref{lmn:xiconv}) with probability one. Certainly, these results establish that the stochastic recursions (\ref{eqn:algamma}-\ref{eqn:xi1}) track the ideal CE method and ergo, they provide a stable and proximal optimization gadget to minimize the error functions MSPBE (or MSBR). The rationale behind the pertinence of the stochastic recursion (\ref{eq:Tt}) is provided in \cite{predictsce2018}. Ostensibly, the purpose is as follows: The threshold sequence $\{\gamma_{\rho}(\mathcal{J}_{p}, \theta_t)\}$ (where $\theta_t$ is generated by Eq. (\ref{eqn:cerecn})) of the ideal CE method is monotonically increasing (Proposition 2 of \cite{predictsce2018}). However, when stochastic approximation iterates are employed to track the ideal model parameters, the monotonicity may not hold always. The purpose of the stochastic recursion (\ref{eq:Tt}) is to ensure that the monotonicity of the threshold sequence is maintained and therefore (4-5) of Lemma \ref{lmn:xiconv} along with an appropriate choice of $\epsilon_1 \in [0,1)$ (Algorithm \ref{algo:sce-mspbem}) ensure that the model sequence $\{\theta_t\}$ is updated infinitely often.
\item
	Finally, we state our main results regarding the convergence of MSPBE and MSBR in Theorems \ref{thm:main} and \ref{thm:main2}, respectively. The theorems analyze the asymptotic behaviour of the model sequence $\{\theta_t\}_{t \in \mathbb{N}}$ for Algorithms \ref{algo:sce-mspbem} and \ref{algo:sce-msbr} respectively. The theorems claim that the model sequence $\{\theta_{t}\}$ generated by Algorithm \ref{algo:sce-mspbem} (Algorithm \ref{algo:sce-msbr}) almost surely converges to $\theta_{p}^{*} = (z_{p}^{*}, 0_{k \times k})^{\top}$ $(\theta_{b}^{*} = (z_{b}^{*}, 0_{k \times k})^{\top})$, the degenerate distribution concentrated at $z_{p}^{*}$ $(z_{b}^{*})$, where $z_{p}^{*}$ $(z_{b}^{*})$ is the solution to the optimization problem (\ref{eqn:mspbeobj}) ((\ref{eqn:msbrobj})) which minimizes the error function MSPBE (MSBR).
\end{enumerate}

\subsection{The Proof of Convergence}
\noindent
For the stochastic recursion (\ref{eqn:alomgupd}), we have the following result:\\
As a proviso, we define the filtration\footnote{For detailed technical information pertaining to filtration and $\sigma$-field, refer \cite{borkar2012probability}.} $\{\mathcal{F}_{t}\}_{t \in \mathbb{N}}$, where the $\sigma$-field \\$\mathcal{F}_t \triangleq \sigma\left(\omega_i, \gamma_i, \gamma^{p}_i, \xi^{(0)}_i, \xi^{(1)}_i, \theta_i, 0 \leq i \leq t; \mathbf{Z}_{i}, 1 \leq i \leq t; \mathbf{s}_{i}, \mathbf{r}_{i}, \mathbf{s}^{\prime}_{i}, 0 \leq i < t \right)$, $t \in \mathbb{N}$, is the $\sigma$-field generated by the specified random variables in the definition.
\noindent
\begin{lemma}\label{lemma:lm1}
Let the step-size sequences $\{\alpha_{t}\}_{t \in \mathbb{N}}$ and $\{\beta_{t}\}_{t \in \mathbb{N}}$ satisfy Eq. (\ref{eqn:learnrt}). For the sample trajectory $\{(\mathbf{s}_{t}, \mathbf{r}_{t}, \mathbf{s}^{\prime}_{t})\}_{t=0}^{\infty}$, we let Assumption (A3) hold. Then, for a given $z \in \mathcal{Z}$, the sequence $\{\omega_{t}\}_{t \in \mathbb{N}}$ defined in Eq. (\ref{eqn:alomgupd}) satisfies with probability one,
\begin{flalign*}
&\lim_{t \rightarrow \infty}\omega^{(0)}_{t} = \omega^{(0)}_{*}, \hspace*{4mm} \lim_{t \rightarrow \infty}\omega^{(1)}_{t} = \omega^{(1)}_{*},\\
&\lim_{t \rightarrow \infty}\omega^{(2)}_{t} = \omega^{(2)}_{*} \hspace{2mm} \mathrm{and} \hspace{2mm} \lim_{t \rightarrow \infty} \bar{\mathcal{J}}_{p}(\omega_{t}, z) = -MSPBE(z), \hspace{15mm}
\end{flalign*}
where $\omega^{(0)}_{*}$, $\omega^{(1)}_{*}$, $\omega^{(2)}_{*}$ and MSPBE are defined in Eq. (\ref{eq:stdetobj}), while $\bar{\mathcal{J}}_{p}(\omega_t,z)$ is defined in Eq. (\ref{eqn:jval}).
\end{lemma}
\begin{Proof}
By rearranging equations in (\ref{eqn:alomgupd}), for $t \in \mathbb{N}$, we get
\begin{equation}\label{eqn:omgrec1}
\omega^{(0)}_{t+1} =  \omega^{(0)}_{t} + \alpha_{t+1}\big(\mathbb{M}^{(0,0)}_{t+1}+h^{(0,0)}(\omega^{(0)}_{t})\big), \hspace*{5mm}
\end{equation}
where $\mathbb{M}^{(0,0)}_{t+1} = \mathbf{r}_{t}\phi_{t} - \mathbb{E}\left[\mathbf{r}_{t}\phi_{t}\right] \hspace{1mm} \mathrm{and}  \hspace{1mm} h^{(0,0)}(x)=\mathbb{E}\left[\mathbf{r}_{t}\phi_{t} \right]-x$.\\
Similarly,
\begin{equation}\label{eqn:omgrec2}
\omega^{(1)}_{t+1} =  \omega^{(1)}_{t} + \alpha_{t+1}\big(\mathbb{M}^{(0,1)}_{t+1}+h^{(0,1)}(\omega^{(1)}_{t})\big), \hspace*{5mm}
\end{equation}
where $\mathbb{M}^{(0,1)}_{t+1} = \phi_{t}(\gamma\phi^{\prime}_{t}-\phi_{t})^{\top} - \mathbb{E}\left[\phi_{t}(\gamma\phi^{\prime}_{t}-\phi_{t})^{\top}\right]$ and \vspace*{3mm}\\
$h^{(0,1)}(x)=\mathbb{E}\left[\phi_{t}(\gamma\phi^{\prime}_{t}-\phi_{t})^{\top}\right]-x$.\vspace*{2mm}\\
Finally,
\begin{equation}\label{eqn:omgrec3}
\omega^{(2)}_{t+1} =  \omega^{(2)}_{t} + \alpha_{t+1}\big(\mathbb{M}^{(0,2)}_{t+1}+h^{(0,2)}(\omega^{(2)}_{t})\big),
\end{equation}
where $\mathbb{M}^{(0,2)}_{t+1} = \mathbb{E}\left[\phi_{t}\phi_{t}^{\top}\omega^{(2)}_{t} \right] - \phi_{t}\phi_{t}^{\top}\omega^{(2)}_{t}  \textrm{ and }
h^{(0,2)}(x) = \mathbb{I}_{k \times k} -\mathbb{E}\left[\phi_{t}\phi_{t}^{\top}x\right]$.\vspace*{4mm}\\
To apply the ODE based analysis, certain necessary conditions on the structural decorum are in order:\\
\begin{enumerate}
\item[(B1)]
$h^{(0,j)}, 0 \leq j \leq 2$ are Lipschitz continuous (easy to verify).
\item[(B2)]
$\{\mathbb{M}^{(0,j)}_{t+1}\}_{t \in \mathbb{N}}$, $0 \leq j \leq 2$ are martingale difference noise sequences, \emph{i.e.}, for each $j$, $\mathbb{M}^{(0,j)}_{t}$ is $\mathcal{F}_{t}$-measurable, integrable and $\mathbb{E}\left[\mathbb{M}^{(0,j)}_{t+1} \vert \mathcal{F}_{t}\right] = 0$, $t \in \mathbb{N}$, $0 \leq j \leq 2$.
\item[(B3)]
Since $\phi_{t}$, $\phi^{\prime}_{t}$ and $\mathbf{r}_{t}$ have uniformly bounded second moments, the noise sequences $\{\mathbb{M}^{(0,j)}_{t+1}\}_{t \in \mathbb{N}}$ have uniformly bounded second moments as well for each $0 \leq j \leq 2$ and hence $\exists K_{0,0}, K_{0,1}, K_{0,2} > 0$ \emph{s.t.}
\begin{flalign}
&\mathbb{E}\left[\Vert \mathbb{M}^{(0,0)}_{t+1} \Vert^{2} \big\vert \mathcal{F}_{t}\right] \leq K_{0,0}(1+\Vert \omega^{(0)}_{t} \Vert^{2}),\hspace*{3mm} t \in \mathbb{N}.\\
&\mathbb{E}\left[\Vert \mathbb{M}^{(0,1)}_{t+1} \Vert^{2} \big\vert \mathcal{F}_{t}\right] \leq K_{0,1}(1+\Vert \omega^{(1)}_{t} \Vert^{2}),\hspace*{3mm} t \in \mathbb{N}.\\
&\mathbb{E}\left[\Vert \mathbb{M}^{(0,2)}_{t+1} \Vert^{2} \big\vert \mathcal{F}_{t}\right] \leq K_{0,2}(1+\Vert \omega^{(2)}_{t} \Vert^{2}),\hspace*{3mm} t \in \mathbb{N}.
\end{flalign}
\item[(B4)]
To establish the stability (boundedness) condition, \emph{i.e.}, $\sup_{t \in \mathbb{N}} \Vert \omega^{(j)}_t \Vert < \infty$ $a.s.$, for each $0 \leq j \leq 2$, we appeal to the Borkar-Meyn theorem (Theorem 2.1 of \cite{borkar2000ode} or Theorem 7, Chapter 3 of \cite{borkar2008stochastic}). Particularly, in order to prove $\sup_{t \in \mathbb{N}} \Vert \omega^{(0)}_{t} \Vert < \infty$  $a.s.$, we study the qualitative behaviour of the dynamical system defined by the following  limiting ODE:
\begin{flalign}\label{eqn:omg-scode0}
\frac{d}{dt}\omega^{(0)}(t) = h^{(0,0)}_{\infty}(\omega^{(0)}(t)), \hspace*{2mm} t \in \bbbr_{+},
\end{flalign}
where
\begin{flalign*}
h^{(0,0)}_{\infty}(x) \triangleq \lim_{c \rightarrow \infty}\frac{h^{(0,0)}(cx)}{c} = \lim_{c \rightarrow \infty}\frac{\mathbb{E}\left[\mathbf{r}_{t}\phi_{t} \right]-cx}{c} = \lim_{c \rightarrow \infty}\frac{\mathbb{E}\left[\mathbf{r}_{t}\phi_{t}\right]}{c} - x = -x.
\end{flalign*}
According to the Borkar-Meyn theorem, the global asymptotic stability of the above limiting system to the origin is sufficient to warrant the stability of the sequence $\{\omega^{(0)}_{t}\}_{t \in \mathbb{N}}$. Now, note that the ODE (\ref{eqn:omg-scode0}) is a linear, first-order ODE with negative rate of change and hence qualitatively the flow induced by the ODE is globally asymptotically stable to the origin. Therefore, we obtain the following:
\begin{flalign}
\sup_{t \in \mathbb{N}}{\Vert \omega^{(0)}_{t} \Vert} < \infty \hspace*{2mm}a.s.
\end{flalign}
Similarly we can show that 
\begin{flalign}
\sup_{t \in \mathbb{N}}{\Vert \omega^{(1)}_{t} \Vert} < \infty \hspace*{2mm} \emph{a.s.}
\end{flalign}
Now, regarding the stability of the sequence $\{\omega^{(2)}_{t}\}_{t \in \mathbb{N}}$, we consider the following limiting ODE:
\begin{flalign}\label{eqn:omg-sode2}
\frac{d}{dt}\omega^{(2)}(t) = h^{(0,2)}_{\infty}(\omega^{(2)}(t) ), \hspace*{2mm} t \in \bbbr_{+},
\end{flalign}
where
\begin{flalign*}
h^{(0,2)}_{\infty}(x) \triangleq \lim_{c \rightarrow \infty}\frac{h^{(0,2)}(cx)}{c} &= \lim_{c \rightarrow \infty}\frac{\mathbb{I}_{k \times k}-\mathbb{E}\left[\phi_{t}\phi_{t}^{\top}cx\right]}{c} \\ &= \lim_{c \rightarrow \infty} \frac{\mathbb{I}_{k \times k}}{c} - x\mathbb{E}\left[\phi_{t}\phi_{t}^{\top} \right] = - x\mathbb{E}\left[\phi_{t}\phi_{t}^{\top} \right].
\end{flalign*}

The system defined by the limiting ODE (\ref{eqn:omg-sode2}) is globally asymptotically stable to the origin since $\mathbb{E}[\phi_t \phi_t^{\top}]$ is positive definite (as it is positive semi-definite (easy to verify) and non-singular (from Assumption (A3))).  Therefore, by Borkar-Meyn theorem, we obtain the following:
\begin{flalign}
\sup_{t \in \mathbb{N}}{\Vert \omega^{(2)}_{t} \Vert} < \infty \hspace*{2mm} \emph{a.s.} 
\end{flalign}
\end{enumerate}
Since we have hitherto established the necessary conditions (B1-B4), now by appealing to Theorem 2, Chapter 2 of \cite{borkar2008stochastic}, we can directly establish the asymptotic equivalence between the individual stochastic recursions (\ref{eqn:omgrec1})-(\ref{eqn:omgrec3}) and the following associated ODEs respectively.
\begin{flalign}
\frac{d}{dt}\omega^{(0)}(t) = \mathbb{E}\left[\mathbf{r}_{t}\phi_{t}\right]-\omega^{(0)}(t), \hspace*{6mm} t \in \bbbr_{+}, \label{eqn:omgode1} \\
\frac{d}{dt}\omega^{(1)}(t) = \mathbb{E}\left[\phi_{t}(\gamma\phi^{\prime}_{t}-\phi_{t})^{\top}\right]-\omega^{(1)}(t)), \hspace*{6mm} t \in \bbbr_{+}, \label{eqn:omgode2} \\
\frac{d}{dt}\omega^{(2)}(t) = \mathbb{I}_{k \times k} -\mathbb{E}\left[\phi_{t}\phi_{t}^{\top}\right]\omega^{(2)}(t), \hspace*{6mm} t \in \bbbr_{+}. \label{eqn:omgode3}
\end{flalign}
Now we study the qualitative behaviour of the above system of first-order, linear ODEs. A simple examination of the trajectories of the ODEs reveals that the point $\mathbb{E}\left[\mathbf{r}_{t}\phi_{t}\right]$ is a globally asymptotically stable equilibrium point of the ODE (\ref{eqn:omgode1}). Similarly for the ODE (\ref{eqn:omgode2}), the point $\mathbb{E}\left[\phi_{t}(\gamma\phi^{\prime}_{t}-\phi_{t})^{\top}\right]$ is a globally asymptotically stable equilibrium. Finally, regarding the limiting behaviour of the ODE (\ref{eqn:omgode3}), we find that the point $\mathbb{E}\left[\phi_{t}\phi_{t}^{\top}\right]^{-1}$  is a globally asymptotically stable equilibrium. This follows since $\mathbb{E}\left[\phi_{t}\phi_{t}^{\top}\right]$ is positive semi-definite (easy to verify) and non-singular (from Assumption (A3)). Formally, 
\begin{flalign}
&\lim_{t \rightarrow \infty} \omega^{(0)}(t) =  \mathbb{E}\left[\mathbf{r}_{t}\phi_{t}\right],\\
&\lim_{t \rightarrow \infty} \omega^{(1)}(t) =  \mathbb{E}\left[\phi_{t}(\gamma\phi^{\prime}_{t}-\phi_{t})^{\top}\right],\\
&\lim_{t \rightarrow \infty} \omega^{(2)}(t) =  \mathbb{E}\left[\phi_{t}\phi_{t}^{\top}\right]^{-1},
\end{flalign}
where the above convergence is achieved independent of the initial values $\omega^{(0)}(0), \omega^{(1)}(0)$ and $\omega^{(2)}(0)$.

Therefore, by the employing the asymptotic equivalence of the stochastic recursions (\ref{eqn:omgrec1})-(\ref{eqn:omgrec3}) and their associated ODEs (\ref{eqn:omgode1})-(\ref{eqn:omgode3}) we obtain the following:
\begin{flalign*}
&\lim_{t \rightarrow \infty}\omega^{(0)}_{t} = \lim_{t \rightarrow \infty} \omega^{(0)}(t) = \mathbb{E}\left[\mathbf{r}_{t}\phi_{t}\right]  \textit{ a.s. } = \omega^{(0)}_{*}.\\ 
&\lim_{t \rightarrow \infty}\omega^{(1)}_{t} = \lim_{t \rightarrow \infty} \omega^{(1)}(t) = \mathbb{E}\left[\phi_t(\gamma\phi^{\prime}_{t} -\phi_t)^{\top}\right] \textit{ a.s. } = \omega^{(1)}_{*}. \\ 
&\lim_{t \rightarrow \infty}\omega^{(2)}_{t} = \lim_{t \rightarrow \infty} \omega^{(2)}(t) = \mathbb{E}\left[\phi_{t}\phi_{t}^{\top}\right]^{-1} \textit{ a.s. } = \omega^{(2)}_{*}.
\end{flalign*}

Putting all the above together, we get, for $z \in \mathcal{Z}$,
$\lim_{t \rightarrow \infty} \bar{\mathcal{J}}_{p}(\omega_t, z) = \bar{\mathcal{J}}_{p}(\omega_{*}, z)$  = $\mathcal{J}_{p}(z)$ \emph{a.s.}
\end{Proof}\\

\noindent
For the stochastic recursion (\ref{eqn:bralupsupd}), we have the following result:\\
Again, as a proviso, we define the filtration $\{\mathcal{F}_{t}\}_{t \in \mathbb{N}}$ where the $\sigma$-field \\$\mathcal{F}_t \triangleq \sigma\left(\upsilon_i, \gamma_i, \gamma^{p}_i, \xi^{(0)}_i, \xi^{(1)}_i, \theta_i, 0 \leq i \leq t; \mathbf{Z}_{i}, 1 \leq i \leq t; \mathbf{s}_{i}, \mathbf{r}_{i}, \mathbf{r}^{\prime}_{i}, \mathbf{s}^{\prime}_{i}, \mathbf{s}^{\prime\prime}_{i}, 0 \leq i < t \right)$, $t \in \mathbb{N}$.
\begin{lemma}\label{lemma:lm2}
	Let the step-size sequences $\{\alpha_{t}\}_{t \in \mathbb{N}}$ and $\{\beta_{t}\}_{t \in \mathbb{N}}$ satisfy Eq. (\ref{eqn:learnrt}). For the sample trajectory $\{(\mathbf{s}_{t}, \mathbf{r}_{t}, \mathbf{r}^{\prime}_{t}, \mathbf{s}^{\prime}_{t}, \mathbf{s}^{\prime\prime}_{t})\}_{t=0}^{\infty}$, we let Assumption $(A3)^{\prime}$ hold. Then, for a given $z \in \mathcal{Z}$, the sequence $\{\upsilon_{t}\}_{t \in \mathbb{N}}$ defined in  Eq. (\ref{eqn:bralupsupd}) satisfies with probability one,
	\begin{flalign*}
	&\lim_{t \rightarrow \infty}\upsilon^{(0)}_{t} = \upsilon^{(0)}_{*}, \hspace*{2mm} \lim_{t \rightarrow \infty} \upsilon^{(1)}_{t} = \upsilon^{(1)}_{*},
	\hspace*{2mm} \lim_{t \rightarrow \infty}\upsilon^{(2)}_{t} = \upsilon^{(2)}_{*}, \\ &\lim_{t \rightarrow \infty} \upsilon^{(3)}_{t} = \upsilon^{(3)}_{*}
	\hspace*{2mm}\mathrm{and} \hspace{2mm} \lim_{t \rightarrow \infty} \bar{\mathcal{J}}_{b}(\upsilon_{t}, z) = -MSBR(z), \hspace{15mm}
	\end{flalign*}
	where $\upsilon^{(0)}_{*}$, $\upsilon^{(1)}_{*}$, $\upsilon^{(2)}_{*}$, $\upsilon^{(3)}_{*}$ and MSBR are defined in Eq. (\ref{eqn:msbrclform}), while $\bar{\mathcal{J}}_{b}(\upsilon_t,z)$ is defined in Eq. (\ref{eqn:brjval}).
\end{lemma}
\begin{Proof}
	By rearranging equations in (\ref{eqn:bralupsupd}), for $t \in \mathbb{N}$, we get
	\begin{equation}\label{eqn:upsrec0}
	\upsilon^{(0)}_{t+1} =  \upsilon^{(0)}_{t} + \alpha_{t+1}\big(\mathbb{M}^{(1,0)}_{t+1}+h^{(1,0)}(\upsilon^{(0)}_{t})\big), \hspace*{5mm}
	\end{equation}
	where $\mathbb{M}^{(1,0)}_{t+1} = \mathbf{r}_{t}\mathbf{r}^{\prime}_{t} - \mathbb{E}^{2}\left[\mathbf{r}_{t}\right] \hspace{1mm} \mathrm{and}  \hspace{1mm} h^{(1,0)}(x)=\mathbb{E}^{2}\left[\mathbf{r}_{t}\right]-x$.\vspace*{4mm}\\
	Similarly,
	\begin{equation}\label{eqn:upsrec1}
	\upsilon^{(1)}_{t+1} =  \upsilon^{(1)}_{t} + \alpha_{t+1}\big(\mathbb{M}^{(1,1)}_{t+1}+h^{(1,1)}(\upsilon^{(1)}_{t})\big), \hspace*{5mm}
	\end{equation}
	where $\mathbb{M}^{(1,1)}_{t+1} = \gamma^{2}\phi^{\prime}_{t}\phi^{\prime\prime\top}_{t} - \gamma^{2}\mathbb{E}\left[\phi^{\prime}_{t}\right]\mathbb{E}\left[\phi^{\prime}_{t}\right]^\top$
	and 
	$h^{(1,1)}(x) = \gamma^{2}\mathbb{E}\left[\phi^{\prime}_{t}\right]\mathbb{E}\left[\phi^{\prime}_{t}\right]^\top-x$.\vspace*{4mm}\\
	Also,
	\begin{equation}\label{eqn:upsrec2}
	\upsilon^{(2)}_{t+1} =  \upsilon^{(2)}_{t} + \alpha_{t+1}\big(\mathbb{M}^{(1,2)}_{t+1}+h^{(1,2)}(\upsilon^{(2)}_{t})\big),
	\end{equation}
	where $\mathbb{M}^{(1,2)}_{t+1} = \mathbf{r}_{t}\big(\phi^{\prime}_{t}-\phi_{t}\big) 
	- \mathbb{E}\left[\mathbf{r}_{t}\big(\phi^{\prime}_{t}-\phi_{t}\big)\right]  \textrm{ and }
	h^{(1,2)}(x) = \mathbb{E}\left[\mathbf{r}_{t}\big(\phi^{\prime}_{t}-\phi_{t}\big)\right] - x$.\vspace*{2mm}\\
	Finally,
	\begin{equation}\label{eqn:upsrec3}
	\upsilon^{(3)}_{t+1} =  \upsilon^{(3)}_{t} + \alpha_{t+1}\big(\mathbb{M}^{(1,3)}_{t+1}+h^{(1,3)}(\upsilon^{(3)}_{t})\big),
	\end{equation}
	where $\mathbb{M}^{(1,3)}_{t+1} = \big(\phi_{t} - 2\gamma\phi^{\prime}_{t}\big)\phi_{t}^{\top}  
	- \mathbb{E}\left[\big(\phi_{t} - 2\gamma\phi^{\prime}_{t}\big)\phi_{t}^{\top}\right]$ and \\$
	h^{(1,3)}(x) = \mathbb{E}\left[\big(\phi_{t} - 2\gamma\phi^{\prime}_{t}\big)\phi_{t}^{\top}\right] - x$.\vspace*{4mm}\\	
	To apply the ODE based analysis, certain necessary conditions on the structural decorum are in order:\\
	\begin{enumerate}
	\item[(C1)]
	$h^{(1,j)}, 0 \leq j \leq 3$ are Lipschitz continuous (easy to verify).\vspace*{1mm}
	\item[(C2)]
	$\{\mathbb{M}^{(1,j)}_{t+1}\}_{t \in \mathbb{N}}$, $0 \leq j \leq 3$ are martingale difference noise sequences, \emph{i.e.}, for each $t \in \mathbb{N}$, $\mathbb{M}^{(1,j)}_{t}$ is $\mathcal{F}_{t}$-measurable, integrable and $\mathbb{E}\left[\mathbb{M}^{(1,j)}_{t+1} \vert \mathcal{F}_t\right] = 0$, $t \in \mathbb{N}$, $0 \leq j \leq 3$.\vspace*{1mm}
	\item[(C3)]
	Since $\phi_{t}$, $\phi^{\prime}_{t}$, $\phi^{\prime\prime}_{t}$, $\mathbf{r}_{t}$ and $\mathbf{r}^{\prime}_{t}$ have uniformly bounded second moments, the noise sequences $\{\mathbb{M}^{(1,j)}_{t+1}\}_{t \in \mathbb{N}}$, $0 \leq j \leq 3$ have uniformly bounded second moments as well and hence $\exists K_{1,0},K_{1,1},K_{1,2},K_{1,3} > 0$ \emph{s.t.}
	\begin{flalign}
	&\mathbb{E}\left[\Vert \mathbb{M}^{(1,0)}_{t+1} \Vert^{2} \vert \mathcal{F}_{t}\right] \leq K_{1,0}(1+\Vert \upsilon^{(0)}_{t} \Vert^{2}), t \in \mathbb{N}.\\
	&\mathbb{E}\left[\Vert \mathbb{M}^{(1,1)}_{t+1} \Vert^{2} \vert \mathcal{F}_{t}\right] \leq K_{1,1}(1+\Vert \upsilon^{(1)}_{t} \Vert^{2}), t \in \mathbb{N}.\\
	&\mathbb{E}\left[\Vert \mathbb{M}^{(1,2)}_{t+1} \Vert^{2} \vert \mathcal{F}_{t}\right] \leq K_{1,2}(1+\Vert \upsilon^{(2)}_{t} \Vert^{2}), t \in \mathbb{N}.\\
	&\mathbb{E}\left[\Vert \mathbb{M}^{(1,3)}_{t+1} \Vert^{2} \vert \mathcal{F}_{t}\right] \leq K_{1,3}(1+\Vert \upsilon^{(3)}_{t} \Vert^{2}), t \in \mathbb{N}.	
	\end{flalign}
	\item[(C4)]
	To establish the stability condition, \emph{i.e.}, $\sup_{t \in \mathbb{N}} \Vert \upsilon^{(j)}_t \Vert < \infty$ $a.s.$, for each $0 \leq j \leq 3$, we appeal to the Borkar-Meyn theorem (Theorem 2.1 of \cite{borkar2000ode} or Theorem 7, Chapter 3 of \cite{borkar2008stochastic}). Indeed to prove $\sup_{t \in \mathbb{N}} \Vert \upsilon^{(0)}_{t} \Vert < \infty$  $a.s.$, we consider 
	the dynamical system defined by the following $\infty$-system ODE:
	\begin{flalign}\label{eq:upsode0}
	\frac{d}{dt}{\upsilon^{(0)}}(t) = h^{(1,0)}_{\infty}(\upsilon^{(0)}(t)),
	\end{flalign}
	 where 	
	 \begin{flalign*}
	 h^{(1,0)}_{\infty}(x) \triangleq \lim_{c \rightarrow 
	\infty}\frac{h^{(1,0)}(cx)}{c} = \lim_{c \rightarrow \infty}\frac{\mathbb{E}^{2}\left[\mathbf{r}_{t}\right]-cx}{c} = \lim_{c \rightarrow \infty}\frac{\mathbb{E}^{2}\left[\mathbf{r}_{t}\right]}{c} - x = -x.
	\end{flalign*}
	 It is easy to verify that the above flow (\ref{eq:upsode0}) is globally asymptotically stable to the origin. Therefore, by appealing to the Borkar-Meyn theorem, we obtain that the iterates $\{\upsilon^{(0)}_{t}\}_{t \in \mathbb{N}}$ are almost surely stable, \emph{i.e.}, 
	 \begin{flalign}
	 	 \sup_{t \in \mathbb{N}}{\Vert \upsilon^{(0)}_{t} \Vert} < \infty \hspace*{2mm} a.s.
	 \end{flalign}
	Similarly we can show that 
	\begin{flalign}
	&\sup_{t \in \mathbb{N}}{\Vert \upsilon^{(1)}_{t} \Vert} < \infty \hspace*{2mm}a.s.\\
	&\sup_{t \in \mathbb{N}}{\Vert \upsilon^{(2)}_{t} \Vert} < \infty \hspace*{2mm}a.s.\\
	&\sup_{t \in \mathbb{N}}{\Vert \upsilon^{(3)}_{t} \Vert} < \infty \hspace*{2mm}a.s.	
	\end{flalign}
	\end{enumerate}	
	Since we have hitherto established the necessary conditions (C1-C4), now by appealing to Theorem 2, Chapter 2 of \cite{borkar2008stochastic}, we can forthwith guarantee the asymptotic equivalence between the recursion (\ref{eqn:upsrec0}) and the following ODE (\emph{i.e.}, the recursion (\ref{eqn:upsrec0}) asymptotically tracks the following ODE):
	\begin{flalign}\label{eqn:upsode0}
	\frac{d}{dt}\upsilon^{(0)}(t) = \mathbb{E}^{2}\left[\mathbf{r}_{t}\right]-\upsilon^{(0)}(t), \hspace*{6mm} t \in \bbbr_{+}.
	\end{flalign}
	Similarly, we can guarantee the independent asymptotic equivalences between the recursions (\ref{eqn:upsrec1})-(\ref{eqn:upsrec3}) and the ODEs (\ref{eqn:upsode1})-(\ref{eqn:upsode3}) respectively.
	\begin{flalign}
	&\frac{d}{dt}\upsilon^{(1)}(t) =  \gamma^{2}\mathbb{E}\left[\phi^{\prime}_{t}\right]\mathbb{E}\left[\phi^{\prime}_{t}\right]^\top-\upsilon^{(1)}(t), \hspace*{10mm} t \in \bbbr_{+}, \label{eqn:upsode1} \\
	&\frac{d}{dt}\upsilon^{(2)}(t) = \mathbb{E}\left[\mathbf{r}_{t}\big(\phi^{\prime}_{t}-\phi_{t}\big)\right] - \upsilon^{(2)}(t), \hspace*{11mm} t \in \bbbr_{+}, \label{eqn:upsode2}\\
	&\frac{d}{dt}\upsilon^{(3)}(t) = \mathbb{E}\left[\big(\phi_t - 2\gamma\phi^{\prime}_{t}\big)\phi_{t}^{\top}\right] - \upsilon^{(3)}(t), \hspace*{6mm} t \in \bbbr_{+}. \label{eqn:upsode3}
	\end{flalign}
	Note that all the above ODEs (\ref{eqn:upsode0})-(\ref{eqn:upsode3}) are linear, first-order ODEs and further qualitative analysis reveals that the individual flows defined by the various ODEs are globally asymptotically stable. An examination of the trajectories of the ODEs attests that the limiting behaviour of the individual flows defined by the ODEs (\ref{eqn:upsode0})-(\ref{eqn:upsode3}) satisfies the following:
	\begin{flalign}\label{eqn:updode-asym-behaviour}
	\left.
	\begin{aligned}
	&\upsilon^{(0)}(t) \rightarrow \mathbb{E}^{2}\left[\mathbf{r}_{t}\right]\hspace*{2mm} \textrm{ as } t \rightarrow \infty.\\
	&\upsilon^{(1)}(t) \rightarrow \gamma^{2}\mathbb{E}\left[\phi^{\prime}_{t}\right]\mathbb{E}\left[\phi^{\prime}_{t}\right]^\top\hspace*{2mm} \textrm{ as } t \rightarrow \infty.\\
	&\upsilon^{(2)}(t) \rightarrow \mathbb{E}\left[\mathbf{r}_{t}\big(\phi^{\prime}_{t}-\phi_{t}\big)\right]\hspace*{2mm} \textrm{ as } t \rightarrow \infty.\\
	&\upsilon^{(3)}(t) \rightarrow \mathbb{E}\left[\big(\phi_t - 2\gamma\phi^{\prime}_{t}\big)\phi_{t}^{\top}\right]\hspace*{2mm} \textrm{ as } t \rightarrow \infty.
	\end{aligned}
	\hspace*{20mm}\right\}	
	\end{flalign}
	
	Finally, Eq. (\ref{eqn:updode-asym-behaviour}) and the previously established asymptotic equivalence between the recursions (\ref{eqn:upsrec0})-(\ref{eqn:upsrec3}) and their respective associated ODEs (\ref{eqn:upsode0})-(\ref{eqn:upsode3}) ascertains the following:
	\begin{flalign*}
	&\lim_{t \rightarrow \infty}\upsilon^{(0)}_{t} = \mathbb{E}^{2}\left[\mathbf{r}_{t}\right]  \textit{ a.s. } = \upsilon^{(0)}_{*}.\\ 
	&\lim_{t \rightarrow \infty}\upsilon^{(1)}_{t} = \gamma^{2}\mathbb{E}\left[\phi^{\prime}_{t}\right]\mathbb{E}\left[\phi^{\prime}_{t}\right]^\top \textit{ a.s. } = \upsilon^{(1)}_{*}. \\ 
	&\lim_{t \rightarrow \infty}\upsilon^{(2)}_{t} = \mathbb{E}\left[\mathbf{r}_{t}\big(\phi^{\prime}_{t}-\phi_{t}\big)\right] \textit{ a.s. } = \upsilon^{(2)}_{*}.\\
	&\lim_{t \rightarrow \infty}\upsilon^{(3)}_{t} = \mathbb{E}\left[\big(\phi_t - 2\gamma\phi^{\prime}_{t}\big)\phi_{t}^{\top}\right] \textit{ a.s. } = \upsilon^{(3)}_{*}.
	\end{flalign*}
	
	Putting all the above together, we get, for $z \in \mathcal{Z}$,
	\begin{flalign*}
	\lim_{t \rightarrow \infty} \bar{\mathcal{J}}_{b}(\upsilon_{t}, z) = \bar{\mathcal{J}}_{b}(\upsilon_{*}, z)  = \mathcal{J}_{b}(z) \hspace*{2mm} a.s.
	\end{flalign*}
\end{Proof}\\\\
\noindent
\noindent
\textbf{Notation: } We denote by $\mathbb{E}_{\widehat{\theta}}[\cdot]$ the expectation \emph{w.r.t.} the mixture PDF $\widehat{f}_{\theta}$ and $\mathbb{P}_{\widehat{\theta}}$ denotes its induced probability measure. Also, $\gamma_{\rho}(\cdot, \widehat{\theta})$ represents the $(1-\rho)$-quantile \emph{w.r.t.} the mixture PDF $\widehat{f}_{\theta}$.\\

The following result characterizes the asymptotic behaviour of the stochastic recursions (\ref{eqn:algamma}-\ref{eqn:xi1}):
\begin{lemma}\label{lmn:xiconv}
Assume $\omega_{t} \equiv \omega$, $\theta_{t} \equiv \theta$, $\forall t \in \mathbb{N}$. Let Assumption (A2) hold. Also, let the step-size sequences $\{\alpha_{t}\}_{t \in \mathbb{N}}$ and $\{\beta_{t}\}_{t \in \mathbb{N}}$ satisfy Eq. (\ref{eqn:learnrt}). Then,
\begin{enumerate}
\item
The sequence $\{\gamma_t\}_{t \in \mathbb{N}}$ generated by  Eq. (\ref{eqn:algamma}) satisfies 
\begin{flalign*}
\lim_{t \rightarrow \infty}\gamma_t = \gamma_{\rho}(\bar{\mathcal{J}}_{p}(\omega, \cdot), \widehat{\theta}) \hspace*{2mm} a.s.
\end{flalign*}
\item
 The sequence $\{\xi^{(0)}_{t}\}_{t \in \mathbb{N}}$ generated by  Eq. (\ref{eqn:xi0}) satisfies
\begin{flalign*}
\lim_{t \rightarrow \infty} \xi^{(0)}_{t} = \xi^{(0)}_{\omega, \theta} = \frac{\mathbb{E}_{\widehat{\theta}}\left[\mathbf{g}_{1}\left(\bar{\mathcal{J}}_{p}({\omega},\mathbf{Z}), \mathbf{Z}, \gamma_{\rho}(\bar{\mathcal{J}}_{p}(\omega, \cdot), \widehat{\theta})\right)\right]}{\mathbb{E}_{\widehat{\theta}}\left[\mathbf{g}_{0}\left(\bar{\mathcal{J}}_{p}({\omega},\mathbf{Z}), \gamma_{\rho}(\bar{\mathcal{J}}_{p}(\omega, \cdot), \widehat{\theta})\right)\right]} \text{ a.s.}
\end{flalign*}
\item
The sequence $\{\xi^{(1)}_{t}\}_{t \in \mathbb{N}}$ generated by  Eq. (\ref{eqn:xi1}) satisfies
\begin{flalign*}
	\lim_{t \rightarrow \infty} \xi^{(1)}_{t} =  \frac{\mathbb{E}_{\widehat{\theta}}\left[\mathbf{g}_{2}\left(\bar{\mathcal{J}}_{p}({\omega}, \mathbf{Z}), \mathbf{Z}, \gamma_{\rho}(\bar{\mathcal{J}}_{p}({\omega}, \cdot), \widehat{\theta}), \xi^{(0)}_{\omega,\theta}\right)\right]}{\mathbb{E}_{\widehat{\theta}}\left[\mathbf{g}_{0}\left(\bar{\mathcal{J}}_{p}({\omega}, \mathbf{Z}), \gamma_{\rho}(\bar{\mathcal{J}}_{p}({\omega}, \cdot), \widehat{\theta})\right)\right]} \text{ a.s.}
\end{flalign*}
\item
For any $T_0 \in (0,1)$, $\{T_t\}_{t \in \mathbb{N}}$ generated by Eq. (\ref{eq:Tt}) satisfies $T_t \in (-1,1)$, $\forall t  \in \mathbb{N}$.
\item
	If $\gamma_{\rho}(\bar{\mathcal{J}}_{p}({\omega}, \cdot), \widehat{\theta}) > \gamma_{\rho}(\bar{\mathcal{J}}_{p}({\omega}, \cdot), \widehat{\theta^{p}})$, then $\{T_t\}_{t \in \mathbb{N}}$ generated by Eq. (\ref{eq:Tt}) satisfies $\lim_{t \rightarrow \infty} T_{t} = 1$ a.s.
\end{enumerate}
\end{lemma}
\begin{Proof}
	Please refer to the proofs of Proposition 1, Lemma 2 and Lemma 3 in \cite{predictsce2018}.
\end{Proof}

\begin{remark}
	Similar results can also be obtained for Algorithm \ref{algo:sce-msbr} with $\bar{\mathcal{J}}_{p}$ replaced by $\bar{\mathcal{J}}_{b}$ and $\omega$ replaced by $\upsilon$.
\end{remark}

\noindent
Finally,  we analyze the asymptotic behaviour of the model sequence $\{\theta_t\}_{t \in \mathbb{N}}$. As a preliminary requirement, we define $\Psi_p(\omega, \theta) = (\Psi^{(0)}_p(\omega, \theta), \Psi^{(1)}_p(\omega, \theta))^{\top}$, where 
\begin{flalign}
	&\Psi^{(0)}_{p}(\omega, \theta) \triangleq \frac{\mathbb{E}_{\widehat{\theta}}\left[ \mathbf{g}_{1}\left(\mathcal{J}_{p}({\omega}, \mathbf{Z}), \mathbf{Z}, \gamma_{\rho}(\mathcal{J}_{p}({\omega}, \cdot), \widehat{\theta})\right)\right]}{\mathbb{E}_{\widehat{\theta}}\left[\mathbf{g}_{0}\left(\mathcal{J}_{p}({\omega}, \mathbf{Z}), \gamma_{\rho}(\mathcal{J}_{p}({\omega}, \cdot), \widehat{\theta})\right)\right]},\hspace*{10mm}\\
	&\Psi^{(1)}_{p}(\omega, \theta) \triangleq  \frac{\mathbb{E}_{\widehat{\theta}}\left[\mathbf{g}_{2}\left(\mathcal{J}_{p}({\omega}, \mathbf{Z}), \mathbf{Z}, \gamma_{\rho}(\mathcal{J}_{p}({\omega}, \cdot), \widehat{\theta}), \Psi^{((0)}_p(\omega, \theta)\right)\right]}{\mathbb{E}_{\widehat{\theta}}\left[\mathbf{g}_{0}\left(\mathcal{J}_{p}({\omega}, \mathbf{Z}), \gamma_{\rho}(\mathcal{J}_{p}({\omega}, \cdot), \widehat{\theta})\right)\right]}.
\end{flalign}
Similarly, we define $\Psi_{b}$ with $\mathcal{J}_{p}$ replaced by $\mathcal{J}_{b}$ and $\omega$ replaced by $\upsilon$.

We now state our main theorems. The first theorem states that the model sequence $\{\theta_{t}\}_{t \in \mathbb{N}}$ generated by Algorithm \ref{algo:sce-mspbem} almost surely converges to $\theta^{p*} = (z_{p}^{*}, 0_{k \times k})^{\top}$, the degenerate distribution concentrated at $z_{p}^{*}$, where $z_{p}^{*}$ is the solution to the optimization problem (\ref{eqn:mspbeobj}) which minimizes the error function MSPBE.
\begin{theorem}\label{thm:main}
	(MSPBE Convergence) Let $S(z) = exp(rz)$, $r \in \bbbr_{+}$.  Let $\rho \in (0,1)$ and $\lambda \in (0,1)$. Let $\theta_0 = (\mu_0, qI_{k \times k})^{\top}$, where $q \in \bbbr_{+}$. Let the step-size sequences $\{\alpha_t\}_{t \in \mathbb{N}}$, $\{\beta_t\}_{t \in \mathbb{N}}$ satisfy Eq. (\ref{eqn:learnrt}). Also let $c_t \rightarrow 0$. Suppose $\{\theta_t = (\mu_t, \Sigma_t)^{\top}\}_{t \in \mathbb{N}}$ is the sequence generated by Algorithm \ref{algo:sce-mspbem} and assume $\theta_{t} \in \Theta$, $\forall t \in \mathbb{N}$. Also, let the Assumptions (A1), (A2) and (A3) hold. Further, we assume that there exists a continuously differentiable function $V:U \rightarrow \bbbr_{+}$, where $U \subseteq \Theta$ is an open neighbourhood of $\theta^{p*}$ with  $\nabla V(\theta)^{\top}\Psi_{p}(\omega_{*}, \theta) < 0$, $\forall \theta \in U\smallsetminus\{\theta^{p*}\}$ and $\nabla V(\theta^{p*})^{\top}\Psi_{p}(\omega_{*}, \theta^{p*}) = 0$. Then, there exists $q^{*} \in \bbbr_{+}$ and $r^{*} \in \bbbr_{+}$ s.t. $\forall q > q^{*}$ and $\forall r > r^{*}$,
\begin{gather*}
	\hspace{1mm} \lim_{t \rightarrow \infty} \bar{\mathcal{J}}_{p}(\omega_{t}, \mu_{t}) = \mathcal{J}_{p}^{*}  \hspace{3mm} and \hspace{2mm} \lim_{t \rightarrow \infty}\theta_{t} = \theta^{p*} = (z_{p}^{*}, 0_{k \times k})^{\top} \textrm{a.s.},
\end{gather*}
where $\mathcal{J}_{p}^{*}$ and $z_{p}^{*}$ are defined in Eq. (\ref{eqn:mspbeobj}). Further, since $\mathcal{J}_{p} = -{MSPBE}$, the algorithm SCE-MSPBEM converges to the global minimum of MSPBE a.s.
\end{theorem}
\begin{Proof}
Please refer to the proof of Theorem 1 in \cite{predictsce2018}.
\end{Proof}\\

Similarly for Algorithm \ref{algo:sce-msbr}, the following theorem states that the model sequence $\{\theta_{t}\}_{t \in \mathbb{N}}$ generated by Algorithm \ref{algo:sce-msbr} almost surely converges to $\theta^{b*} = (z_{b}^{*}, 0_{k \times k})^{\top}$, the degenerate distribution concentrated at $z_{b}^{*}$, where $z_{b}^{*}$ is the solution to the optimization problem (\ref{eqn:msbrobj}) which minimizes the error function MSBR.
\begin{theorem}\label{thm:main2}
	(MSBR Convergence) Let $S(z) = exp(rz)$, $r \in \bbbr_{+}$.  Let $\rho \in (0,1)$ and $\lambda \in (0,1)$. Let $\theta_0 = (\mu_0, qI_{k \times k})^{\top}$, where $q \in \bbbr_{+}$. Let the step-size sequences $\{\alpha_t\}_{t \in \mathbb{N}}$, $\{\beta_t\}_{t \in \mathbb{N}}$ satisfy Eq. (\ref{eqn:learnrt}). Also let $c_t \rightarrow 0$. Suppose $\{\theta_t = (\mu_t, \Sigma_t)^{\top}\}_{t \in \mathbb{N}}$ is the sequence generated by Algorithm \ref{algo:sce-msbr} and assume $\theta_{t} \in \Theta$, $\forall t \in \mathbb{N}$. Also, let the Assumptions (A1), (A2) and (A3)$^{\prime}$ hold. Further, we assume that there exists a continuously differentiable function $V:U \rightarrow \bbbr_{+}$, where $U \subseteq \Theta$ is an open neighbourhood of $\theta^{b*}$ with $\nabla V(\theta)^{\top}\Psi_{b}(\upsilon_{*}, \theta) < 0$, $\forall \theta \in U\smallsetminus\{\theta^{b*}\}$ and $\nabla V(\theta^{b*})^{\top}\Psi_{b}(\upsilon_{*}, \theta^{b*}) = 0$. Then, there exists $q^{*} \in \bbbr_{+}$ and $r^{*} \in \bbbr_{+}$ s.t. $\forall q > q^{*}$ and $\forall r > r^{*}$,
\begin{gather*}
	\hspace{1mm} \lim_{t \rightarrow \infty} \bar{\mathcal{J}}_{b}(\upsilon_{t}, \mu_{t}) = \mathcal{J}_{b}^{*}  \hspace{3mm} and \hspace{2mm} \lim_{t \rightarrow \infty}\theta_{t} = \theta^{b*} = (z_{b}^{*}, 0_{k \times k})^{\top} \textrm{a.s.},
\end{gather*}
where $\mathcal{J}_{b}^{*}$ and $z_{b}^{*}$ are defined in Eq. (\ref{eqn:msbrobj}). Further, since $\mathcal{J}_{b} = -{MSBR}$, the algorithm SCE-MSBRM converges to the global minimum of MSBR a.s.
\end{theorem}
\begin{Proof}
Please refer to the proof of Theorem 1 in \cite{predictsce2018}.
\end{Proof}

\subsection{Discussion of the Proposed Algorithms}
\textit{The computational load of the algorithms SCE-MSPBEM and SCE-MSBRM is $\Theta(k^{2})$ per iteration} which is primarily attributed to the computation of Eqs. (\ref{eqn:alomgupd}) and (\ref{eqn:bralupsupd}) respectively. Least squares algorithms like LSTD and LSPE also require $\Theta(k^{2})$ per iteration. However, LSTD requires an extra operation of inverting the $k \times k$ matrix $A_{T}$ (Algorithm \ref{algo:lstdlbda}) which requires an extra computational effort of $\Theta(k^{3})$. (Note that LSPE also requires a $k \times k$ matrix inversion). This makes the \textit{overall complexity of LSTD and LSPE to be $\Theta(k^{3})$}. Further in some cases the matrix $A_{T}$ may not be invertible. In that case, the pseudo inverse of $A_{T}$ needs to be obtained in LSTD and LSPE which is computationally even more expensive. Our algorithm does not require such an inversion procedure. Also \textit{even though the complexity of the first order temporal difference algorithms such as TD($\lambda$) and GTD2 is $\Theta(k)$, the approximations they produced in the experiments we conducted turned out to be inferior to ours and also showed a slower rate of convergence than our algorithm.} Another noteworthy characteristic exhibited by our algorithm is \textit{stability}. Recall that the convergence of TD($0$) is guaranteed by the requirements that the Markov chain of $\mathrm{P}^{\pi}$ should be ergodic with the sampling distribution $\nu$ as its stationary distribution. The classic example of Baird's 7-star \cite{baird1995residual} violates those restrictions and hence TD(0) is seen to diverge. However, our algorithm does not impose such restrictions and shows stable behaviour even in non-ergodic cases such as the Baird's example.

But the significant feature of SCE-MSPBEM/SCE-MSBRM is its ability to find the global optimum. This particular characteristic of the algorithm enables it to produce high quality solutions when applied to non-linear function approximation, where the convexity of the objective function does not hold in general. Also note that SCE-MSPBEM/SCE-MSBRM is a gradient-free technique and hence does not require strong structural restrictions on the objective function. 

\section{Experimental Results}
We present here a numerical comparison of our algorithms with various state-of-the-art algorithms in the literature on some benchmark reinforcement learning problems. In each of the experiments, a random trajectory $\{(\mathbf{s}_t, \mathbf{r}_{t}, \mathbf{s}^{\prime}_{t})\}_{t=0}^{\infty}$ is chosen and all the algorithms are updated using it. Each $\mathbf{s}_t$ in $\{(\mathbf{s}_t, \mathbf{r}_{t}, \mathbf{s}^{\prime}_{t}), t \geq 0\}$ is sampled using an arbitrary distribution $\nu$ over $\mathbb{S}$. The algorithms are run on $10$ independent trajectories and the average of the results obtained is plotted. The $x$-axis in the plots is $t/1000$, where $t$ is the iteration number. The function $S(\cdot)$ is chosen as $S(x) = \exp{(rx)}$, where $r \in \bbbr_{+}$ is chosen appropriately. In all the test cases, the evolution of the model sequence $\{\theta_t\}$  across independent trials was almost homogeneous and hence we omit the standard error bars from our plots.\\

We evaluated the performance of our algorithms on the following benchmark problems:
\begin{enumerate}
	\item
	Linearized cart-pole balancing \cite{dann2014policy}.
	\item
	5-Link actuated pendulum balancing \cite{dann2014policy}.
	\item
	Baird's $7$-star MDP \cite{baird1995residual}.
	\item
	$10$-state ring MDP \cite{kveton2006solving}.
	\item
	MDPs with radial basis functions and Fourier basis functions \cite{konidaris2011value}.
	\item
	Settings involving non-linear function approximation \cite{tsitsiklis1997analysis}.
\end{enumerate}
\subsection{Experiment 1: Linearized Cart-Pole Balancing \cite{dann2014policy}}
\begin{itemize}
	\item 
	\textbf{Setup: } A pole with mass $m$ and length $l$ is connected to a cart of mass $M$. It can rotate $360^{\circ}$ and the cart is free to move in either direction within the bounds of a linear track.
	\item 
	\textbf{Goal: } To balance the pole upright and the cart at the centre of the track.
	\item 
	\textbf{State space: }The 4-tuple $(x, \dot{x}, \psi, \dot{\psi})^{\top}$ $\in \bbbr^{4}$, where $\psi$ is  the angle of the pendulum with respect to the vertical axis, $\dot{\psi}$ is the angular velocity, $x$ the relative cart position from the centre of the track and $\dot{x}$ is its velocity.
	\item 
	\textbf{Control space: }The controller applies a horizontal force $a \in \bbbr$ on the cart parallel to the track. The stochastic policy used in this setting corresponds to $\pi(a|s) = \mathcal{N}(a | \beta_{1}^{\top}s, \sigma_1^{2})$, where $\beta_1 \in \bbbr^{4}$ and $\sigma_1 \in \bbbr$.
	\item 
	\textbf{System dynamics: }
	The dynamical equations of the system are given by
	\begin{equation}
	\ddot{\psi} = \frac{-3ml\dot{\psi}^{2}\sin{\psi}\cos{\psi}+(6M+m)g\sin{\psi}-6(a-b\dot{\psi})\cos{\psi}}{4l(M+m)-3ml\cos{\psi}},
	\end{equation}
	\begin{equation}
	\ddot{x} = \frac{-2ml\dot{\psi}^{2}\sin{\psi}+3mg\sin{\psi}\cos{\psi}+4a-4b\dot{\psi}}{4(M+m)-3m\cos{\psi}}.
	\end{equation}
	By making further assumptions on the initial conditions, the system dynamics can be approximated accurately by the linear system 
	\begin{equation}
	\begin{bmatrix} 
	x_{t+1}\\
	\dot{x}_{t+1}\\
	\psi_{t+1}\\
	\dot{\psi}_{t+1}
	\end{bmatrix} = \begin{bmatrix} 
	x_{t}\\
	\dot{x}_{t}\\
	\psi_{t}\\
	\dot{\psi}_{t}
	\end{bmatrix} + \Delta t \begin{bmatrix}
	\dot{\psi}_{t} \\
	\frac{3(M+m)\psi_t-3a+3b\dot{\psi_t}}{4Ml-ml} \\
	\dot{x}_{t} \\
	\frac{3mg\psi_t + 4a - 4b\dot{\psi_t}}{4M-m}
	\end{bmatrix} + 
	\begin{bmatrix}
	0 \\
	0 \\
	0 \\
	\mathbf{z}
	\end{bmatrix},
	\end{equation}
	where $\Delta t$ is the integration time step, \emph{i.e.}, the time difference between two transitions and $\mathbf{z}$ is a Gaussian noise on the velocity of the cart with standard deviation $\sigma_{2}$.
	\item 
	\textbf{Reward function: }
	$\mathrm{R}(s, a) = \mathrm{R}(\psi, \dot{\psi}, x, \dot{x}, a) = -100\psi^2 - x^2 - \frac{1}{10}a^2$.
	\item 
	\textbf{Feature vectors: } $\phi(s \in \bbbr^{4}) = (1, s_{1}^{2}, s_{2}^{2} \dots, s_{1}s_{2}, s_{1}s_{3}, \dots, s_{3}s_{4})^{\top} \in \bbbr^{11}$.
	\item 
	\textbf{Evaluation policy: }The policy evaluated in the experiment is the optimal policy $\pi^{*}(a | s) = \mathcal{N}(a | {\beta_{1}^{*}}^{\top}s, {\sigma_{1}^{*}}^{2})$. The parameters $\beta_{1}^{*}$ and $\sigma_{1}^{*}$ are computed using dynamic programming. The feature set chosen above is a perfect feature set, \emph{i.e.}, $V^{\pi^{*}} \in \{\Phi z \vert z \in \bbbr^{k}\}$.
\end{itemize}
\begin{figure}[!h]
	\centering
	{\includegraphics[height=50mm, width=75mm]{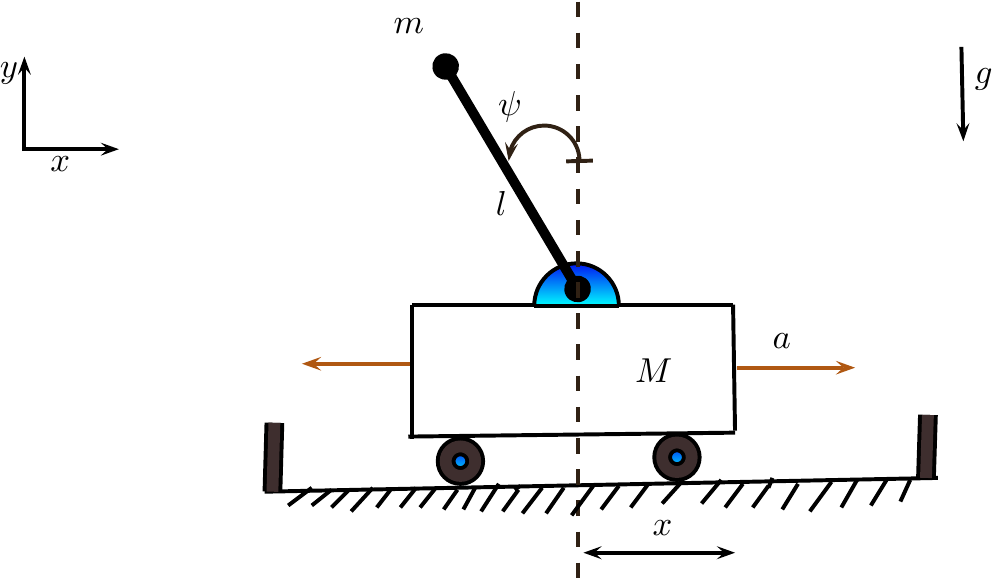}}
	\caption{The cart-pole system. The goal is to keep the pole in the  upright position and the cart at the center of the track by pushing the cart with a force $a$ either to the left or the right. The system is parametrized by the position $x$ of the cart, the angle of the pole $\psi$, the velocity $\dot{x}$ and the angular velocity $\dot{\psi}$.}
\end{figure}%
\begin{figure}[!h]
	\begin{subfigure}[h]{0.5\textwidth}
		{\includegraphics[height=55mm, width=58mm]{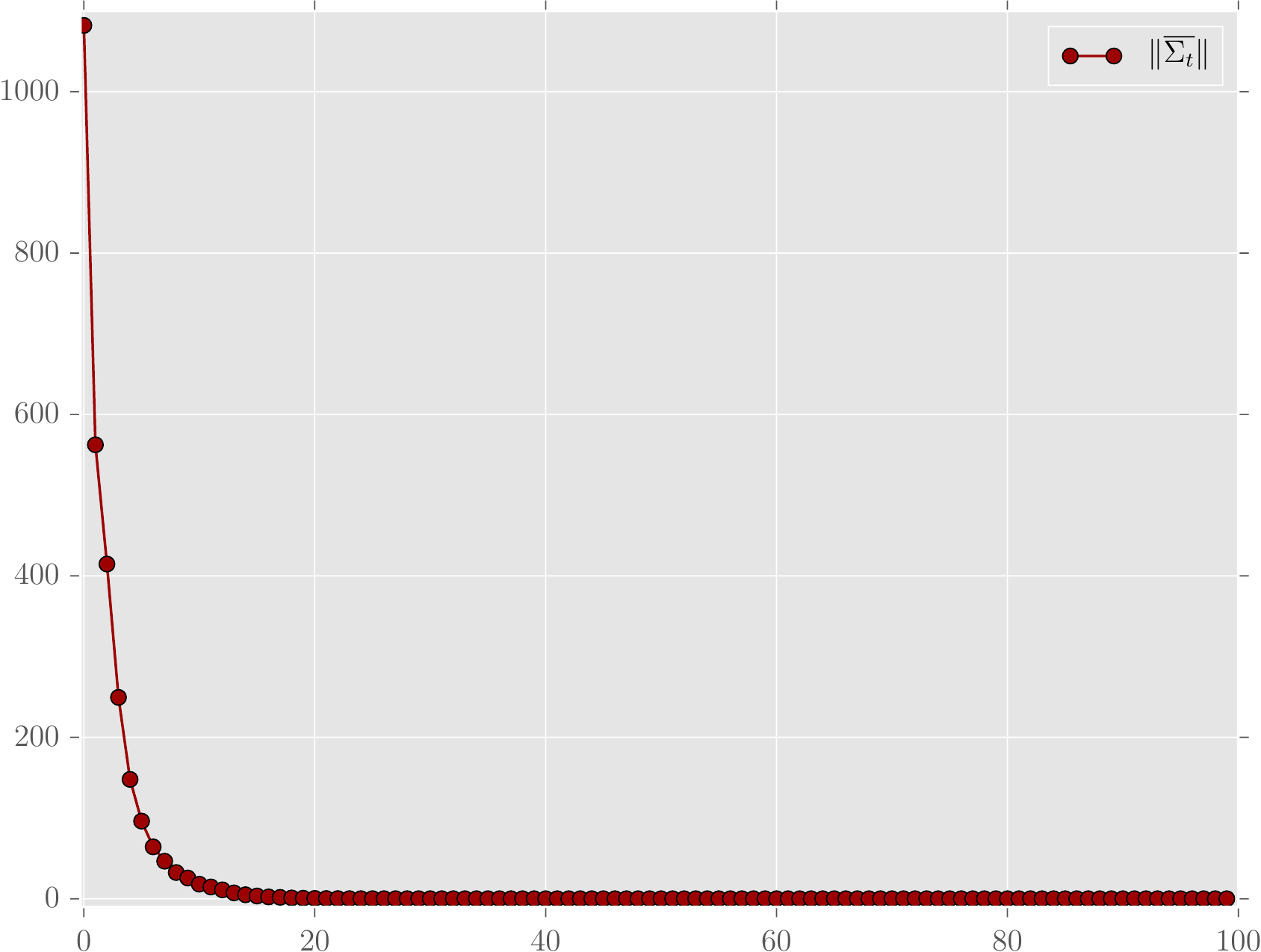}}
		\subcaption{Plot of $\Vert \Sigma_{t} \Vert_{F}$  (where $\Vert\cdot\Vert_{F}$ is the Frobenius norm)}
	\end{subfigure}\hspace*{2mm}
	\begin{subfigure}[h]{0.5\textwidth}\vspace*{-4mm}
		{\includegraphics[height=55mm, width=58mm]{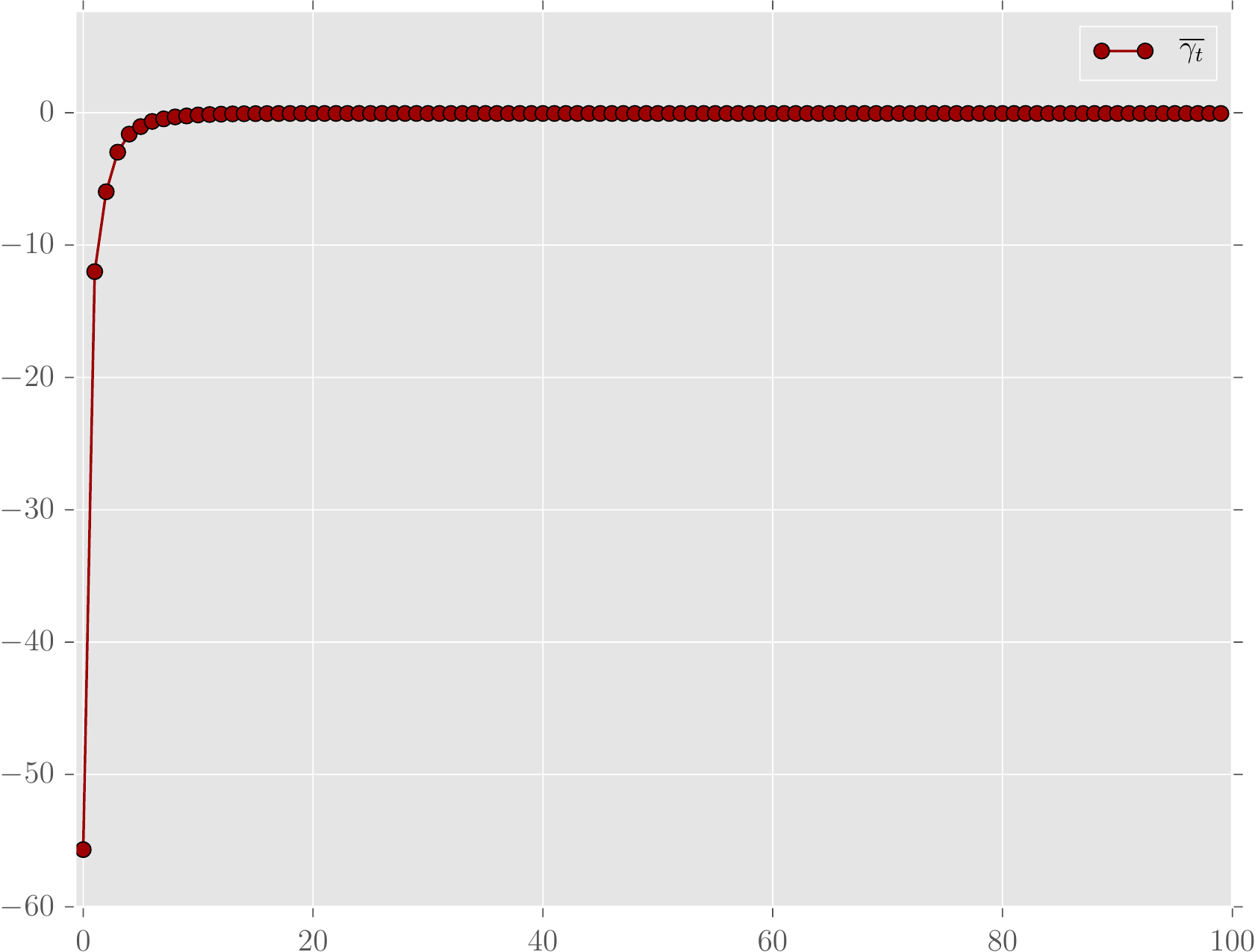}}
		\subcaption{Plot of $\gamma^{p}_{t}$}
	\end{subfigure}\\
	\begin{subfigure}[h]{0.5\textwidth}
		{\includegraphics[height=55mm, width=58mm]{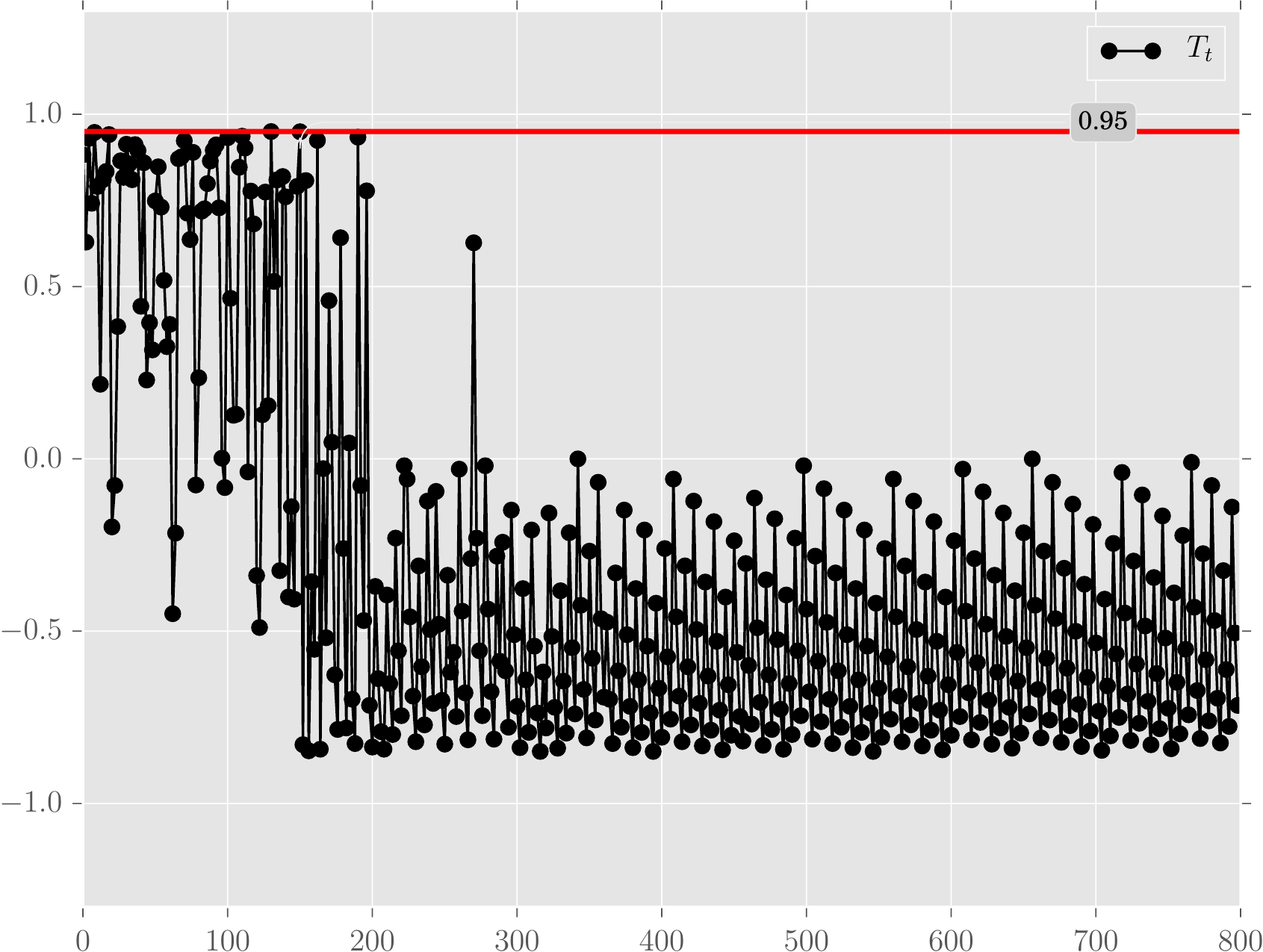}}
		\subcaption{Plot of $T_{t}$}
	\end{subfigure}\hspace*{2mm}
	\begin{subfigure}[h]{0.5\textwidth}
		{\includegraphics[height=55mm, width=58mm]{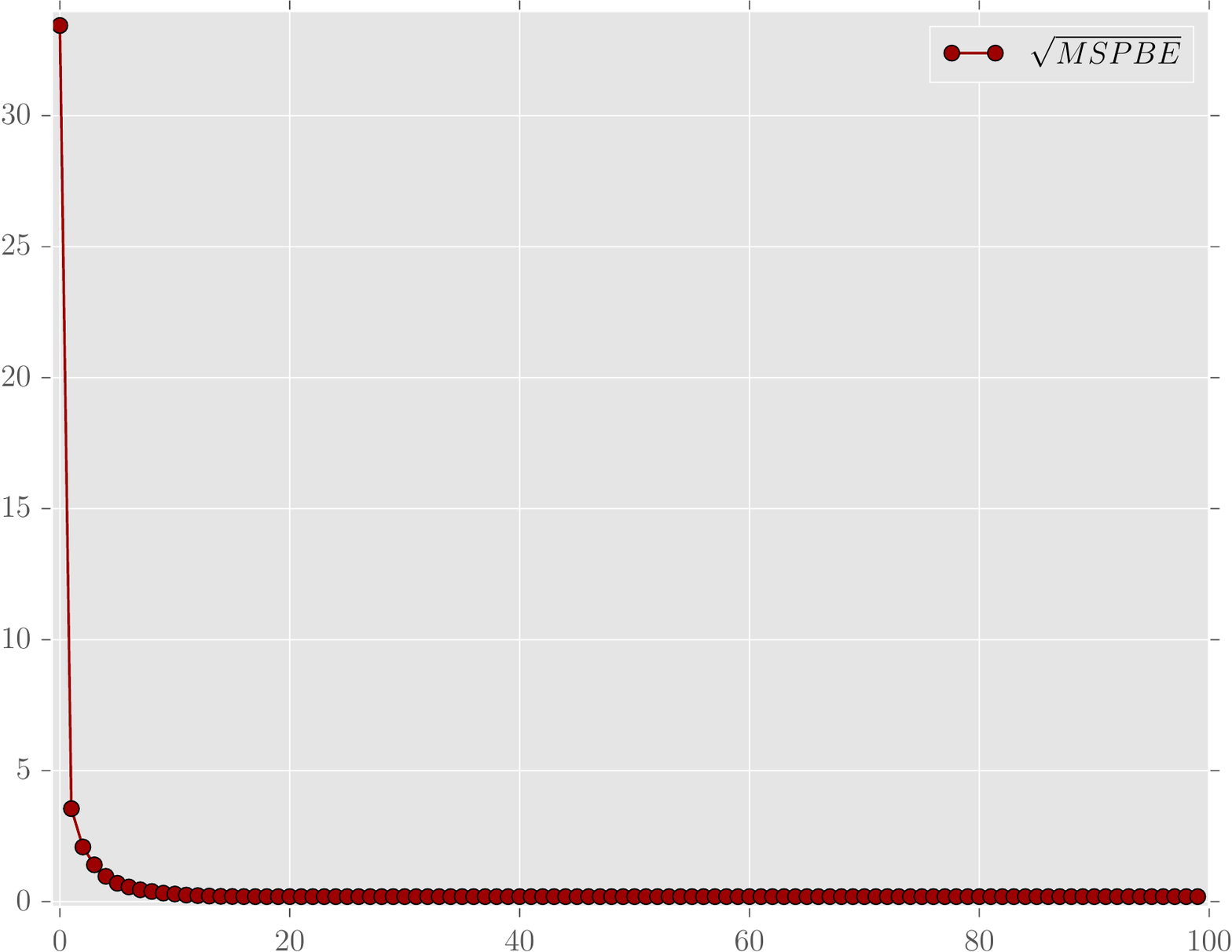}}
		\subcaption{Plot of $\sqrt{\mathrm{MSPBE}(\mu_{t})}$}
	\end{subfigure}%
	\caption{The cart-pole setting. The evolutionary trajectory of the variables $\Vert \Sigma_{t} \Vert_{F}$  (where $\Vert\cdot\Vert_{F}$ is the Frobenius norm), $\gamma^{p}_{t}$, $T_{t}$ and $\sqrt{\mathrm{MSPBE}(\mu_{t})}$. Note that both $\gamma^{p}_{t}$ and $\sqrt{\mathrm{MSPBE}(\mu_{t})}$  converge to $0$ as $t \rightarrow \infty$, while $\Vert \Sigma_{t} \Vert_{F}$ also converges to $0$. This implies that the model $\theta_{t} = (\mu_t,\Sigma_t)^{\top}$ converges to the degenerate distribution concentrated on $z^{*}$. The evolutionary track of $T_{t}$ shows that $T_{t}$ does not cross the $\epsilon_{1} = 0.95$ line after the model $\theta_t = (\mu_{t}, \Sigma_{t})^{\top}$ reaches a close neighbourhood of its limit. }\label{fig:cartpoleres}
\end{figure}
Here the sample trajectory is obtained by a continuous roll-out of a particular realization of the underlying Markov chain and hence it is of on-policy nature. Therefore, the sampling distribution is the stationary distribution (steady-state distribution) of the Markov chain induced by the policy being evaluated (see Remark \ref{rem:rem1}).
The various parameter values we used in our experiment are provided in Table \ref{tab:predcartpole5} of Appendix.
The results of the experiments are shown in Fig. \ref{fig:cartpoleres}.
\newpage
\subsection{Experiment 2: 5-Link Actuated Pendulum Balancing \cite{dann2014policy}}
\begin{figure}[!h]
	\hspace*{-3mm}
	\begin{subfigure}[h]{0.5\textwidth}
		\includegraphics[height=52mm, width=60mm]{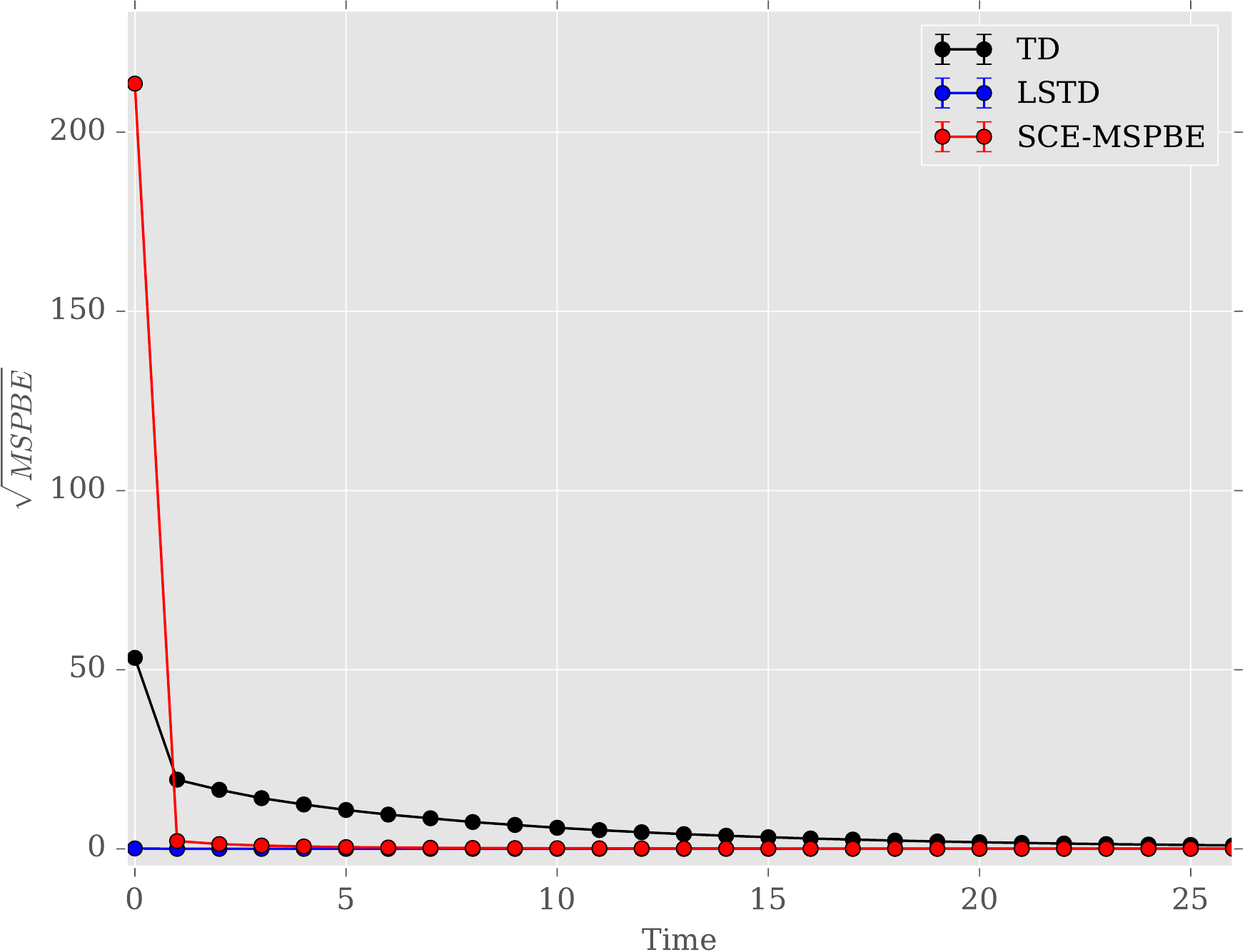}
		\subcaption{$\sqrt{\mathrm{MSPBE}(\mu_{t})}$}
	\end{subfigure}\hspace*{2mm}
	\begin{subfigure}[h]{0.5\textwidth}
		\includegraphics[width=60mm, height=52mm]{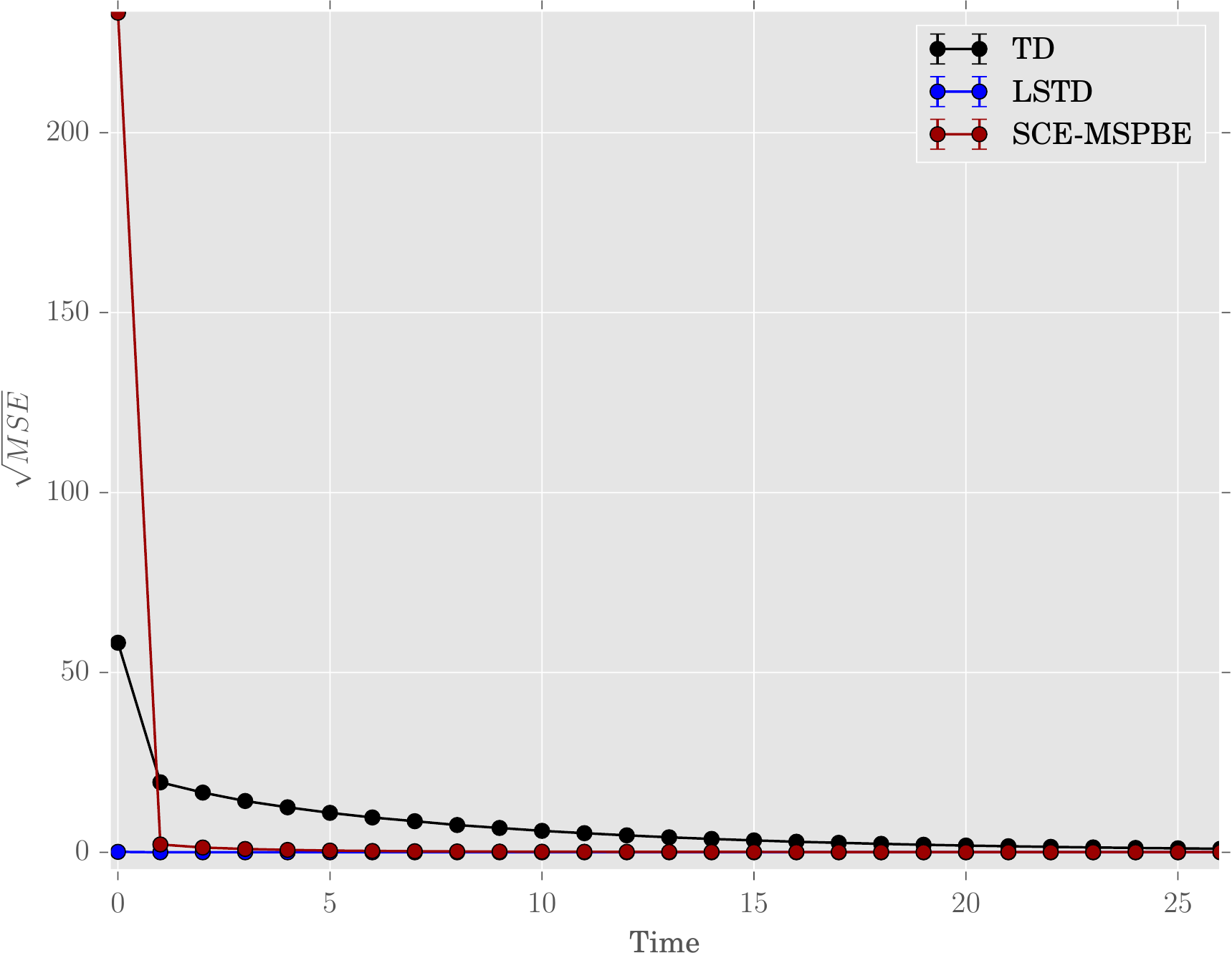}
		\subcaption{$\sqrt{\mathrm{MSE}(\mu_{t})}$}
	\end{subfigure}	
	\caption{$5$-link actuated pendulum setting. The respective trajectories of the $\sqrt{\mathrm{MSPBE}}$ and $\sqrt{\mathrm{MSE}}$ generated by TD(0), LSTD(0) and SCE-MSPBEM algorithms are plotted. The graph on the left is for $\sqrt{\mathrm{MSPBE}}$ , while on the right is that of $\sqrt{\mathrm{MSE}}$. Note that $\sqrt{\mathrm{MSE}}$ also converges to $0$ since the feature set is perfect.}\label{fig:linkresinvpend}
\end{figure}

\begin{itemize}
	\item 
	\textbf{Setup: } $5$ independent poles each with mass $m$ and length $l$ with the top pole being a pendulum connected using $5$ rotational joints.
	\item
	\textbf{Goal: } To keep all the poles in the upright position by applying independent torques at each joint.
	\item
	\textbf{State space: }The state $s = (q, \dot{q})^{\top} \in \bbbr^{10}$, where $q = (\psi_{1}, \psi_{2}, \psi_{3}, \psi_{4}, \psi_{5}) \in \bbbr^{5}$ and $\dot{q} = (\dot{\psi}_{1}, \dot{\psi}_{2}, \dot{\psi}_{3}, \dot{\psi}_{4}, \dot{\psi}_{5})  \in \bbbr^{5}$ with $\psi_{i}$ being the angle of the pole $i$ with respect to the vertical axis and $\dot{\psi}_{i}$ the angular velocity.
	\item
	\textbf{Control space: }The action $a = (a_{1}, a_{2}, \dots, a_{5})^{\top} \in \bbbr^{5}$, where $a_{i}$ is the torque applied to the joint $i$. The stochastic policy  used in this setting corresponds to $\pi(a|s) = \mathcal{N}_{5}(a | \beta_{1}^{\top}s, \sigma_1^{2})$, where $\beta_1 \in \bbbr^{10 \times 5}$ and $\sigma_1 \in \bbbr^{5 \times 5}$.
	\item
	\textbf{System dynamics: } The approximate linear system dynamics is given by
	\begin{equation} 
	\begin{bmatrix} 
	q_{t+1}\\
	\dot{q}_{t+1}
	\end{bmatrix} = 
	\begin{bmatrix} 
	I && \Delta t\hspace*{1mm} I\\
	-\Delta t \hspace*{1mm}M^{-1}U && I
	\end{bmatrix}\begin{bmatrix}q_{t}\\ \dot{q}_{t}\end{bmatrix} + \Delta t \begin{bmatrix}
	0 \\
	M^{-1}
	\end{bmatrix}a + 
	\mathbf{z},
	\end{equation}
	where $\Delta t$ is the integration time step, \emph{i.e.}, the time difference between two transitions, $M$ is the mass matrix in the upright position where $M_{ij} = l^{2}(6-max(i,j))m$ and $U$ is a diagonal matrix with $U_{ii} = -gl(6-i)m$.  Each component of $\mathbf{z}$ is a Gaussian noise.
	\item
	\textbf{Reward function: } $\mathrm{R}(q, \dot{q}, a) = -q^{\top}q$.
	\item
	\textbf{Feature vectors: } $\phi(s \in \bbbr^{10}) = (1, s_{1}^{2}, s_{2}^{2} \dots, s_{1}s_{2}, s_{1}s_{3}, \dots, s_{9}s_{10})^{\top} \in \bbbr^{46}$.
	\item
	\textbf{Evaluation policy: }The policy evaluated in the experiment is the optimal policy $\pi^{*}(a | s) = \mathcal{N}(a | {\beta_{1}^{*}}^{\top}s, {\sigma_{1}^{*}}^{2})$. The parameters $\beta_{1}^{*}$ and $\sigma_{1}^{*}$ are computed using dynamic programming. The feature set chosen above is a perfect feature set, \emph{i.e.}, $V^{\pi^{*}} \in \{\Phi z \vert z \in \bbbr^{k}\}$.
\end{itemize}
\begin{figure}[!h]
	\centering
	\includegraphics[height=55mm, width=55mm]{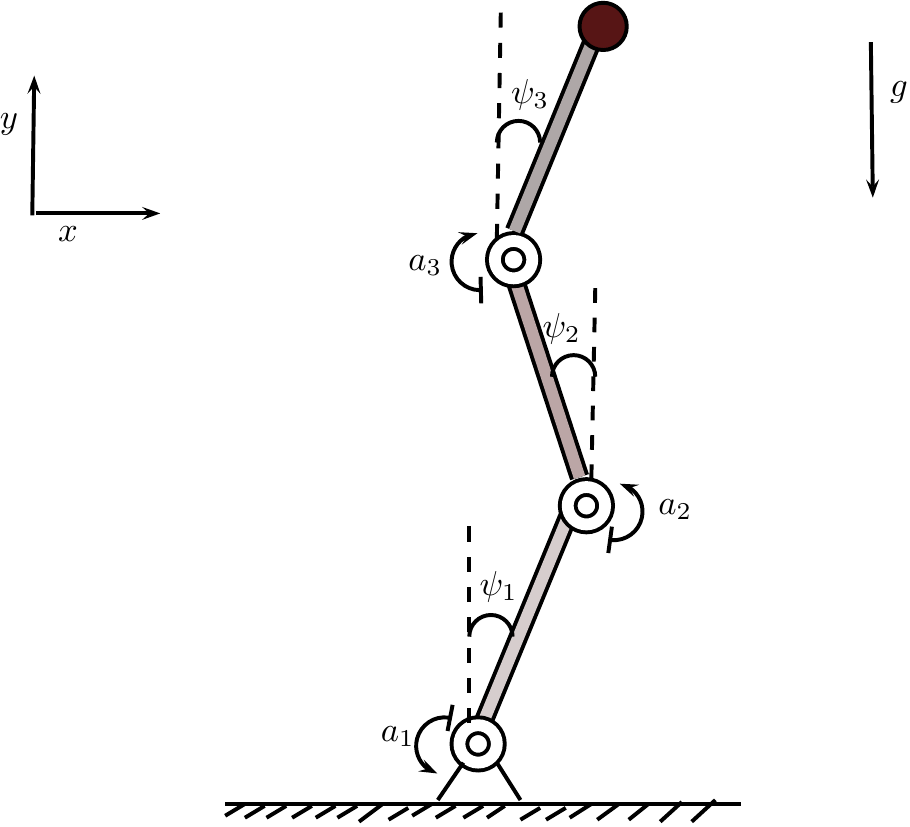}
	\caption{A $3$-link actuated pendulum setting. Each rotational joint $i$, $1 \leq i \leq 3$ is actuated by a torque $a_{i}$ . The system is parametrized by the angle $\psi_{i}$ against the vertical direction and the angular velocity $\dot{\psi}_{i}$. The goal is to balance the pole in the upright direction, \emph{i.e.}, all $\psi_{i}$ should be as close to $0$ as possible. The $5$-link actuated pendulum setting that we actually consider in the experiments is similar to this but with two additional links.}
\end{figure}
Similar to the earlier experiment, here also the sample trajectory is of on-policy nature and therefore the sampling distribution is the steady-state distribution of the Markov chain induced by the policy being evaluated (see Remark \ref{rem:rem1}).
The various parameter values we used in our experiment are provided in Table \ref{tab:invpendpred5} of Appendix. Note that we have used constant step-sizes in this experiment. The results of the experiment are shown in Fig. \ref{fig:linkresinvpend}.
	\subsection{Experiment 3: Baird's 7-Star MDP \cite{baird1995residual}}\label{subsec:bairdmdp}
Our algorithm was also tested on Baird's star problem \cite{baird1995residual}. We call it the stability test because the Markov chain in this case is not ergodic and this is a classic example where TD($0$) is seen to diverge \cite{baird1995residual}. We consider here an MDP with $\vert \mathbb{S} \vert = 7$, $\vert \mathbb{A} \vert = 2$ and $k = 8$. We let the sampling distribution $\nu$ to be the uniform distribution over $\mathbb{S}$. The feature matrix $\Phi$ and the transition matrix $\P_{\pi}$ are given by\\\\
{\setlength{\abovedisplayskip}{0pt}\setlength{\belowdisplayskip}{8pt}\begin{flalign}\label{eq:phibairds}
\Phi = \begin{pmatrix}
	1 & 2 & 0 & 0 & 0 & 0 & 0 & 0\\
	1 & 0 & 2 & 0 & 0 & 0 & 0 & 0\\
	1 & 0 & 0 & 2 & 0 & 0 & 0 & 0\\
	1 & 0 & 0 & 0 & 2 & 0 & 0 & 0\\
	1 & 0 & 0 & 0 & 0 & 2 & 0 & 0\\
	1 & 0 & 0 & 0 & 0 & 0 & 2 & 0\\
	2 & 0 & 0 & 0 & 0 & 0 & 0 & 1\\
\end{pmatrix}
\hspace*{1cm}
\P_{\pi} = \begin{pmatrix}
	0 & 0 & 0 & 0 & 0 & 0 & 1 \\
	0 & 0 & 0 & 0 & 0 & 0 & 1 \\
	0 & 0 & 0 & 0 & 0 & 0 & 1 \\
	0 & 0 & 0 & 0 & 0 & 0 & 1 \\
	0 & 0 & 0 & 0 & 0 & 0 & 1 \\
	0 & 0 & 0 & 0 & 0 & 0 & 1 \\
	0 & 0 & 0 & 0 & 0 & 0 & 1 \\
	\end{pmatrix}.
\end{flalign}
The reward function is given by $\mathrm{R}(s, s^{\prime}) = 0$, $\forall s, s^{\prime} \in \mathbb{S}$. The performance comparison of the algorithms GTD$2$, TD($0$) and LSTD($0$) with SCE-MSPBEM is shown in Fig. \ref{fig:starperf}. Here, the performance metric used for comparison is the $\sqrt{\mathrm{MSE}(\cdot)}$ of the prediction vector generated by the corresponding algorithm at time $t$.
\begin{figure}[!h]
	\begin{center}
		\includegraphics[scale=0.20]{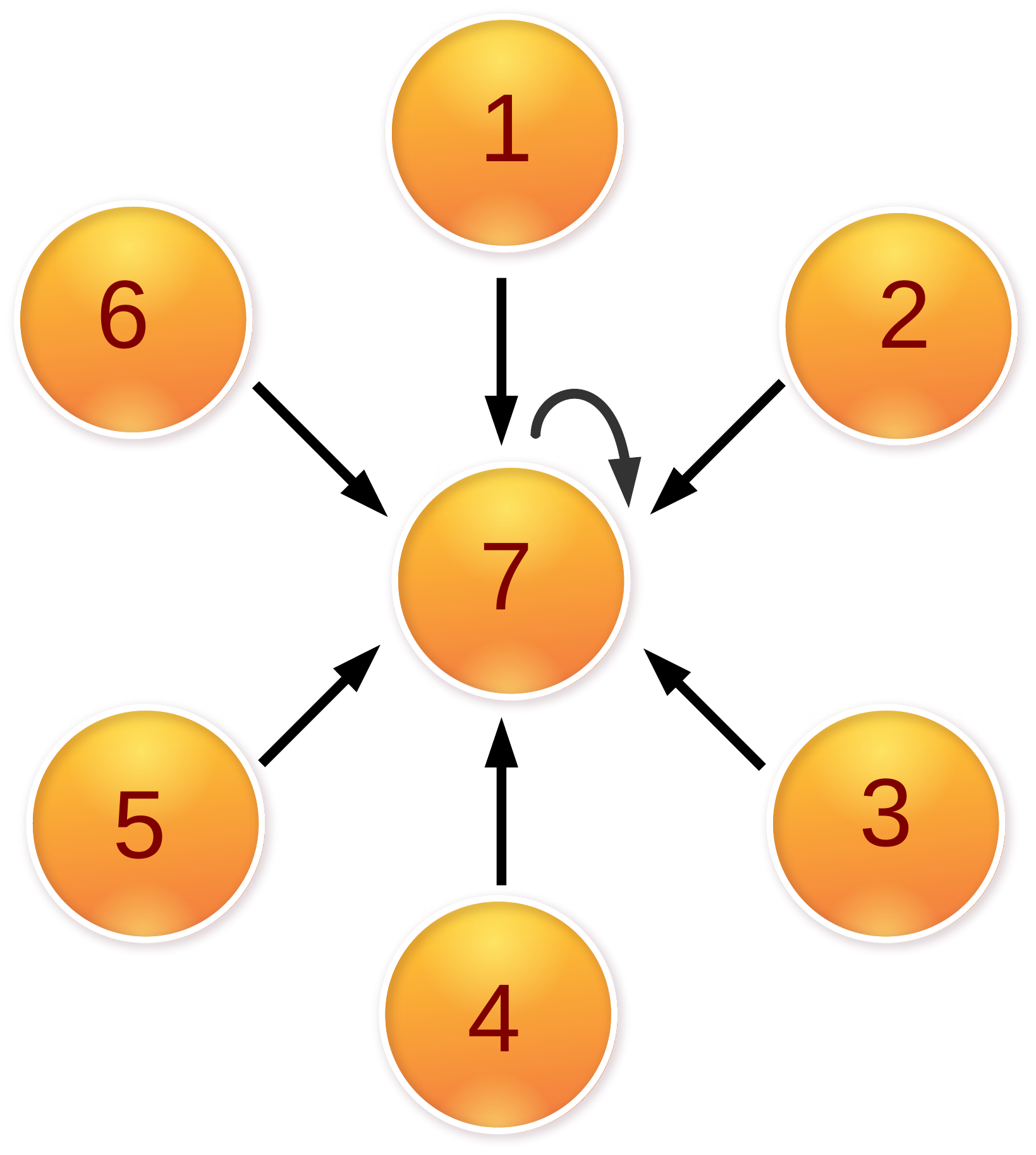}
	\end{center}
	\caption{Baird's $7$-star MDP}
\end{figure}
The algorithm parameter values used in the experiment are provided in Table \ref{tab:baird7star5} of Appendix.


A careful analysis in \cite{schoknecht2002convergent} has shown that when the discount factor $\gamma \leq 0.88$, with appropriate learning rate, TD($0$) converges. Nonetheless, it is also shown in the same paper that for discount factor $\gamma = 0.9$, TD($0$) will diverge for all values of the learning rate. This is explicitly demonstrated in Fig. \ref{fig:starperf}. However our algorithm SCE-MSPBEM converges in both cases, which demonstrates the stable behaviour exhibited by our algorithm.\\

The algorithms were also compared on the same Baird's $7$-star, but with a different feature matrix $\Phi_{1}$ as under.\vspace*{2mm}\\
\hspace*{30mm}\resizebox{0.3\linewidth}{!}{$\Phi_{1} = \begin{pmatrix}
	1 & 2 & 0 & 0 & 0 & 0 & 1 & 0\\
	1 & 0 & 2 & 0 & 0 & 0 & 0 & 0\\
	1 & 0 & 0 & 2 & 0 & 0 & 0 & 0\\
	1 & 0 & 0 & 0 & 2 & 0 & 0 & 0\\
	1 & 0 & 0 & 0 & 0 & 0 & 0 & 2\\
	1 & 0 & 0 & 0 & 0 & 0 & 0 & 3\\
	2 & 0 & 0 & 0 & 0 & 0 & 0 & 1\\
	\end{pmatrix}$}.
\vspace*{5mm}\\
In this case, the reward function is given by $\mathrm{R}(s, s^{\prime})=2.0$, $\forall s, s^{\prime} \in \mathbb{S}$. Note that $\Phi_{1}$ gives an imperfect feature set. The algorithm parameter values used are same as earlier. The results are shown in Fig. \ref{fig:starperf2}. In this case also, TD($0$) diverges. However, SCE-MSPBEM is seen to exhibit good stable behaviour.
\begin{figure}[!h]
	\hspace*{-4mm}
	\begin{subfigure}[h]{0.48\textwidth}
		\includegraphics[width=58mm, height=52mm]{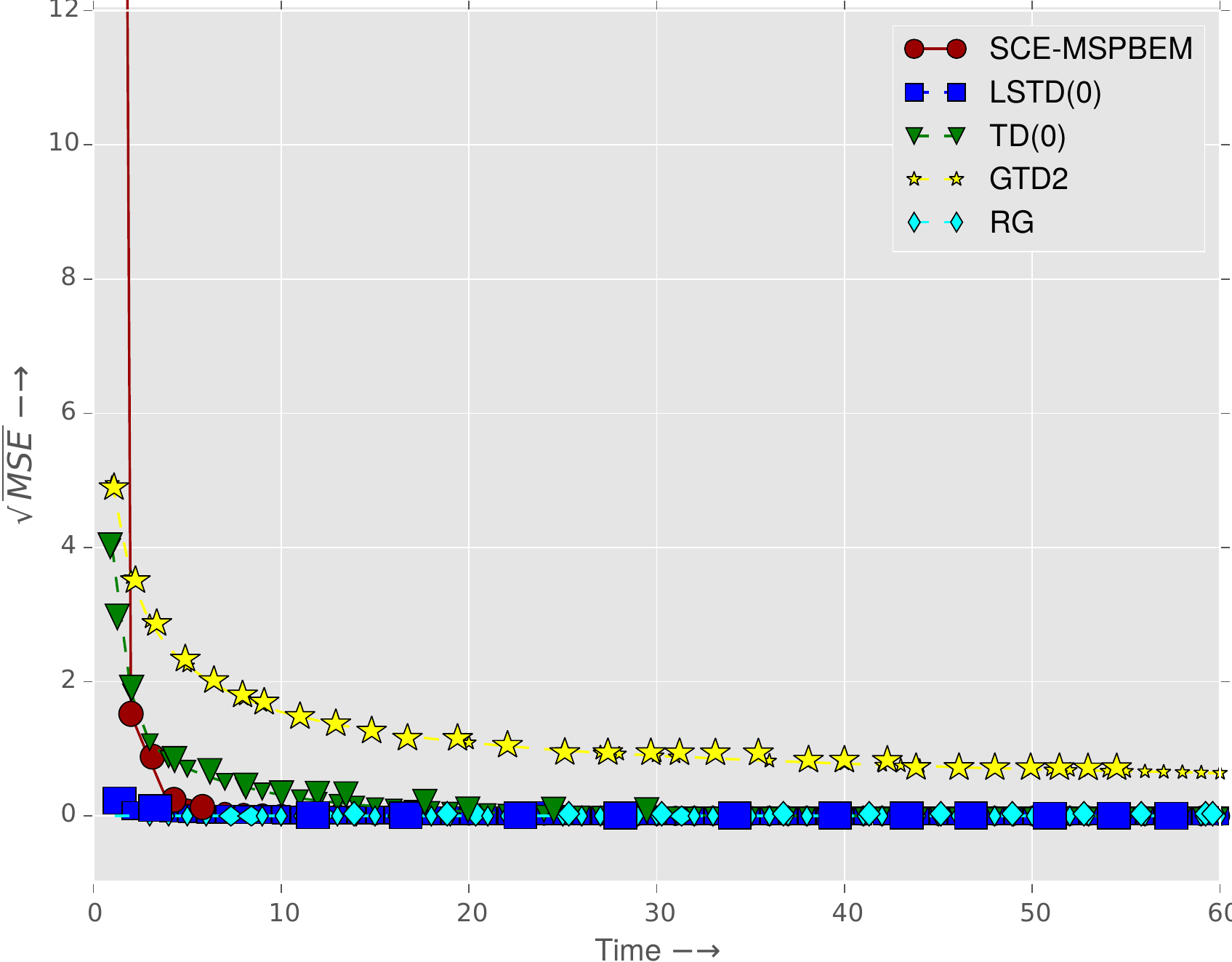}
		\subcaption{Discount factor $\gamma = 0.1$}
	\end{subfigure}\hspace*{1mm}
	\begin{subfigure}[h]{0.52\textwidth}
		\includegraphics[width=62mm, height=52mm]{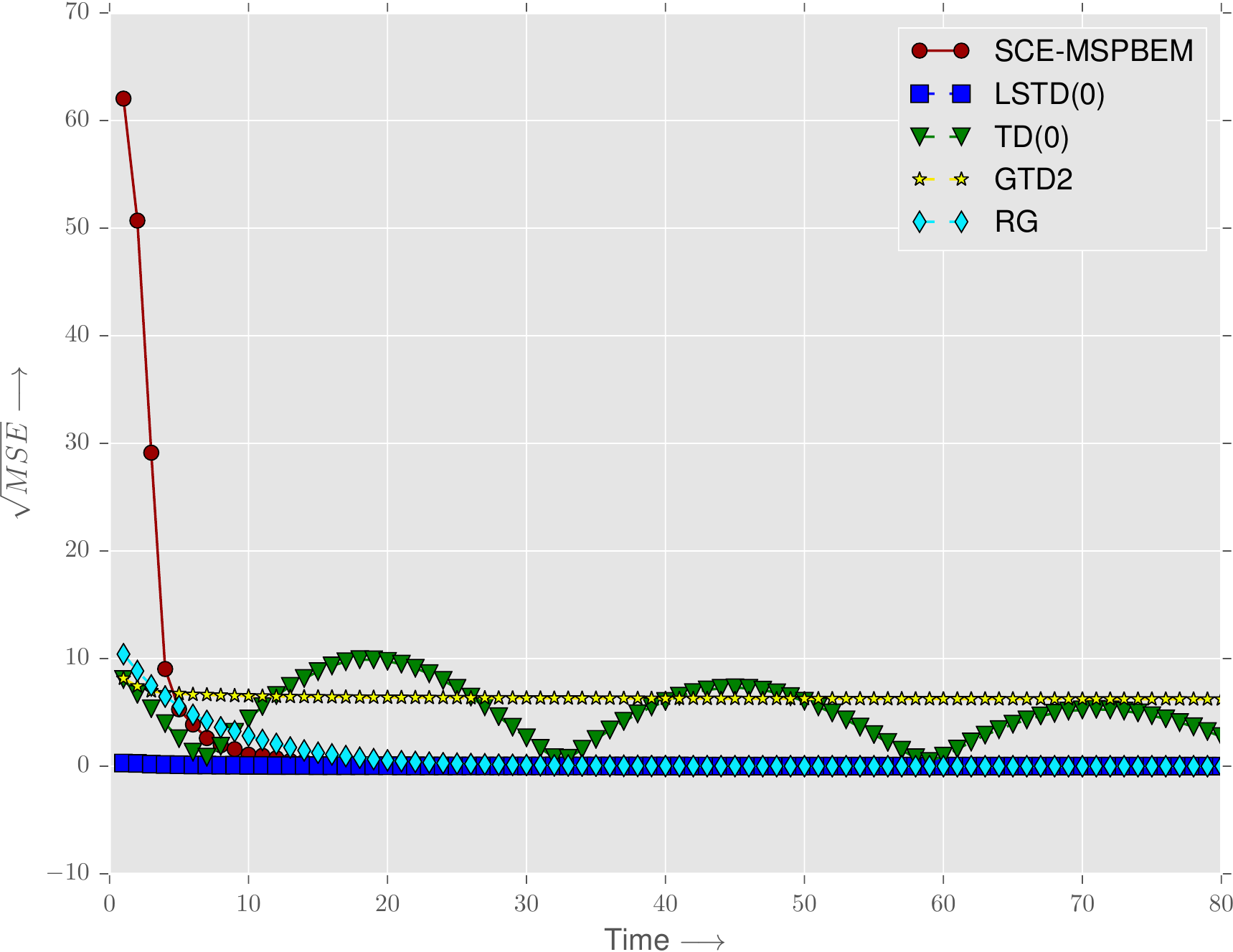}
		\subcaption{Discount factor $\gamma = 0.9$}
	\end{subfigure}%
	\caption{Baird's 7-Star MDP with perfect feature set. For $\gamma = 0.1$, all the algorithms converge with SCE-MSPBEM converging faster compared to a few algorithms, while being on par with the remaining algorithms. Note that this performance is obtained despite the fact that the initial value of SCE-MSPBEM is far from the solution compared to the rest of the algorithms. For $\gamma = 0.9$, TD(0) does not converge (which is in compliance with the observations made in \cite{schoknecht2002convergent}), while the performance of GTD$2$ is slow. However, SCE-MSPBEM exhibits good convergence behaviour which demonstrates the stable nature of the algorithm. }\label{fig:starperf}
\end{figure}
\begin{figure}[!h]
	\hspace*{-4mm}
	\begin{subfigure}[h]{0.53\textwidth}
		\includegraphics[width=62mm, height=52mm]{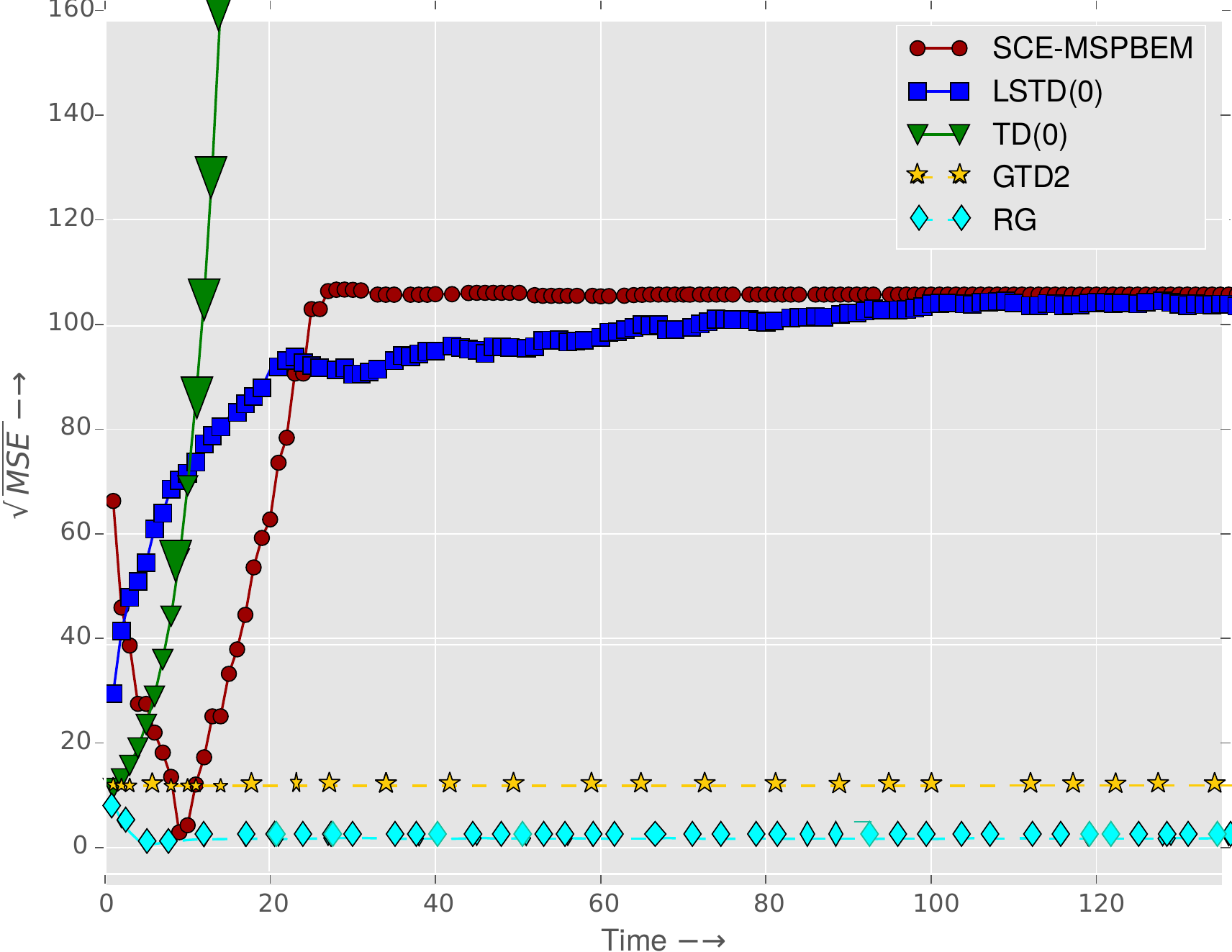}
		\subcaption{$\sqrt{\mathrm{MSE}}$.}
	\end{subfigure}\hspace*{1mm}
	\begin{subfigure}[h]{0.47\textwidth}
		\includegraphics[width=58mm, height=52mm]{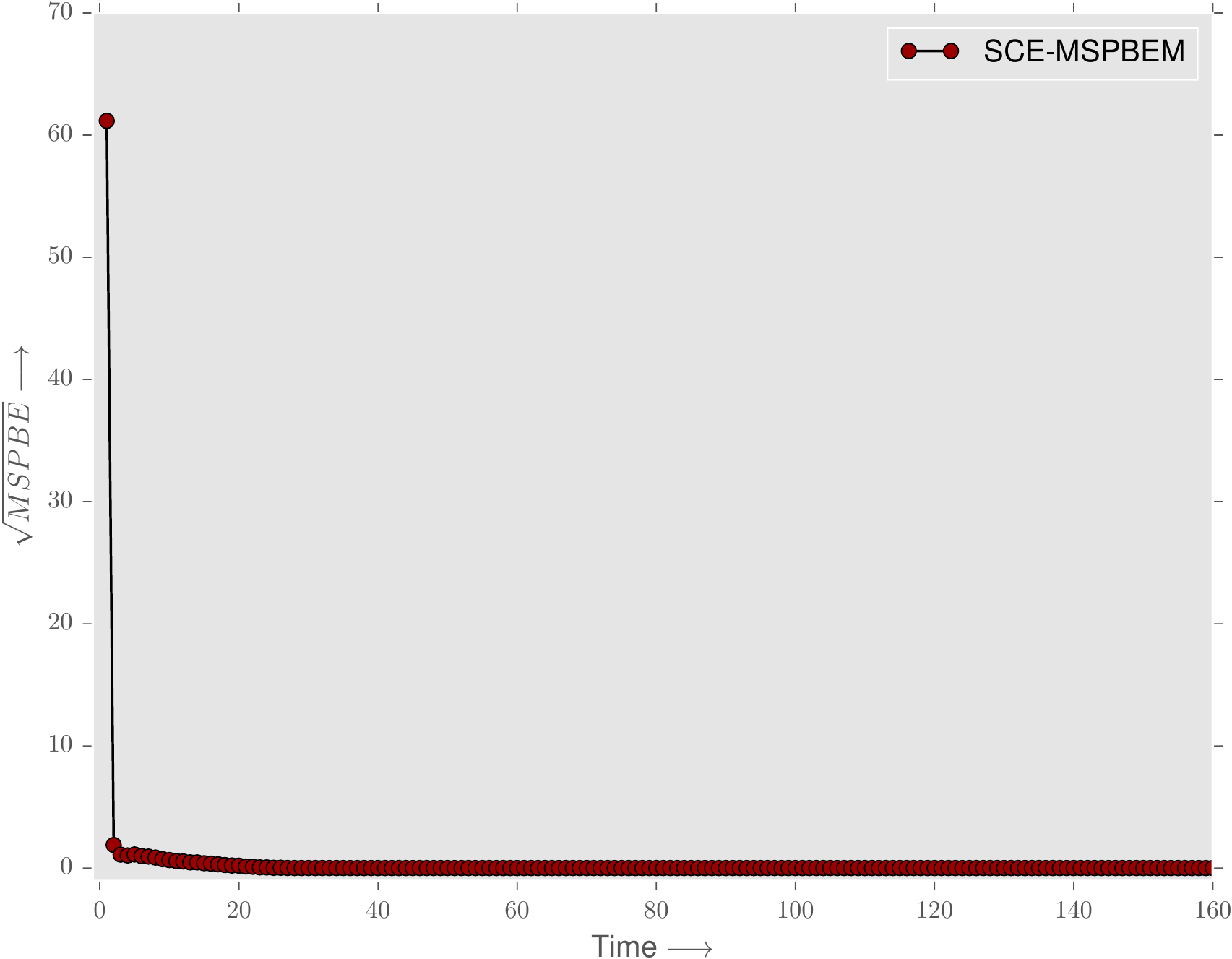}
		\subcaption{$\sqrt{\mathrm{MSPBE}}$.}
	\end{subfigure}%
	\caption{Baird's 7-Star MDP with imperfect feature set.	 Here the discount factor $\gamma = 0.99$. In this case, TD(0) diverges. However, $\sqrt{\mathrm{MSE}}$ of SCE-MSPBEM and LSTD($0$) converge to the same limit point (=$103.0$) with SCE-MSPBEM converging faster than LSTD. Also note that the RG method converges to a different limit (= $1.6919$). This is because the feature set is imperfect and also the fact that RG minimizes MSBR, while SCE-MSPBEM and LSTD minimize MSPBE. To verify this fact, note that in (b), $\sqrt{\mathrm{MSPBE}(\mu_{t})}$ of SCE-MSPBEM converges to $0$ which is indeed the minimum of MSPBE.}\label{fig:starperf2}
\end{figure}
\newpage
\subsection{Experiment 4: 10-State Ring MDP \cite{kveton2006solving}}\label{subsec:ringmdp}
Next, we studied the performance comparisons of the algorithms on a $10$-ring MDP with $\vert\mathbb{S}\vert = 10$ and $k = 8$. 
\begin{figure}[h]
	\begin{center}
		\includegraphics[scale=0.21]{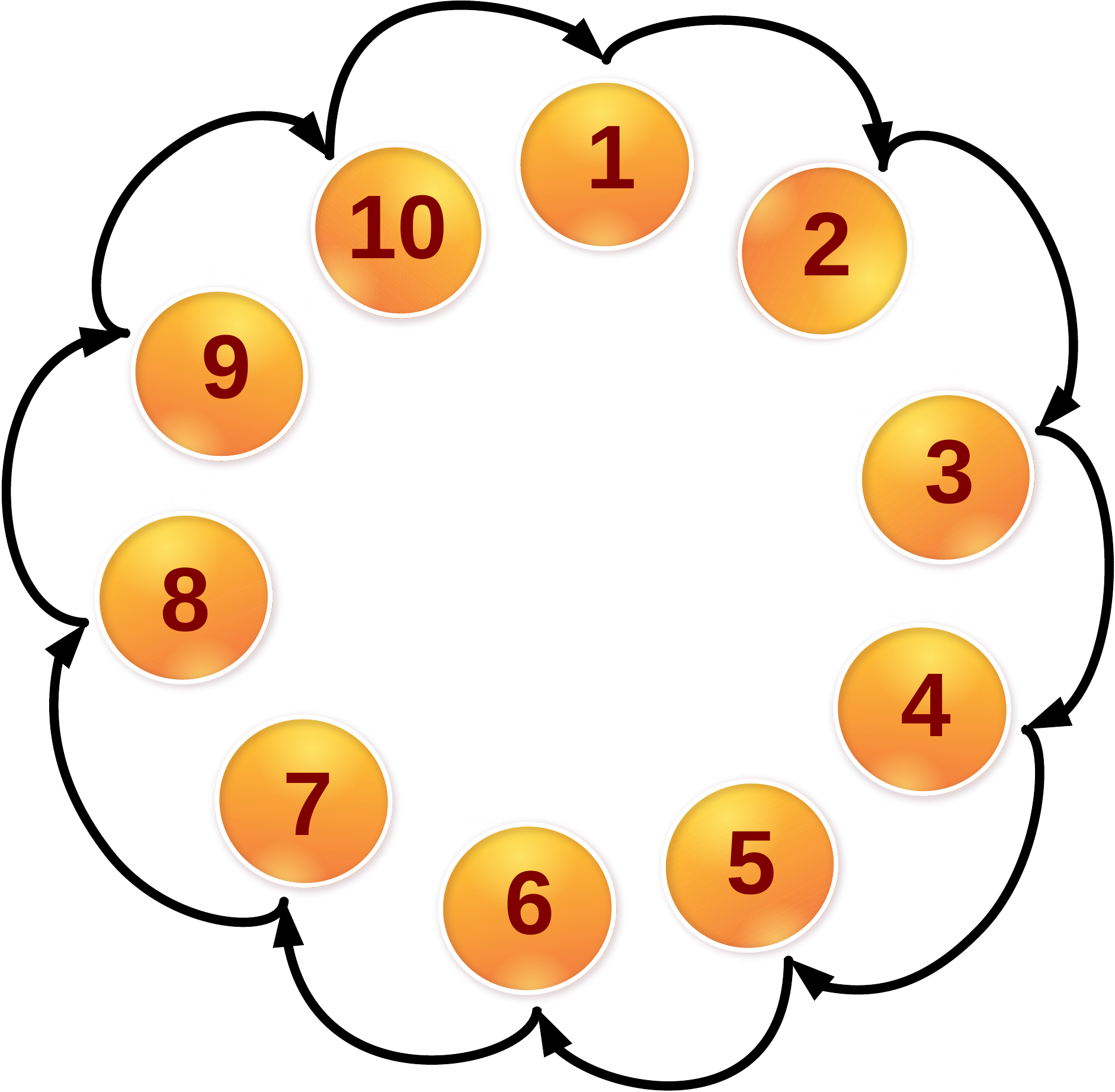}
	\end{center}
	\caption{10-Ring MDP}
	\vskip -5mm
\end{figure}
We let the sampling distribution $\nu$ to be the uniform distribution over $\mathbb{S}$. The transition matrix $\P_{\pi}$ and the feature matrix $\Phi$ are given by\vspace*{5mm}\\\\
{\setlength{\abovedisplayskip}{0pt}\setlength{\belowdisplayskip}{8pt}\begin{flalign}\label{eq:phiring}
\P_{\pi} = \begin{pmatrix}
	0 & 1 & 0 & 0 & 0 & 0 & 0 & 0 & 0 & 0 \\
	0 & 0 & 1 & 0 & 0 & 0 & 0 & 0 & 0 & 0 \\
	0 & 0 & 0 & 1 & 0 & 0 & 0 & 0 & 0 & 0 \\
	0 & 0 & 0 & 0 & 1 & 0 & 0 & 0 & 0 & 0 \\
	0 & 0 & 0 & 0 & 0 & 1 & 0 & 0 & 0 & 0 \\
	0 & 0 & 0 & 0 & 0 & 0 & 1 & 0 & 0 & 0 \\
	0 & 0 & 0 & 0 & 0 & 0 & 0 & 1 & 0 & 0 \\
	0 & 0 & 0 & 0 & 0 & 0 & 0 & 0 & 1 & 0 \\
	0 & 0 & 0 & 0 & 0 & 0 & 0 & 0 & 0 & 1 \\
	1 & 0 & 0 & 0 & 0 & 0 & 0 & 0 & 0 & 0 
	\end{pmatrix},
\hspace*{5mm}
\Phi = \begin{pmatrix}
	1 & 0 & 0 & 0 & 0 & 0 & 0 & 0 \\
	0 & 1 & 0 & 0 & 0 & 0 & 0 & 0 \\
	0 & 0 & 1 & 0 & 0 & 0 & 0 & 0 \\
	0 & 0 & 0 & 1 & 0 & 0 & 0 & 0 \\
	0 & 0 & 0 & 0 & 1 & 0 & 0 & 0 \\
	0 & 0 & 0 & 0 & 0 & 1 & 0 & 0 \\
	0 & 0 & 0 & 0 & 0 & 0 & 1 & 0 \\
	0 & 0 & 0 & 0 & 0 & 0 & 0 & 1 \\
	0 & 0 & 0 & 0 & 0 & 0 & 0 & 1 \\
	0 & 0 & 0 & 0 & 0 & 1 & 0 & 0
	\end{pmatrix}.
\end{flalign}}
The reward function is $\mathrm{R}(s, s^{\prime})=1.0, \forall s, s^{\prime} \in \mathbb{S}$.

The performance comparisons of the algorithms GTD$2$, TD($0$) and LSTD($0$) with SCE-MSPBEM are shown in Fig. \ref{fig:ringper}. The performance metric used here is the $\sqrt{\mathrm{MSE}(\cdot)}$ of the prediction vector generated by the corresponding algorithm at time $t$. The Markov chain in this case is ergodic and the uniform distribution over $\mathbb{S}$ is indeed the stationary distribution of the Markov chain. So theoretically all the algorithms should converge and the results in Fig. \ref{fig:ringper} confirm this. However, there is a significant difference in the rate of convergence of the various algorithms for large values of the discount factor $\gamma$. For $\gamma = 0.99$, the results show that GTD$2$ and RG trail behind other methods, while our method is only behind LSTD and outperforms TD($0$), RG and GTD$2$. The algorithm parameter values used in the experiment are provided in Table \ref{tab:10ring5} of Appendix.
\begin{figure}[!h]
	\begin{subfigure}[h]{0.5\textwidth}\hspace*{-4mm}
		\includegraphics[width=60mm, height=55mm]{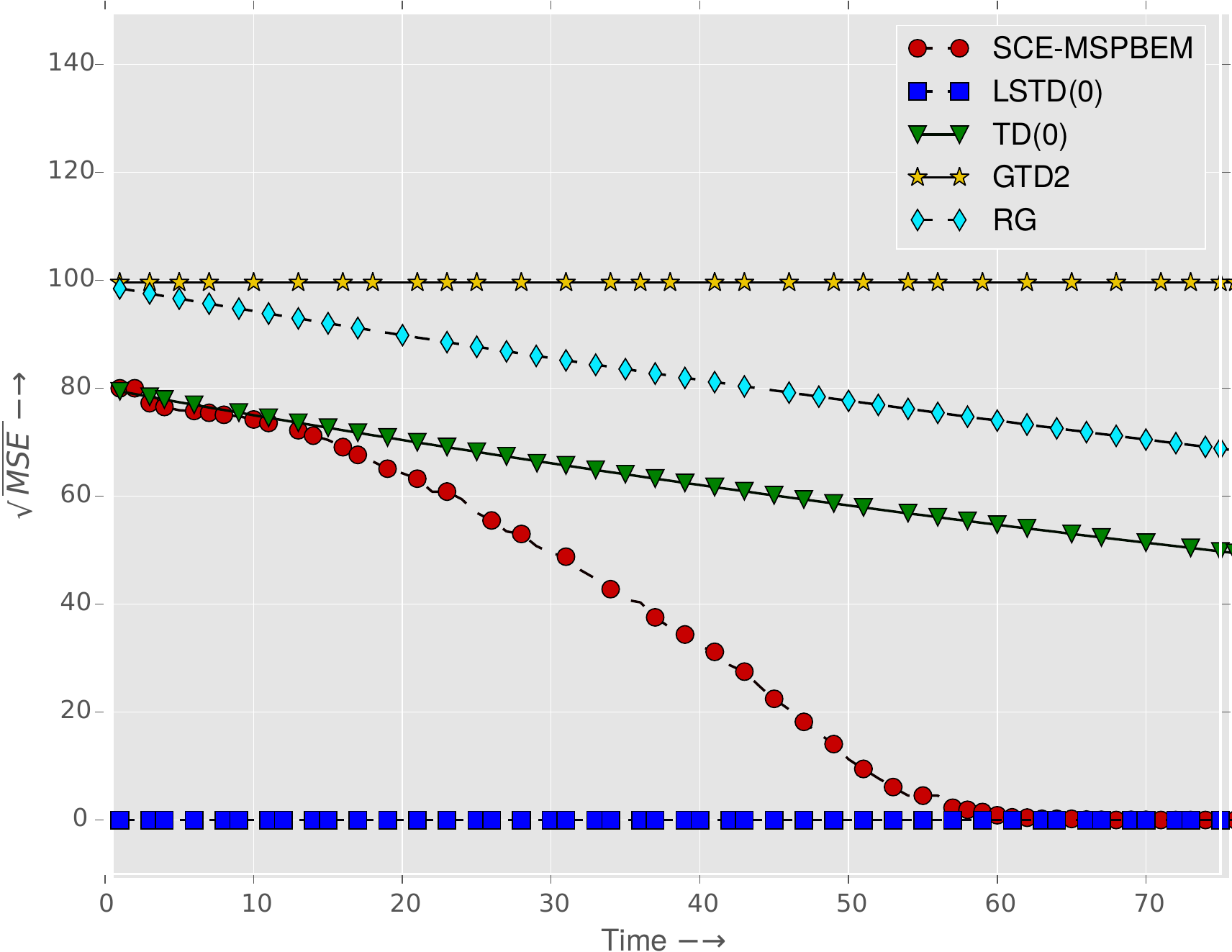}
		\subcaption{Discount factor $\gamma = 0.99$}
	\end{subfigure}\hspace*{1mm}
	\begin{subfigure}[h]{0.5\textwidth}
		\includegraphics[width=60mm, height=55mm]{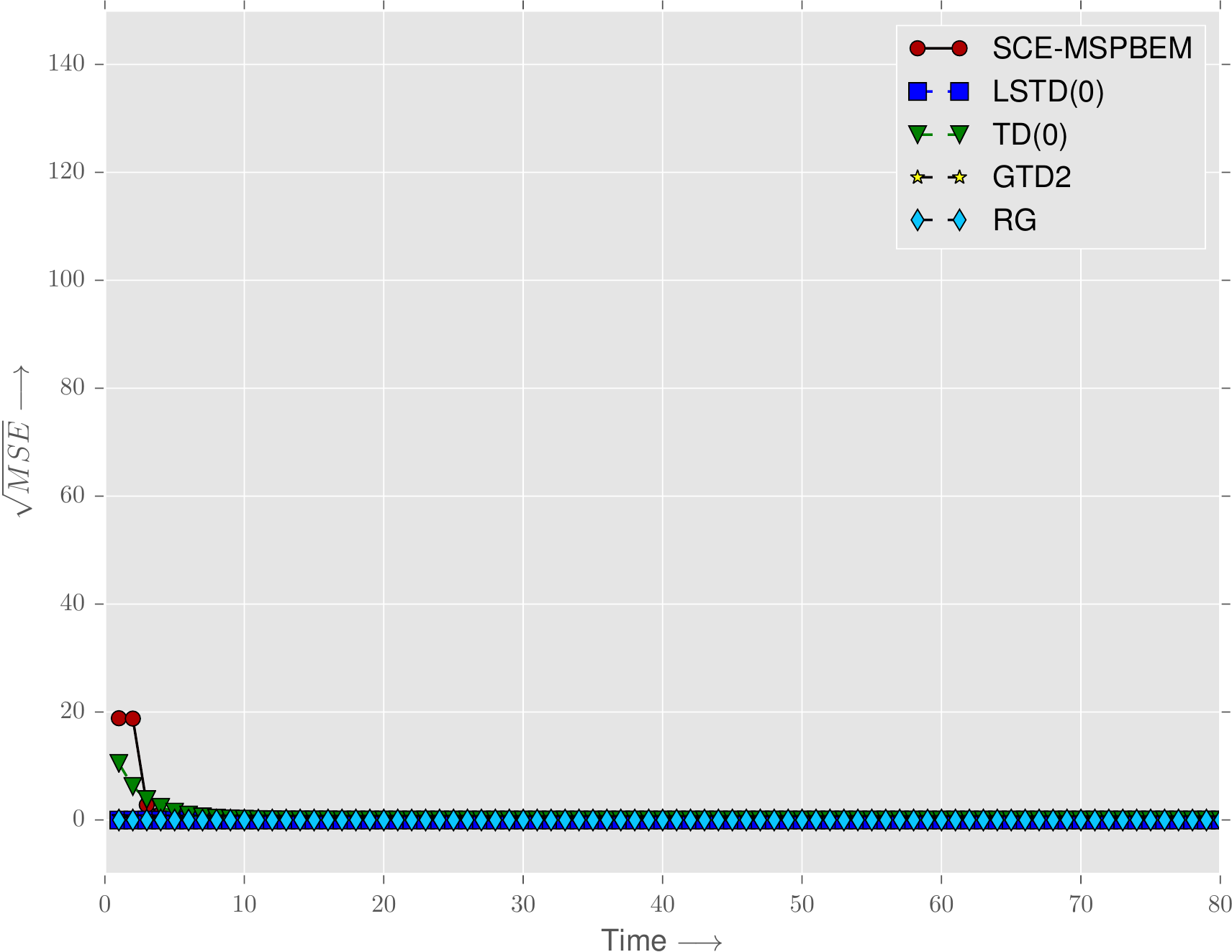}
		\subcaption{Discount factor $\gamma = 0.1$}
	\end{subfigure}%
	\caption{10-Ring MDP with perfect feature set: For $\gamma = 0.1$, all the algorithms exhibit almost the same rate of convergence. For $\gamma = 0.99$, SCE-MSPBEM converges faster than TD($0$), GTD$2$ and RG.}\label{fig:ringper}
\end{figure}

\subsection{Experiment 5: Random MDP with Radial Basis Functions and Fourier Basis}
These toy experiments are designed by us. Here, the tests are performed using standard basis functions to demonstrate that the algorithm is not dependent on any particular feature set. Two types of feature sets are considered here: Fourier basis functions and radial basis functions (RBF).\\\\
The Fourier basis functions \cite{konidaris2011value} are defined as follows:
\begin{eqnarray}
\phi_i(s) = \begin{cases}  1  &\mbox{if } i = 1, \\
\cos{\frac{(i+1)\pi s}{2}} & \mbox{if } i \mbox{ is odd}, \\
\sin{\frac{i\pi s}{2}} & \mbox{if } i \mbox{ is even}.
\end{cases}
\end{eqnarray}\\
	The radial basis functions are defined as follows:\vspace*{2mm}
\begin{equation}
	\phi_{i}(s) = \exp{\left(-\frac{(s-m_{i})^{2}}{2v_{i}^{2}}\right)},
\end{equation}
where $m_i \in \bbbr$ and $v_i \in \bbbr$ are fixed \emph{a priori}.\\

In both the cases, the reward function is given by
\begin{equation}\label{eqn:rwdrndchap5}
\mathrm{R}(s, s^{\prime}) = G(s)G(s^{\prime})\left(\frac{1}{(1.0+s^{\prime})^{0.25}}\right), \hspace*{1cm} \forall s, s^{\prime} \in \mathbb{S},
\end{equation}
where the vector $G \in (0,1)^{\vert \mathbb{S} \vert}$ is initialized for the algorithm with $G(s) \sim U(0,1), \forall s \in \mathbb{S}$.\\

Also in both the cases, the transition probability matrix $\P_{\pi}$ is  generated as follows:
\begin{equation}\label{eqn:prndchap5}
\P_{\pi}(s, s^{\prime}) =  {\vert \mathbb{S} \vert \choose s^{\prime}}b(s)^{s^{\prime}}(1.0-b(s))^{\vert \mathbb{S} \vert - s^{\prime}}, \hspace*{1cm} \forall s, s^{\prime} \in \mathbb{S},
\end{equation}
where the vector $b \in (0,1)^{\vert \mathbb{S} \vert}$ is initialized for the algorithm with $b(s) \sim U(0,1), \forall s \in \mathbb{S}$. It is easy to verify that the Markov chain defined by $\P_{\pi}$ is ergodic in nature.

In the case of RBF, we let $\vert\mathbb{S}\vert = 1000$, $\vert\mathbb{A}\vert = 200$, $k = 50$, $m_i = 10+20(i-1)$ and $v_i = 10$, while for Fourier basis functions, we let $\vert\mathbb{S}\vert = 1000$, $\vert\mathbb{A}\vert = 200$, $k = 50$. In both the cases, the distribution $\nu$ is the stationary distribution of the Markov chain. The simulation is run sufficiently long to ensure that the chain achieves its steady state behaviour, \emph{i.e.}, the states appear with the stationary distribution.
The algorithm parameter values used in the experiment are provided in Table \ref{tab:lrgmdppred5} of Appendix and the results obtained are provided in Figs. \ref{fig:fourier} and \ref{fig:rbfres}.

Also note that when Fourier basis is used, the discount factor $\gamma=0.9$ and for RBFs, $\gamma=0.01$. SCE-MSPBEM exhibits good convergence behaviour in both cases, which shows the non-dependence of SCE-MSPBEM on the discount factor $\gamma$. This is important because in \cite{schoknecht2003td}, the performance of TD methods is shown to be dependent on the discount factor $\gamma$.

\begin{figure}[h]
	\centering
	\includegraphics[height=70mm, width=105mm]{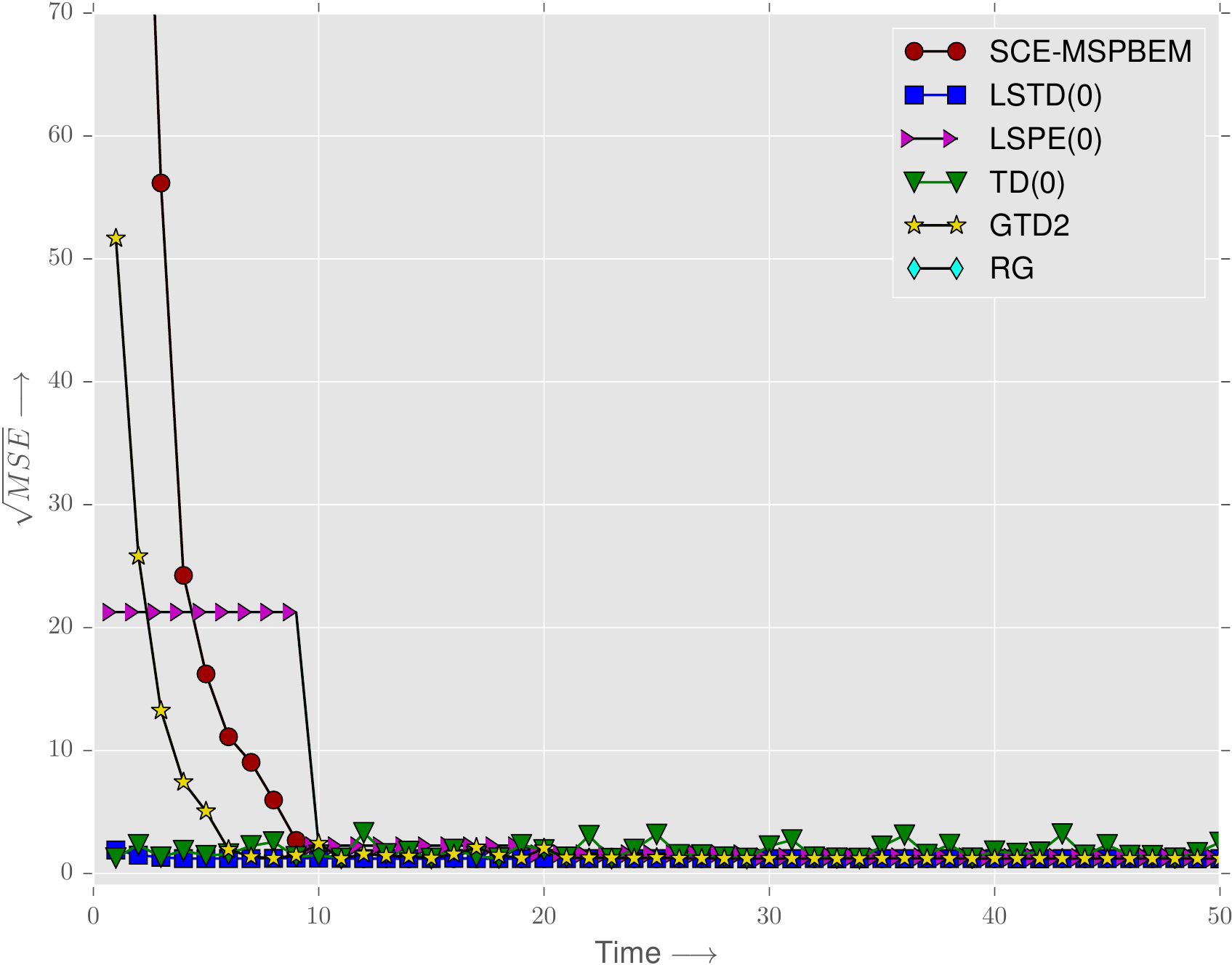}
	\caption{Fourier basis function: Here, $\vert\mathbb{S}\vert = 1000$, $\vert\mathbb{A}\vert = 200$, $k = 50$ and $\gamma = 0.9$. In this case, SCE-MSPBEM shows good convergence behaviour.}\label{fig:fourier}
\end{figure}
\begin{figure}[h]
	\centering
	\includegraphics[height=70mm, width=105mm]{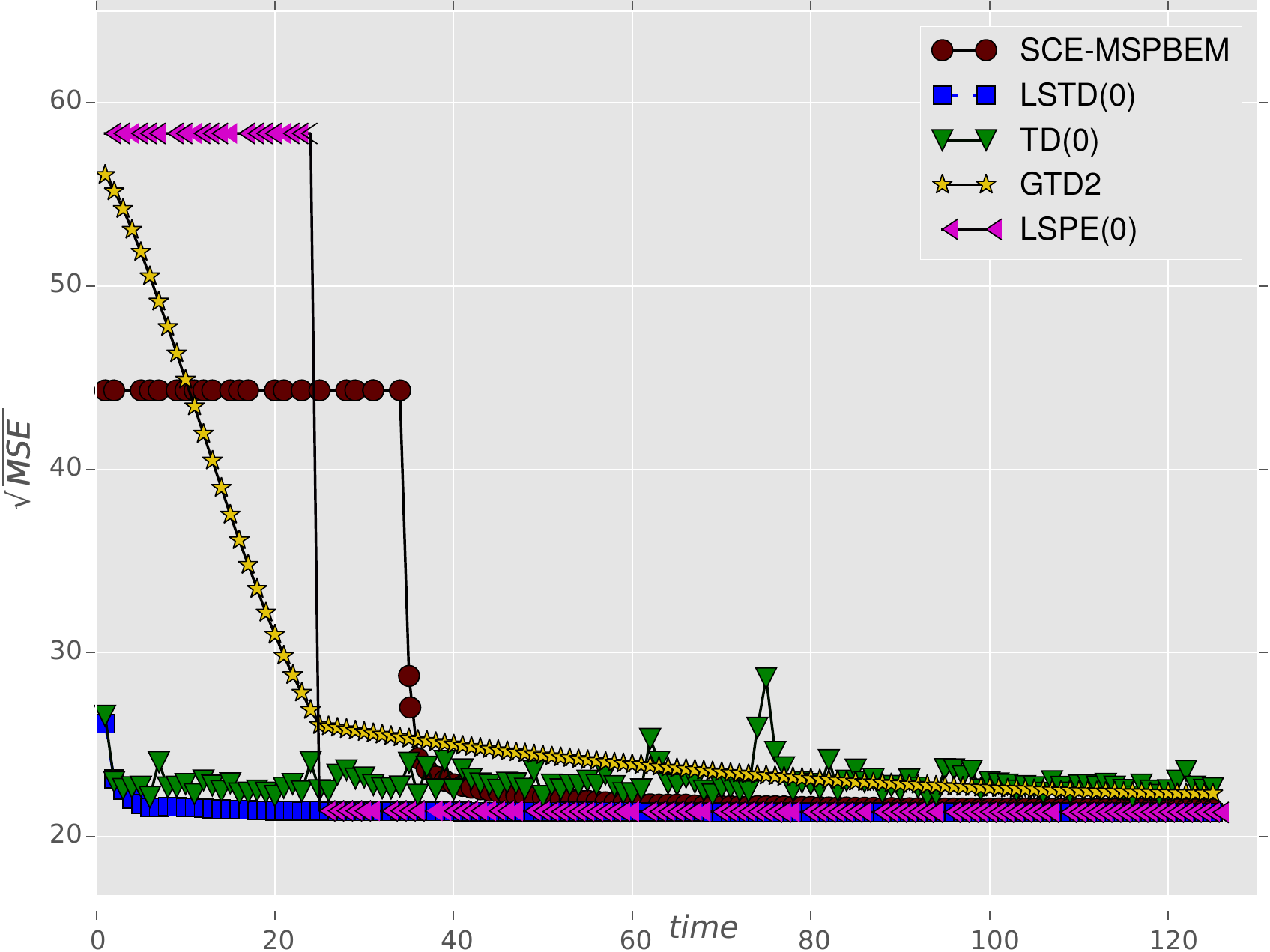}
	\caption{Radial basis function. Here, $\vert\mathbb{S}\vert = 1000$, $\vert\mathbb{A}\vert = 200$, $k = 50$ and $\gamma = 0.01$. In this case, SCE-MSPBEM converges to the same limit point as other algorithms.}\label{fig:rbfres}
\end{figure}

\vspace*{2mm}To measure how well our algorithm scales with respect to the size of the state space, we applied it on a medium sized MDP\footnote{This is the biggest MDP we could deploy on a machine with 3.2GHz processor  and 8GB of memory.}, where $|\mathbb{S}| = 2^{15}, |\mathbb{A}| = 50$, $k = 100$ and $\gamma=0.9$. This is the stress test. The reward function $\mathrm{R}$ and the transition probability matrix $\P_{\pi}$ are generated using Eqs. (\ref{eqn:rwdrndchap5}) and (\ref{eqn:prndchap5}) respectively. RBFs are used as the features in this case. Since the MDP is huge, the algorithms were run on Amazon cloud servers. The true value function $V^{\pi}$ was computed and the $\sqrt{\textrm{MSE}}$s of the prediction vectors generated by the different algorithms were compared. The performance results are shown in Table \ref{tab:comptable}. The results show that the performance of our algorithm does not seem affected by the complexity of the MDP.\\\\
\begin{table}[!h]
	\begin{center}
		\caption{Performance comparison of various algorithms on a medium sized MDP. Here $|\mathbb{S}| = 2^{15}, |\mathbb{A}| = 50, k = 100,$ and $\gamma = 0.9$. RBF is used as the feature set. The feature set is imperfect. The entries in the table correspond to the $\sqrt{\mathrm{MSE}}$ values obtained from the respective algorithms on $7$ different random MDPs. While the entries of SCE-MSPBEM, and LSTD($0$) appear to be similar, they actually differed in decimal digits that are not shown here for lack of space.}\label{tab:comptable}
		\begin{tabular}{|c | c | c | c | c | c |}
			\specialrule{.2em}{.04em}{.04em} 
			\textbf{Ex\#} & \textbf{SCE-MSPBEM} & \textbf{LSTD(0)} & \textbf{TD(0)} & \textbf{LSPE(0)} & \textbf{GTD$2$} \\
			\specialrule{.2em}{.04em}{.04em} 
			\noalign{\smallskip}
			1 & 23.339 &  23.339 & 24.581 & 23.354 & 24.932 \\ 
			2 & 23.142 &  23.142 & 24.372 & 23.178 & 24.755 \\ 
			3 & 23.332 &  23.332 & 24.537 & 23.446 & 24.881 \\ 
			4 & 22.978 &  22.978 & 24.194 & 22.987 & 24.532 \\  
			5 & 22.950 &  22.950 & 24.203 & 22.965 & 24.554 \\ 
			6 & 23.060 &  23.060 & 24.253 & 23.084 & 24.607 \\ 
			7 & 23.228 &  23.228 & 24.481 & 23.244 & 24.835 \\
			\specialrule{.2em}{.04em}{.04em} 
		\end{tabular}
	\end{center}
\end{table}
\newpage
\subsection{Experiment 6: Non-linear Function Approximation of Value Function\cite{tsitsiklis1997analysis}}
To demonstrate the flexibility and robustness of our approach, we also consider a few non-linear function approximation RL settings. The landscape in the non-linear setting is mostly non-convex and therefore multiple local optima exist. The stable non-linear function approximation extension of GTD$2$ is only shown to converge to the local optima \cite{bhatnagar2009convergent}. We believe that the non-linear setting offers the perfect scaffolding to demonstrate the global convergence property of our approach.
\subsubsection{Experiment 6.1: Van Roy and Tsitsiklis MDP \cite{tsitsiklis1997analysis}} This particular setting is designed in \cite{tsitsiklis1997analysis} to show the divergence of the standard TD($0$) algorithm in reinforcement learning under a non-linear approximation architecture. 
We consider here a discrete time Markov chain with state space $\mathbb{S} = \{1,2,3\}$, discount factor $\gamma = 0.9$, the reward function $\R(s, s^{\prime}) = 0, \forall s, s^{\prime} \in \mathbb{S}$ and the transition probability matrix as under:
{\setlength{\abovedisplayskip}{1pt}\setlength{\belowdisplayskip}{6pt}\begin{flalign*}
\mathrm{P} = \begin{bmatrix} 
1/2 && 0 && 1/2 \\
1/2 && 1/2 && 0 \\
0 && 1/2 && 1/2
\end{bmatrix}.
\end{flalign*}}
Note that the Markov chain is ergodic and hence the sample trajectory is obtained by following the dynamics of the Markov chain. Therefore the steady-state distribution of the Markov chain is indeed the sampling distribution. 
Here, we minimize the MSBR error function and it demands double sampling as prescribed in Assumption $(A3)^{\prime}$. The optimization is as follows:\\
\begin{gather}\label{eqn:nlopt}
\eta^{*} \in \argmin_{\eta \in \mathcal{Q} \subseteq \bbbr} \mathbb{E}\left[\mathbb{E}^{2}\left[\delta_{t}\right] \vert \mathbf{s}_{t}\right],
\end{gather}
where $\delta_{t} = \mathbf{r}_{t}+\gamma \psi_{\eta}(\mathbf{s}^{\prime}_{t})-\psi_{\eta}(\mathbf{s}_{t})$. We also have \\
\begin{flalign}\label{eqn:nlfapprpsi}
\psi_{\eta}(s) = (a(s)\cos{(\tau \eta)} - b(s)\sin{(\tau \eta)})e^{\epsilon \eta},
\end{flalign}
where $a = [100, -70, -30]^{\top}$, $b = [23.094, -98.15, 75.056]^{\top}$, $\tau = 0.01$ and $\epsilon = 0.001$. Here $\psi_{\eta}$ defines the projected non-linear manifold.  The true value function of this particular setting is $V = (0, 0, 0)^{\top}$. 

Now the challenge here is to best approximate the true value function $V$ using the family of non-linear functions $\psi_{\eta}$ parametrized by $\eta \in \bbbr$ by solving the optimization problem (\ref{eqn:nlopt}). It is easy to see that $\psi_{-\infty} = V$ and hence is a degenerate setting.\\

The objective function in Eq. (\ref{eqn:nlopt}) can be rearranged as
\begin{flalign*}
&\left[\upsilon_{11}, \upsilon_{21}, \omega_{31}\right]\left[1.0, 2.0e^{\epsilon \eta}\cos{(\tau\eta)}, -2.0e^{\epsilon \eta}\sin{(\tau\eta)}\right]^{\top} + \\
&\left[e^{\epsilon \eta}\cos{(\tau\eta)}, -e^{\epsilon \eta}\sin{(\tau\eta)}\right]\begin{bmatrix} 
\upsilon_{22} && \upsilon_{23} \\
\upsilon_{32} && \upsilon_{33} 
\end{bmatrix}\left[e^{\epsilon \eta}\cos{(\tau\eta)}, -e^{\epsilon \eta}\sin{(\tau\eta)}\right]^{\top},
\end{flalign*}
where  $\upsilon \in \bbbr^{3 \times 3}$ with  $\upsilon = (\upsilon_{ij})_{1 \leq i,j \leq 3} \triangleq \mathbb{E}[\mathit{h}_{t}]\mathbb{E}[\mathit{h}^{\prime}_{t}]^{\top}$. Here $\mathit{h}_{t} = [\mathbf{r}_{t}, a(\mathbf{s}^{\prime}_{t})-a(\mathbf{s}_{t}), b(\mathbf{s}^{\prime}_{t})-b(\mathbf{s}_{t})]^{\top}$ and $\mathit{h}^{\prime}_{t} = [\mathbf{r}^{\prime}_{t}, a(\mathbf{s}^{\prime\prime}_{t})-a(\mathbf{s}_{t}), b(\mathbf{s}^{\prime\prime}_{t})-b(\mathbf{s}_{t})]^{\top}$.\\

Now we maintain the time indexed random vector $\upsilon^{(t)} \in \bbbr^{3 \times 3}$   with  $\upsilon^{(t)} = (\upsilon^{(t)}_{ij})_{1 \leq i,j \leq 3}$ and employ the following recursion to track $\upsilon$:\\
\begin{equation}\label{eq:nlfnarecu}
	\upsilon^{(t+1)} = \upsilon^{(t)} + \alpha_{t+1}(\mathit{h}_{t} {\mathit{h}^{\prime}_{t}}^{\top} - \upsilon^{(t)}).
\end{equation}
Also, we define\\
\begin{flalign}\label{eqn:nlobjfuncce}
\bar{\mathcal{J}}_{b}&(\upsilon^{(t)}, \eta) = \left[\upsilon^{(t)}_{11}, \upsilon^{(t)}_{21}, \upsilon^{(t)}_{31}\right]\left[1.0, 2.0e^{\epsilon \eta}\cos{(\tau\eta)}, -2.0e^{\epsilon \eta}\sin{(\tau\eta)}\right]^{\top} + \nonumber\\
&\left[e^{\epsilon \eta}\cos{(\tau\eta)}, -e^{\epsilon \eta}\sin{(\tau\eta)}\right]\begin{bmatrix} 
\upsilon^{(t)}_{22} && \upsilon^{(t)}_{23} \\
\upsilon^{(t)}_{32} && \upsilon^{(t)}_{33} 
\end{bmatrix}\left[e^{\epsilon \eta}\cos{(\tau\eta)}, -e^{\epsilon \eta}\sin{(\tau\eta)}\right]^{\top}.
\end{flalign}

Now we solve the optimization problem (\ref{eqn:nlopt}) using Algorithm \ref{algo:sce-msbr} with the objective function defined in Eq.  (\ref{eqn:nlobjfuncce}) (\emph{i.e.}, using Eq. (\ref{eq:nlfnarecu}) instead of Eq. (\ref{eqn:bralupsupd}) and Eq. (\ref{eqn:nlobjfuncce}) instead of Eq. (\ref{eqn:brjval}) respectively).

The various parameter values used in the experiment are provided in Table \ref{tab:nlfaparvals} of Appendix.
The results of the experiment are shown in Fig. \ref{fig:linkres}. The $x$-axis is the iteration number $t$. The performance measure considered here is the mean squared error (MSE) which is defined in Eq. (\ref{eq:msedef}). Algorithm \ref{algo:sce-msbr} is seen to clearly outperform TD($0$) and GTD2 here.
\begin{figure}[h]
	\centering
	{\hspace*{2mm}\includegraphics[height=60mm, width=90mm]{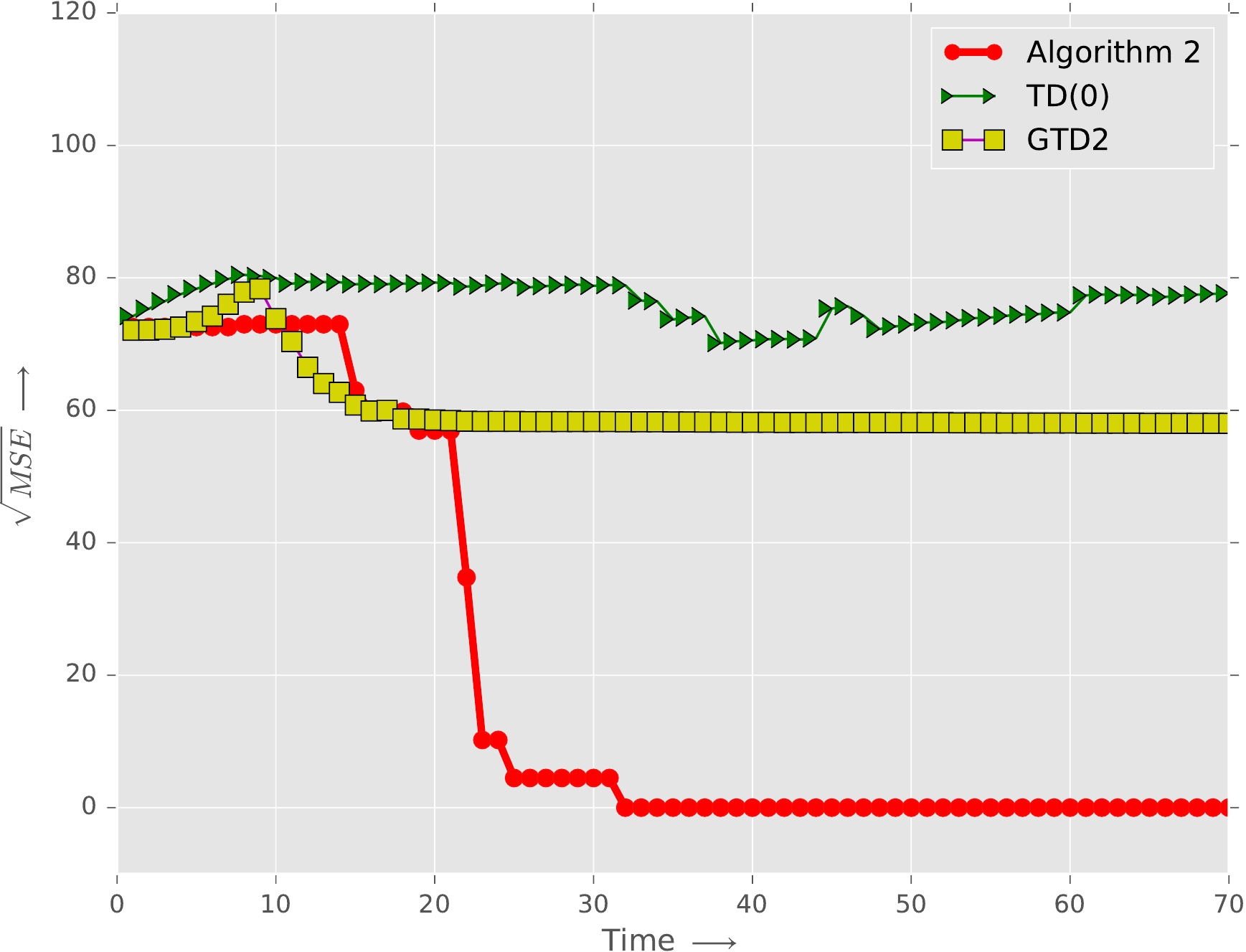}\hspace*{4mm}}
	\caption{Non-linear function approximation on Van Roy and Tsitsiklis MDP. The  plot shows the trajectory of $\sqrt{\mathrm{MSE}}$ generated by TD($0$), GTD$2$ and our algorithm against the iteration number $t$. Algorithm \ref{algo:sce-msbr} (SCE-MSBRM) predicts the true value function $V$ which follows from the observation that $\sqrt{\mathrm{MSE}}$ converges to $0$. TD($0$) slowly diverges, while GTD$2$ converges to a sub-optimal solution. This experiment further demonstrates the effectiveness of our proposed scheme to non-convex settings.}\label{fig:linkres}
\end{figure}%

\noindent
\subsubsection{Experiment 6.2: Baird's $7$-star MDP using Non-Linear Function Approximation}
Here, we consider the Baird's $7$-star MDP defined in Section \ref{subsec:bairdmdp} with discount factor $\gamma = 0.9$, $k=8$ and the sampling distribution to be the uniform distribution over $\mathbb{S}$. To perform the non-linear function approximation, we consider the non-linear manifold given by $\{\Phi h(z) \vert z \in \bbbr^{8}\}$, where $h(z) \triangleq (\cos^{2}{(z_1)}\exp{(0.01z_1)}, \cos^{2}{(z_2)}\exp{(0.01z_2)}, \dots, \cos^{2}{(z_8)}\exp{(0.01z_8)})^{\top}$ and $\Phi$ is defined in Eq. (\ref{eq:phibairds}). The reward function is given by $\R(s,s^{\prime}) = 0, \forall s,s^{\prime} \in \mathbb{S}$ and hence the true value function is $(0, 0, \dots, 0)^{\top}_{7 \times 1}$. Due to the unique nature of the non-linear manifold, one can directly apply SCE-MSBRM (Algorithm \ref{algo:sce-msbr}) with $h(z)$ replacing $z$ in Eq. (\ref{eqn:brjval}).  This setting presents a hard and challenging task for TD($\lambda$) since  we already experienced the erratic and unstable behaviour of TD($\lambda$) in the linear function approximation version of Baird's $7$-star. This setting also proves to be a litmus test for determining the confines of the stable characteristic of the non-linear function approximation version of GTD2. The results obtained are provided in Fig. \ref{fig:bairdnl}.  The various parameter values used in the experiment are provided in Table \ref{tab:bairdnlfaparvals} of Appendix. It can be seen that whereas both TD($0$) and GTD2 diverge here, SCE-MSBRM is converging to the true value.
\begin{figure}[h]
	\centering
	{\hspace*{2mm}\includegraphics[height=55mm, width=90mm]{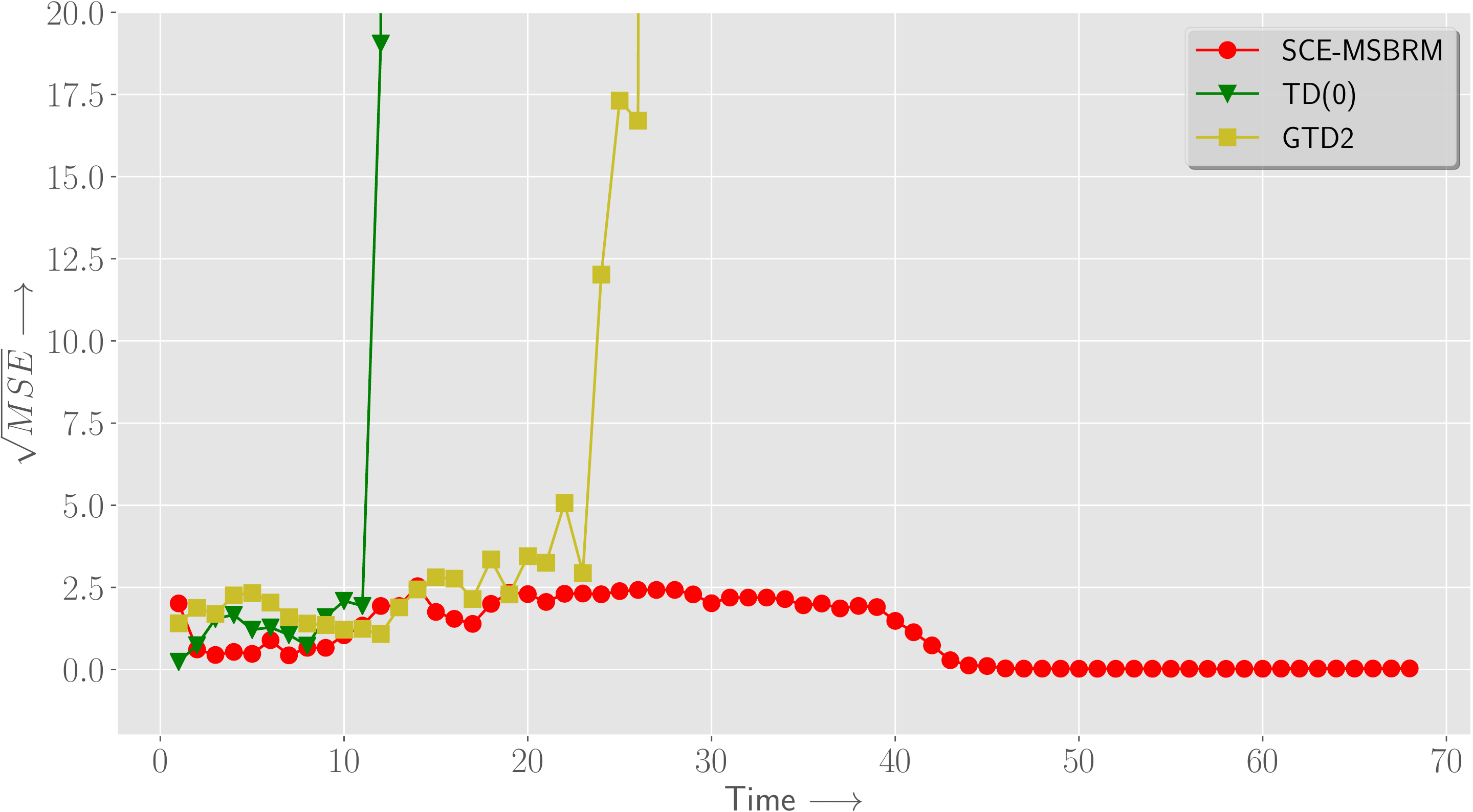}\hspace*{4mm}}
	\caption{Non-linear function approximation on Baird's $7$-star MDP: The  plot shows the trajectory of $\sqrt{\mathrm{MSE}}$ generated by TD($0$), GTD$2$ and SCE-MSBRM against $t/100$, where $t$ is the iteration number. SCE-MSBRM predicts the true value function $V$ which follows from the observation that $\sqrt{\mathrm{MSE}}$ converges to $0$. Note that both TD($0$) and GTD$2$ diverge. The divergence of TD($0$) is expected since it also diverges in the linear case (see Fig. \ref{fig:starperf}). However, the divergence of the stable non-linear GTD2 is not unforeseen, but subjective. The rationale behind this erratic behaviour is ascribed to the absence of the projection ($\Pi_{C}$ defined in Algorithm \ref{algo:gtd2}) of the iterates in the experiment conducted. The projection operator $\Pi_{C}$ is necessary to ensure the stability of the algorithm, however its computation is hard and hence the omission.}\label{fig:bairdnl}
\end{figure}%
\subsubsection{Experiment 6.3:  $10$-ring MDP using Non-linear Function Approximation}
Here, we consider the $10$-ring MDP defined in Section \ref{subsec:ringmdp} with discount factor $\gamma = 0.99$, $k=8$ and the sampling distribution as the stationary distribution of the underlying Markov chain.
Here, we consider the non-linear manifold given by $\{\Phi h(z) \vert z \in \bbbr^{8}\}$, where $h(z) \triangleq (\cos^{2}{(z_1)}\exp{(0.1z_1)}, \cos^{2}{(z_2)}\exp{(0.1z_2)}, \dots,$ $\cos^{2}{(z_8)}\exp{(0.1z_8)})^{\top}$ and $\Phi$ is defined in Eq. (\ref{eq:phiring}). The reward function is given by $\R(s,s^{\prime}) = 0, \forall s,s^{\prime} \in \mathbb{S}$ and hence the true value function is $(0, 0, \dots, 0)^{\top}_{10 \times 1}$. Similar to the previous experiment, here also one can directly apply SCE-MSBRM (Algorithm \ref{algo:sce-msbr}) with $h(z)$ replacing $z$ in Eq. (\ref{eqn:brjval}).   The results obtained are provided in Fig. \ref{fig:ringnl}.  The various parameter values we used are provided in Table \ref{tab:ringnlfaparvals} of Appendix. GTD2 does not converge to the true value here while both SCE-MSBRM and TD($0$) do, with TD($0$) marginally better.
\begin{figure}[h]
	\centering
	{\hspace*{2mm}\includegraphics[height=55mm, width=90mm]{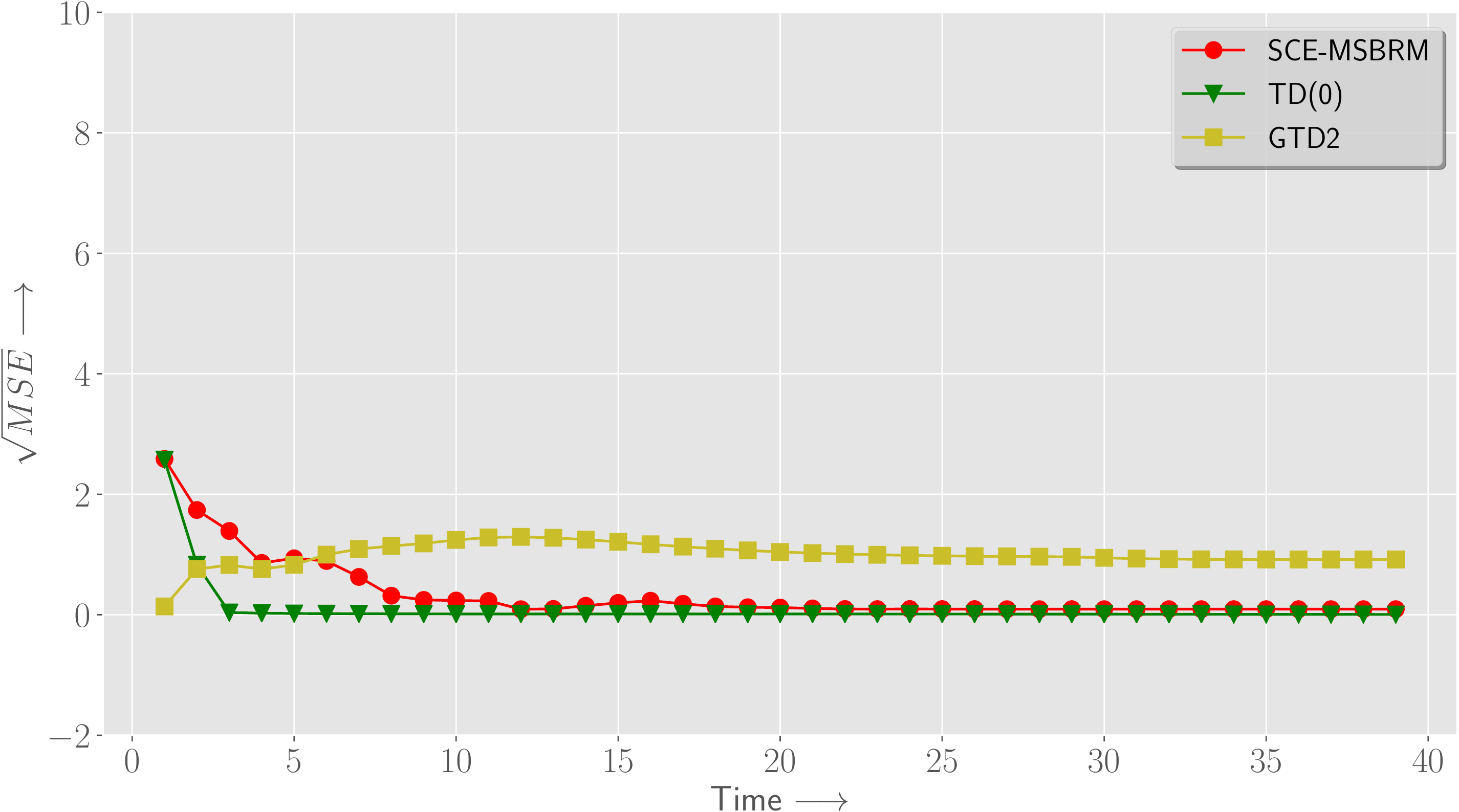}\hspace*{4mm}}
	\caption{Non-linear function approximation on $10$-ring MDP: The  plot shows the trajectory of $\sqrt{\mathrm{MSE}}$ generated by TD($0$), GTD$2$ and SCE-MSBRM against $t/150$, where $t$ is the iteration number. SCE-MSBRM predicts the true value function $V$ which follows from the observation that $\sqrt{\mathrm{MSE}}$ converges to $0$. Note that TD($0$) also converges to the true value function while GTD$2$ could only find sub-optimal solution here.}\label{fig:ringnl}
\end{figure}%
\section{Conclusion}
We proposed, for the first time, an application of the cross entropy (CE) method to the prediction problem in reinforcement learning (RL) under the linear function approximation architecture. This task is accomplished by employing the multi-timescale stochastic approximation variant of the cross entropy optimization method to minimize the mean squared projected Bellman error (MSPBE) and mean squared Bellman error (MSBR) objectives. The proofs of convergence of the algorithms to the optimum values using the ODE method are also provided. The theoretical analysis is supplemented by extensive experimental evaluation which is shown to corroborate the claims. Experimental comparisons with the state-of-the-art algorithms show the superiority in terms of stability and accuracy while being competitive enough with regard to computational efficiency and rate of convergence. As future work, one may design similar cross entropy approaches for both prediction and control problems. More numerical experiments involving other non-linear function approximators, delayed rewards \emph{etc.} may be tried as well.

%
%
%
\clearpage
\begin{appendices}
\section{Linear Function Approximation (LFA) based Prediction Algorithms}
\vspace*{-10mm}
\fbox{\begin{minipage}[t]{55mm}
\null 
\begin{algorithm}[H]
\vspace*{1mm}
\hspace*{-4mm}$\delta_{t} = \mathbf{r}_{t} + \gamma \mathbf{z}_{t}^{\top}\phi(\mathbf{s}^{\prime}_{t}) - \mathbf{z}_{t}^{\top}\phi(\mathbf{s}_{t})$;\\
\hspace*{-4mm}$\mathbf{e}_{t+1} = \phi(\mathbf{s}_t) + \gamma\lambda \mathbf{e}_{t}$;\\
\hspace*{-4mm}$\mathbf{z}_{t+1} = \mathbf{z}_{t} + \alpha_{t+1}\delta_{t}\mathbf{e}_{t+1}$.\vspace*{4mm}\\
\caption{TD($\lambda$) LFA\label{algo:tdlbdalfa}}
\end{algorithm}
	$\alpha_{t} > 0$ satisfies $\sum_{t=1}^{\infty}\alpha_{t} = \infty$, $\sum_{t=1}^{\infty}\alpha^{2}_{t} < \infty$.
\end{minipage}}\vspace*{4mm}
\fbox{\begin{minipage}[t]{62mm}
\null
\begin{algorithm}[H]
\vspace*{1mm}
	\hspace*{-4mm}$\delta_{t} = \mathbf{r}_{t} + \gamma \mathbf{z}_{t}^{\top}\phi(\mathbf{s}^{\prime}_{t}) - \mathbf{z}_{t}^{\top}\phi(\mathbf{s}_{t})$;\\
	\hspace*{-4mm}$\mathbf{z}_{t+1} = \mathbf{z}_{t} + \alpha_{t+1}\delta_{t}\left(\phi(\mathbf{s}_{t})-\gamma\phi(\mathbf{s}^{\prime}_{t+1})\right)$;\vspace*{4mm}\\
\caption{RG LFA}\label{algo:resgd}
\end{algorithm}
$\alpha_{t} > 0$ satisfies $\sum_{t=1}^{\infty}\alpha_{t} = \infty$, $\sum_{t=1}^{\infty}\alpha^{2}_{t} < \infty$.
\end{minipage}}\vspace*{2mm}\\\\
\fbox{
\begin{minipage}[t]{102mm}
\null
\begin{algorithm}[H]
\vspace*{1mm}
	$\delta_{t} = \mathbf{r}_{t} + \gamma \mathbf{z}_{t}^{\top}\phi(\mathbf{s}^{\prime}_{t}) - \mathbf{z}_{t}^{\top}\phi(\mathbf{s}_{t})$;\\
	$\mathbf{z}_{t+1} = \mathbf{z}_{t} + \alpha_{t+1}\left(\phi(\mathbf{s}_{t}) - \gamma \phi(\mathbf{s}^{\prime}_{t})\right)(\phi(\mathbf{s}_t)^{\top}\mathbf{v}_{t})$;\\
$\mathbf{v}_{t+1} = \mathbf{v}_{t} + \beta_{t+1}(\delta_{t}- \phi(\mathbf{s}_t)^{\top}\mathbf{v}_{t})\phi(\mathbf{s}_t)$;
\vspace*{1mm}
\caption{GTD2 LFA}\label{algo:gtd2}
\end{algorithm}
$\alpha_{t}, \beta_{t} > 0$ satisfy $\sum_{t=1}^{\infty}\alpha_{t} = \infty$, $\sum_{t=1}^{\infty}\alpha^{2}_{t} < \infty$ and 
	$\beta_{t} = \eta \alpha_{t}$, where $\eta > 0$.
\end{minipage}}\vspace*{2mm}\\\\
\fbox{
\begin{minipage}[t]{102mm}
\null
\begin{algorithm}[H]
$\delta_{t} = \mathbf{r}_{t} + \gamma \mathbf{z}_{t}^{\top}\phi(\mathbf{s}^{\prime}_{t}) - \mathbf{z}_{t}^{\top}\phi(\mathbf{s}_{t})$;\\
$\mathbf{z}_{t+1} = \mathbf{z}_{t} + \alpha_{t+1}\delta_{t}\phi(\mathbf{s}_{t}) - \gamma \phi(\mathbf{s}^{\prime}_{t})(\phi(\mathbf{s}_t)^{\top}\mathbf{v}_{t})$;\\
$\mathbf{v}_{t+1} = \mathbf{v}_{t} + \beta_{t+1}(\delta_{t}- \phi(\mathbf{s}_t)^{\top}\mathbf{v}_{t})\phi(\mathbf{s}_t)$;
\caption{TDC LFA}\label{algo:tdc}
\end{algorithm}
	$\alpha_{t}, \beta_{t} > 0$ satisfy $\sum_{t=1}^{\infty}\alpha_{t} = \sum_{t=1}^{\infty}\beta_{t} = \infty$, $\sum_{t=1}^{\infty}\left(\alpha^{2}_{t} + \beta^{2}_{t}\right) < \infty$ and 
	$\frac{\alpha_{t}}{\beta_{t}} \rightarrow 0$.
\end{minipage}}\vspace*{2mm}\\\\
\fbox{\begin{minipage}[t]{102mm}
\null
\begin{algorithm}[H]
$\mathbf{A}_0 = \epsilon \mathbb{I}_{k_1 \times k_1}$, $\epsilon > 0$, $t = 0$, $\mathbf{b}_0 = \mathbf{e}_{0} = 0_{k \times 1}$;\\
\While{stopping criteria not satisfied}{
$\mathbf{e}_{t+1} = \phi(\mathbf{s}_t) + \gamma\lambda \mathbf{e}_{t}$;\\
	$\mathbf{A}_{t+1} = \mathbf{A}_{t} + \mathbf{e}_{t+1}\left(\phi(\mathbf{s}_{t}) - \gamma\phi(\mathbf{s}^{\prime}_{t})\right)^{\top}$;\\
$\mathbf{b}_{t+1} = \mathbf{b}_{t} + \mathbf{e}_{t+1}\mathbf{r}_{t}$;}
\Return $\mathbf{A}_{T}^{-1}\mathbf{b}_{T}$;
\caption{LSTD($\lambda$) LFA}\label{algo:lstdlbda}
\end{algorithm}
\end{minipage}}\vspace*{2mm}\\\\
\fbox{\begin{minipage}[t]{102mm}
\null
\begin{algorithm}[H]
$\mathbf{A}_0 = \mathbf{B}_0 = \epsilon \I_{k \times k}$, $\epsilon > 0$, $t = 0$, $\mathbf{b}_{0} = \mathbf{e}_{0} = 0_{k \times 1}$;\\
\While{stopping criteria not satisfied}{
$\mathbf{B}_{t+1} = \mathbf{B}_{t} + \phi(\mathbf{s}_{t})\phi(\mathbf{s}_{t})^{\top}$;\\
$\mathbf{e}_{t+1} = \phi(\mathbf{s}_t) + \gamma\lambda \mathbf{e}_{t}$;\\
	$\mathbf{A}_{t+1} = \mathbf{A}_{t} + \mathbf{e}_{t+1}\left(\phi(\mathbf{s}_{t}) - \gamma\phi(\mathbf{s}^{\prime}_{t})\right)^{\top}$;\\
$\mathbf{b}_{t+1} = \mathbf{b}_{t} + \mathbf{e}_{t+1}\mathbf{r}_{t}$;}
\Return $\mathbf{A}_{T}^{-1}\mathbf{B}_{T}\mathbf{b}_{T}$;
\caption{LSPE($\lambda$) LFA}\label{algo:lspelbda}
\end{algorithm}
\end{minipage}}\vspace*{2mm}\\
\section{Non-Linear Function Approximation (NLFA) based Prediction Algorithms}
\fbox{
\begin{minipage}[t]{102mm}
\null
\begin{algorithm}[H]
\vspace*{1mm}
$\delta_{t} = \mathbf{r}_{t} + \gamma V_{\theta_t}(\mathbf{s}^{\prime}_{t}) - V_{\theta_t}(\mathbf{s}_{t})$;\\
$\mathbf{w}_{t+1} = \mathbf{w}_{t} + \beta_{t+1}\left(\delta_{t} - \nabla V_{\theta_t}(\mathbf{s}_{t})^{\top}\mathbf{w}_{t}\right)\nabla V_{\theta_t}(\mathbf{s}_{t})$;\\
$h_{t} = (\delta_{t} - \nabla V_{\theta_t}(\mathbf{s}_{t})^{\top}\mathbf{w}_{t})\nabla^{2}V_{\theta_t}(\mathbf{s}_{t})\mathbf{w}_{t}$;\\
$\theta_{t+1} = \Pi_{C}\Big(\theta_{t} + \alpha_{t+1}\{\delta_{t}\nabla V_{\theta_t}(\mathbf{s}_{t}) - $\\\hspace*{25mm}$\gamma \nabla V_{\theta_t}(\mathbf{s}^{\prime}_{t})(\nabla V_{\theta_t}(\mathbf{s}_{t})^{\top}\mathbf{w}_{t}) - h_{t+1}\}\Big)$;\\
\vspace*{1mm}
\caption{GTD2 NLFA}\label{algo:gtd2}
\end{algorithm}
$\alpha_{t}, \beta_{t} > 0$ satisfy $\sum_{t=1}^{\infty}\alpha_{t} = \infty$, $\sum_{t=1}^{\infty}\alpha^{2}_{t} < \infty$ and 
$\frac{\alpha_{t}}{\beta_{t}} \rightarrow 0$.\\
	Here $\{V_{\theta} \in \bbbr^{\vert \mathbb{S} \vert} \vert \theta \in \bbbr^{n}\}$ is the differentiable sub-manifold of $\bbbr^{\vert \mathbb{S} \vert}$.\\
$C$ is a predetermined compact subset of $\bbbr^{n}$ and $\Pi_{C}$ is the projection operator on $C$ w.r.t. some appropriate norm.
\end{minipage}}
\section{Parameter Values used in Various Experiments}
\begin{itemize}[itemsep=2mm,leftmargin=4mm]
\item[]
\begin{table}[h]
 	\caption{The experiment parameter values and the algorithm parameter values used in the cart-pole experiment (Experiment 1)}\label{tab:predcartpole5}
 	\hspace*{0cm}\begin{tabular}{ | l | c |}
 		\specialrule{.2em}{.02em}{.02em} 
 		Gravitational acceleration ($g$) & $9.8\frac{m}{s^{2}}$ \\ \hline
 		Mass of the pole ($m$) & $0.5kg$ \\ \hline
 		Mass of the cart ($M$) & $0.5kg$ \\ \hline
 		Length of the pole ($l$) & $0.6m$ \\ \hline
 		Friction coefficient ($b$) & $0.1N(ms)^{-1}$ \\ \hline
 		Integration time step ($\Delta t$) & $0.1s$\\ \hline
 		Standard deviation of $z$ ($\sigma_{2}$) & $0.01$ \\ \hline
 		Discount factor ($\gamma$) & $0.95$ \\ \hline
 		\specialrule{.1em}{.02em}{.02em} 
 	\end{tabular}
 	\hspace*{15mm} \begin{tabular}{ | c | c |}
 		\specialrule{.2em}{.04em}{.04em} 
 		$\alpha_{t}$\hspace*{15mm} & $t^{-1.0}$ \\ \hline
 		$\beta_{t}$\hspace*{15mm} & $t^{-0.6}$ \\ \hline
 		$c_t$\hspace*{15mm} & $0.01$ \\ \hline
 		$\lambda$\hspace*{15mm} & $0.01$ \\ \hline
 		$\epsilon_1$\hspace*{15mm} & $0.95$ \\ \hline
 		$\rho$\hspace*{15mm} & $0.1$ \\
 		\specialrule{.1em}{.02em}{.02em} 
 	\end{tabular}
\end{table}
\item[]
\begin{table}
	\caption{The experiment parameter values and the algorithm parameter values used in the $5$-link actuated pendulum experiment (Experiment 2)}\label{tab:invpendpred5}
	\hspace*{1mm}\begin{tabular}{ | l | c |}
		\specialrule{.2em}{.02em}{.02em} 
		Gravitational acceleration ($g$) & $9.8\frac{m}{s^{2}}$ \\ \hline
		Mass of the pole ($m$) & $1.0kg$ \\ \hline
		Length of the pole ($l$) & $1.0m$ \\ \hline
		Integration time step ($\Delta t$) & $0.1s$\\ \hline
		Discount factor ($\gamma$) & $0.95$ \\ \hline
		\specialrule{.1em}{.02em}{.02em} 
	\end{tabular}
	\hspace*{15mm} \begin{tabular}{ | c | c |}
		\specialrule{.2em}{.04em}{.04em} 
		$\alpha_{t}$\hspace*{25mm} & $0.001$ \\ \hline
		$\beta_{t}$\hspace*{25mm} & $0.05$ \\ \hline
		$c_t$\hspace*{25mm} & $0.05$ \\ \hline
		$\lambda$\hspace*{25mm} & $0.01$ \\ \hline
		$\epsilon_1$\hspace*{25mm} & $0.95$ \\ \hline
		$\rho$\hspace*{25mm} & $0.1$ \\
		\specialrule{.1em}{.02em}{.02em} 
	\end{tabular}
\end{table}
\item[]
\begin{table}[h]
	\begin{center}
		\caption{Algorithm parameter values used in the Baird's $7$-star experiment (Experiment 3)}\label{tab:baird7star5}
		\begin{tabular}{ | c | c |}
			\specialrule{.2em}{.04em}{.04em} 
			$\alpha_{t}$\hspace*{35mm} & $0.001$ \\ \hline
			$\beta_{t}$\hspace*{35mm} & $0.05$ \\ \hline
			$c_t$\hspace*{35mm} & $0.01$ \\ \hline
			$\lambda$\hspace*{35mm} & $0.01$ \\ \hline
			$\epsilon_1$\hspace*{35mm} & $0.8$ \\ \hline
			$\rho$\hspace*{35mm} & $0.1$ \\
			\specialrule{.1em}{.02em}{.02em} 
		\end{tabular}
	\end{center}
\end{table}
\item[]
\begin{table}[h]
	\begin{center}
		\caption{Algorithm parameter values used in the $10$-state ring experiment  (Experiment 4)}\label{tab:10ring5}
		\begin{tabular}{ | c | c |}
			\specialrule{.2em}{.04em}{.04em} 
			$\alpha_{t}$\hspace*{25mm} & $0.001$ \\ \hline
			$\beta_{t}$\hspace*{25mm} & $0.05$ \\ \hline
			$c_t$\hspace*{25mm} & $0.075$ \\ \hline
			$\lambda$\hspace*{25mm} & $0.001$ \\ \hline
			$\epsilon_1$\hspace*{25mm} & $0.85$ \\ \hline
			$\rho$\hspace*{25mm} & $0.1$ \\
			\specialrule{.1em}{.02em}{.02em} 
		\end{tabular}
	\end{center}
\end{table}
\item[]
\begin{table}[!h]
	\begin{center}
		\caption{Algorithm parameter values used in the random MDP experiment (Experiment 5)}\label{tab:lrgmdppred5}
		\begin{tabular}{ | c | c |}
			\specialrule{.2em}{.04em}{.04em} 
			\multicolumn{2}{|c|}{Both RBF \& Fourier Basis}  \\ \hline
			$\alpha_{t}$\hspace*{35mm} & $0.001$\hspace*{10mm} \\ \hline
			$\beta_{t}$\hspace*{35mm} & $0.05$\hspace*{10mm} \\ \hline
			$c_t$\hspace*{35mm} & $0.075$\hspace*{10mm} \\ \hline
			$\lambda$\hspace*{35mm} & $0.001$\hspace*{10mm} \\ \hline
			$\epsilon_1$\hspace*{35mm} & $0.85$\hspace*{10mm} \\ \hline
			$\rho$\hspace*{35mm} & $0.1$\hspace*{10mm} \\						
			\specialrule{.1em}{.02em}{.02em} 
		\end{tabular}
	\end{center}
\end{table}
\item[]
\begin{table}[h]
	\caption{Algorithm parameter values used in the Van Roy and Tsitsiklis non-linear function approximation experiment (Experiment 6.1)}\label{tab:nlfaparvals}
	\centering
	\begin{tabular}{ | c | c | c | c | c | c | c |}
		\specialrule{.2em}{.04em}{.04em} 
		$S(\cdot)$ & $\alpha_{t}$ & $\beta_{t}$ & $\lambda$ & $c_t$ & $\epsilon_1$ & $\rho$ \\ \hline
		$\exp{(rx)}, r = 10^{-6}$ & $\frac{1}{t}$ &$0.9$ & $0.01$ & $0.03$ & $0.95$ & $0.1$  \\ \hline
		\specialrule{.1em}{.02em}{.02em} 
	\end{tabular}
\end{table}
\item[]
\begin{table}[h]
	\caption{Algorithm parameter values used in the Baird's $7$-star non-linear function approximation experiment (Experiment 6.2)}\label{tab:bairdnlfaparvals}
	\centering
	\begin{tabular}{ | c | c | c | c | c | c | c |}
		\specialrule{.2em}{.04em}{.04em} 
		$S(\cdot)$ & $\alpha_{t}$ & $\beta_{t}$ & $\lambda$ & $c_t$ & $\epsilon_1$ & $\rho$ \\ \hline
		$\exp{(rx)}, r = 0.2$ & $0.02$ &$0.1$ & $0.001$ & $0.05$ & $0.8$ & $0.1$  \\ \hline
		\specialrule{.1em}{.02em}{.02em} 
	\end{tabular}
\end{table}
\item[]
\begin{table}[h]
	\caption{Algorithm parameter values used in the $10$-ring MDP non-linear function approximation experiment (Experiment 6.3)}\label{tab:ringnlfaparvals}
	\centering
	\begin{tabular}{ | c | c | c | c | c | c | c |}
		\specialrule{.2em}{.04em}{.04em} 
		$S(\cdot)$ & $\alpha_{t}$ & $\beta_{t}$ & $\lambda$ & $c_t$ & $\epsilon_1$ & $\rho$ \\ \hline
		$\exp{(rx)}, r = 0.05$ & $0.04$ &$0.2$ & $0.001$ & $0.08$ & $0.8$ & $0.1$  \\ \hline
		\specialrule{.1em}{.02em}{.02em} 
	\end{tabular}
\end{table}
\end{itemize}
\clearpage
\section{Illustration of CE Optimization Procedure}
\begin{figure}[h]
	\centering
	\hspace*{-10mm}{\includegraphics[height=30mm, width=98mm]{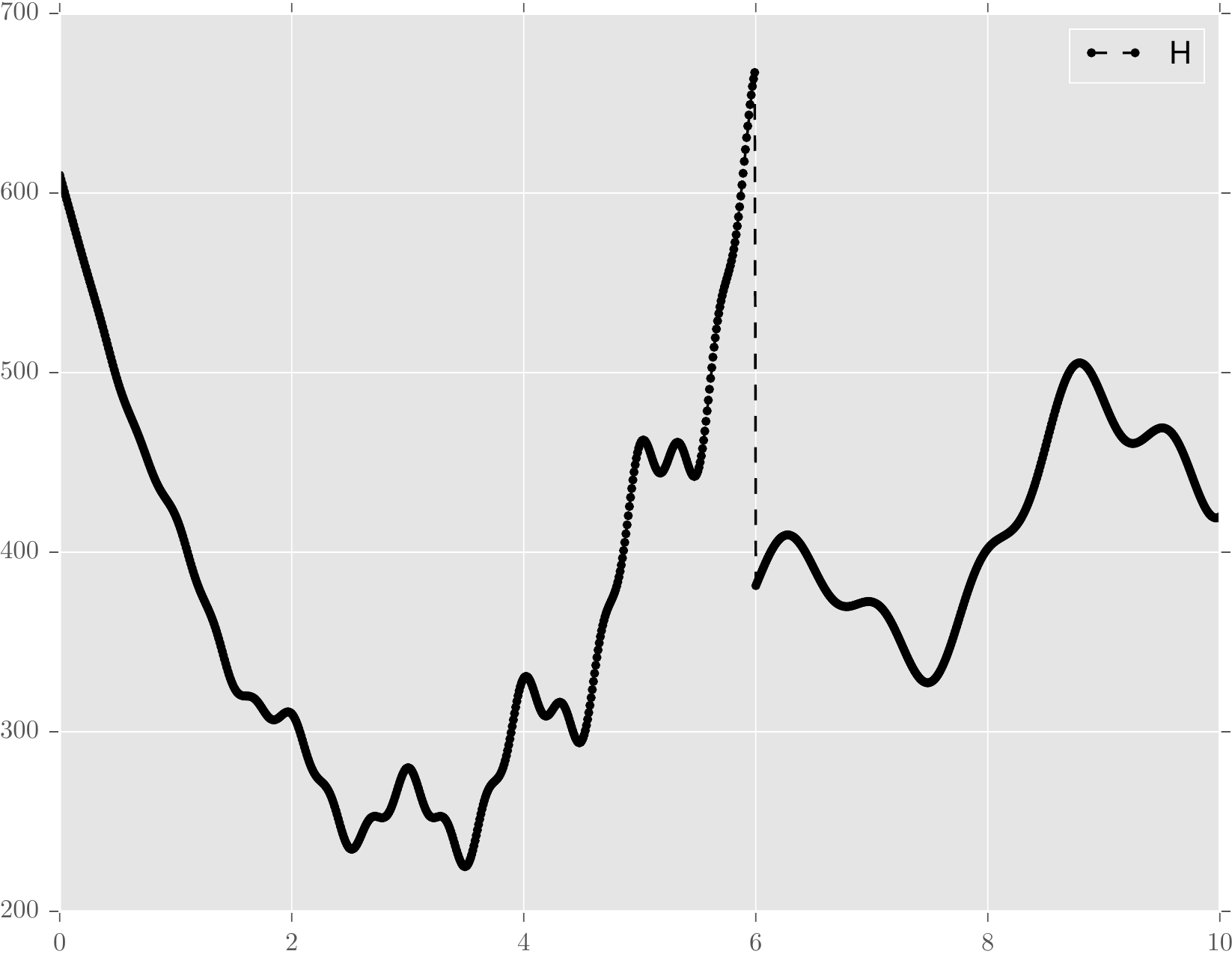}}\\
	\textbf{The objective function $\mathcal{H}:\bbbr \rightarrow \bbbr$ with global maximum at $x^{*} = 6.0$. The function also has a discontinuity at $x^{*}$.}
	\vspace*{2mm}\\
	{\includegraphics[height=85mm, width=125mm]{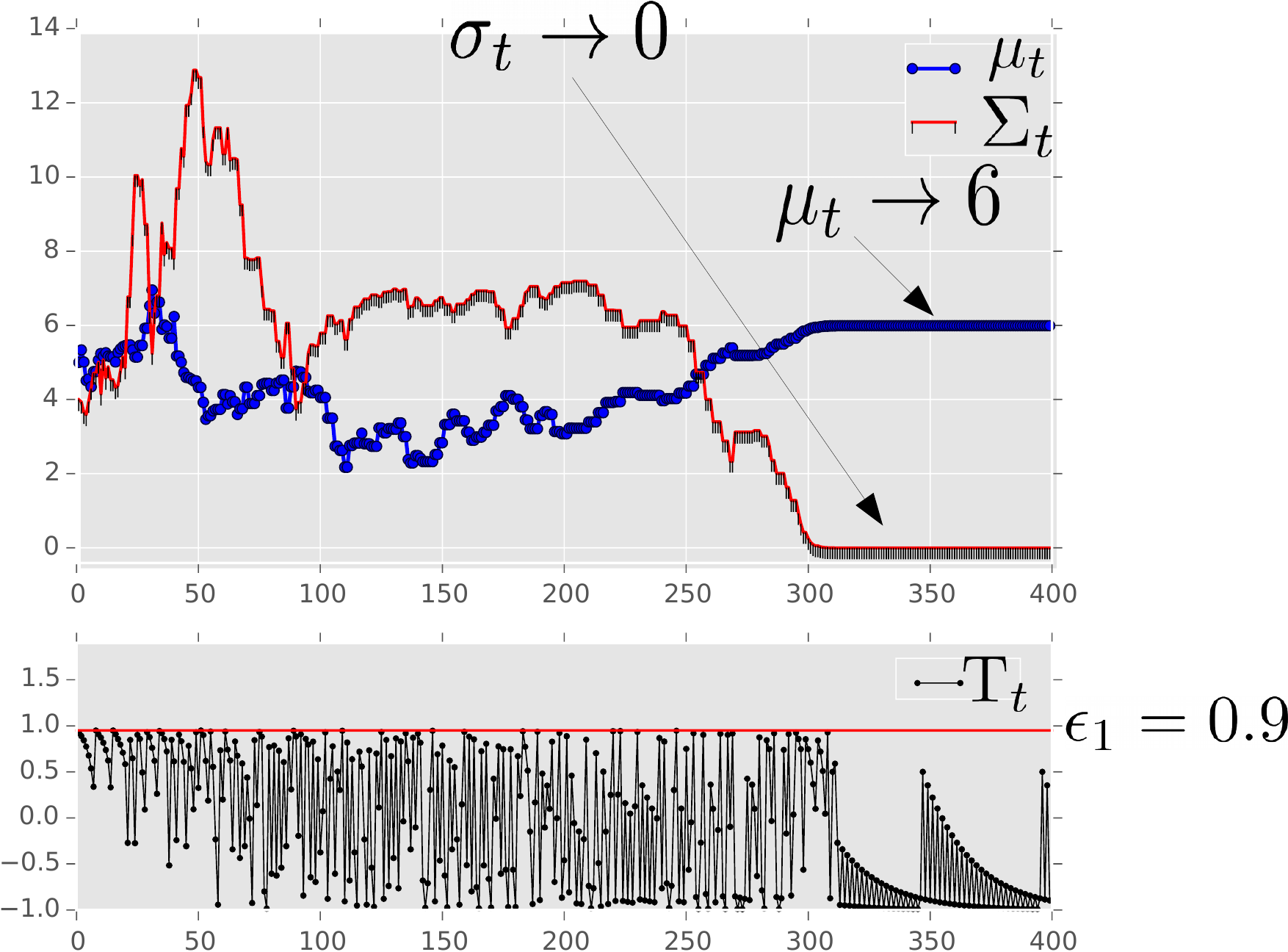}}\\
	\textbf{Here the model parameter $\theta_t$ is given by $\theta_t = (\mu_t, \sigma_t)^{\top}$, where $\mu_t \in \bbbr$ and $\sigma_t \in \bbbr_{+}$. The mean parameter $\mu_{t}$ converges to $x^{*}$ and the variance parameter $\sigma_{t}$  converges to 0. }\\
	\textbf{The top horizontal line in the third figure is $\epsilon_{1}=0.9$. In the third figure, note that $T_{t}$ hits $\epsilon_{1}$ many times till $\theta_{t}$ converges. But once $\theta_{t}$ reaches its limit, $T_{t}$ ceases to hit $\epsilon_1$.}\\
	\caption{Illustration of the CE method on a deterministic optimization problem}\label{fig:evtrack}
\end{figure}
\section{Borkar-Meyn Theorem (Theorem 2.1 of \cite{borkar2000ode})}
\begin{theorem}
	\textbf{For the stochastic recursion of $x_{n} \in \bbbr^{d}$ given by}\\
	\begin{flalign}
		x_{n+1} = x_{n} + a_{n}\left(h(x_{n}) + \mathbb{M}_{n+1}\right), \hspace*{2mm} n \in \mathbb{N},
	\end{flalign}
\noindent
	\textbf{if the following assumptions are satisfied:}
\begin{itemize}
\item
		The map $h:\bbbr^{d} \rightarrow \bbbr^{d}$ is Lipschitz, i.e., $\Vert h(x) - h(y) \Vert \leq L\Vert x - y \Vert$, for some $0 < L < \infty$.
\item
	Step-sizes $\{a_n\}$ are positive scalars satisfying\\
\begin{flalign*}
	\sum_{n} a_n = \infty, \hspace*{2mm}\sum_{n} a_n^{2} < \infty.
\end{flalign*}
\item
	$\{\mathbb{M}_{n+1}\}_{n \in \mathbb{N}}$ is a martingale difference noise w.r.t. the increasing family of $\sigma$-fields\\
		\begin{flalign*}
			\mathcal{F}_{n} \triangleq \sigma(x_{m},\mathbb{M}_{m},m \leq n), \hspace*{2mm} n \in \mathbb{N}.
		\end{flalign*}
That is,
	\begin{flalign*}
		\mathbb{E}\left[\mathbb{M}_{n+1} \vert \mathcal{F}_{n}\right] = 0 \hspace*{2mm} a.s., \hspace*{2mm} n \in \mathbb{N}.
	\end{flalign*}
	Furthermore, $\{\mathbb{M}_{n+1}\}_{n \in \mathbb{N}}$ are square-integrable with\\
	\begin{flalign*}
		\mathbb{E}\left[\Vert \mathbb{M}_{n+1} \Vert^{2} \vert \mathcal{F}_{n}\right] \leq K(1+\Vert x_{n} \Vert^{2}) \hspace*{2mm} a.s., \hspace*{2mm} n \in \mathbb{N},
	\end{flalign*}
	for some constant $K > 0$.
\item
	The functions $h_{c}(x) \triangleq \frac{h(cx)}{x}$, $c \geq 1$, $x \in \bbbr^{d}$, satisfy $h_{c}(x) \rightarrow h_{\infty}(x)$ as $c \rightarrow \infty$, uniformly on compacts for some $h_{\infty} \in C(\bbbr^{d})$. Furthermore, the ODE 
\begin{flalign}
\dot{x}(t) = h_{\infty}(x(t))
\end{flalign}
has the origin as its unique globally asymptotically stable equilibrium,
\end{itemize}
	\textbf{then}
	\begin{flalign*}
		\sup_{n \in \mathbb{N}} \Vert x_{n} \Vert < \infty \hspace*{2mm}a.s.
	\end{flalign*}
\end{theorem}
\end{appendices}
\clearpage


\bibliographystyle{spmpsci}      


\bibliography{cepredict.bib}

\end{document}